\pdfoutput=1

\documentclass{article} 

\usepackage{iclr2025_conference,times}

\usepackage{amsmath,amsfonts,bm}

\newcommand{\figleft}{{\em (Left)}}

\newcommand{\figright}{{\em (Right)}}
\newcommand{\figtop}{{\em (Top)}}
\newcommand{\figbottom}{{\em (Bottom)}}

\def\eqref#1{equation~\ref{#1}}

\def\1{\bm{1}}

\DeclareMathAlphabet{\mathsfit}{\encodingdefault}{\sfdefault}{m}{sl}
\SetMathAlphabet{\mathsfit}{bold}{\encodingdefault}{\sfdefault}{bx}{n}

\usepackage[utf8]{inputenc} 
\usepackage[T1]{fontenc}    
\usepackage{hyperref}       
\usepackage{url}            
\usepackage{booktabs}       
\usepackage{nicefrac}       
\usepackage{microtype}      
\usepackage{xcolor}         

\usepackage{amssymb} 
\usepackage{pifont} 
\usepackage{tabularx} 
\usepackage{arydshln} 
\usepackage{soul,todonotes} 
\usepackage{mathtools} 
\usepackage{adjustbox} 
\usepackage{multirow} 
\usepackage{enumitem}

\usepackage{CJKutf8} 

\definecolor{citecolor}{HTML}{0071BC}
\definecolor{linkcolor}{HTML}{ED1C24}
\definecolor{darkblue}{rgb}{0, 0, 0.5}
\hypersetup{pagebackref=true, breaklinks=true, letterpaper=true,bookmarksnumbered=true, colorlinks=true, citecolor=darkblue, linkcolor=linkcolor, urlcolor=purple} 
\usepackage{bbding} 
\usepackage{wasysym}
\usepackage{graphicx}
\usepackage{wrapfig}
\usepackage{tikz}
\usepackage[edges]{forest}
\usepackage{pgfplots}
\usepackage{nicefrac}
\usepackage{verbatim}
\usepackage[british,american]{babel}
\usepackage{courier}

\usepackage{hwemoji}

\usepackage[ruled,linesnumbered]{algorithm2e}
\SetKwInOut{Parameter}{Parameters}
\usepackage{amsthm} 
\theoremstyle{definition}

\usepackage{makecell}
\usepackage{tabulary}
\definecolor{demphcolor}{RGB}{144,144,144}
\newcommand{\demph}[1]{\textcolor{demphcolor}{#1}}
\definecolor{mygray}{gray}{0.4}

\newlength\savewidth

\newcommand{\tablestyle}[2]{\setlength{\tabcolsep}{#1}\renewcommand{\arraystretch}{#2}\centering} 

\newcommand{\vqa}[1]{VQAv2}
\newcommand{\datasetname}[0]{\textsc{MMGiC}}
\newcommand{\datasetnamecaptiononly}[0]{\datasetname{}(C)}
\newcommand{\datasetnamenoregion}[0]{\datasetname{}(CLD)}

\newcommand{\datasetnamefull}[0]{\datasetname{}(CLDR)}
\newcommand{\icdatasetname}[0]{{IC}}
\newcommand{\icdatasetnamepart}[0]{\icdatasetname{}-\textsc{Part}}
\newcommand{\datasetnamewithic}[0]{\datasetname{}\,\&\,\icdatasetname{}}
\newcommand{\sftprefix}{{MLLM-}}
\newcommand{\methodname}[0]{\sftprefix{}\datasetname{}}
\newcommand{\methodnameic}[0]{\sftprefix{}IC}
\newcommand{\methodnamewithic}[0]{\sftprefix{}\datasetnamewithic{}}
\newcommand{\methodnameall}[0]{\sftprefix{}\{\icdatasetname{}, \datasetname{}, \datasetnamewithic{}\}}

\definecolor{indexcolor}{HTML}{6200EA}
\definecolor{traincolor}{HTML}{F2F2F2}
\definecolor{templatecolor}{HTML}{527CC8}
\definecolor{imageplaceholdercolor}{HTML}{EE7D31}
\definecolor{annotationplaceholdercolor}{HTML}{028310}

\definecolor{kmycolor}{rgb}{0.858, 0.188, 0.478}

\newcommand{\cmark}{\ding{51}}

\newcommand{\Tref}[1]{Table~\ref{#1}}

\newcommand{\Fref}[1]{Figure~\ref{#1}}
\newcommand{\Sref}[1]{Section~\ref{#1}}

\newcommand{\Apref}[1]{Appendix~\ref{#1}}

\newcommand{\fref}[1]{Fig.~\ref{#1}}
\newcommand{\sref}[1]{Sec.~\ref{#1}}

\newcommand{\apref}[1]{App.~\ref{#1}}

\newcommand{\eg}[0]{\textit{e.g.},}
\newcommand{\ie}[0]{\textit{i.e.},}
\newcommand{\etc}[0]{\textit{etc}.}

\definecolor{gain}{HTML}{34a853}  %

\definecolor{lost}{HTML}{ea4335}  %

\definecolor{baselinecolor}{gray}{.9}

\title{
    Exploring Multi-Grained Concept Annotations \\ for Multimodal Large Language Models
}

\iclrfinalcopy

\author{
    Xiao Xu\textsuperscript{\rm 1}\thanks{Work done while visiting the National University of Singapore.}, \ 
	Tianhao Niu\textsuperscript{\rm 1},
    Yuxi Xie\textsuperscript{\rm 2},
    Libo Qin\textsuperscript{\rm 3},
    Wanxiang Che\textsuperscript{\rm 1\thanks{Corresponding author.}},
    Min-Yen Kan\textsuperscript{\rm 2$^\dagger$} \\
    \textsuperscript{\rm 1}Harbin Institute of Technology,
    \textsuperscript{\rm 2}National University of Singapore,
    \textsuperscript{\rm 3}Central South University\\
    \texttt{\{xxu,thniu,car\}@ir.hit.edu.cn}, \texttt{lbqin@csu.edu.cn}, \\
    \texttt{xieyuxi@u.nus.edu}, \texttt{kanmy@comp.nus.edu.sg}
}

\begin{document}

\maketitle
\begin{abstract}
   \label{sec:abstract}
   \setcounter{footnote}{0}
   Multimodal Large Language Models (MLLMs) excel in vision--language tasks by pre-training solely on coarse-grained concept annotations (\eg{} image captions).
   We hypothesize that integrating fine-grained concept annotations (\eg{} object labels and object regions) will further improve performance, as both data granularities complement each other in terms of breadth and depth in concept representation.
   
   We introduce a new dataset featuring \textbf{M}ultimodal \textbf{M}ulti-\textbf{G}ra\textbf{i}ned \textbf{C}oncept annotations (\textbf{\datasetname{}}) for MLLMs.
   In constructing \datasetname{}, we explore the impact of different data recipes on multimodal comprehension and generation.
   Our analyses reveal that multi-grained concept annotations integrate and complement each other, under our structured template and a general MLLM framework.
   
   We clearly explore and demonstrate the potential of \datasetname{} to help MLLMs better locate and learn concepts, aligning vision and language at multiple granularities.
   We further validate our hypothesis by investigating the fair comparison and effective collaboration between \datasetname{} and image--caption data on $12$ multimodal comprehension and generation benchmarks, \eg{} their appropriate combination achieve $3.95\%$ and $2.34\%$ absolute improvements over image--caption data alone on POPE and SEED-Bench.
   Code, data and models will be available at \url{https://github.com/LooperXX/MMGiC}.
\end{abstract}
\section{Introduction}
\label{sec:introduction}

With the rapid development of Large Language Models (LLMs)~\citep{brown2020language,chowdhery2023palm,touvron2023llama,touvron2023llama2,qin2024large} and Visual Foundation Models (VFMs)~\citep{radford2021learning,rombach2022high,dehghani2023scaling}, researchers have started to explore the potential of unifying them into Multimodal Large Language Models (MLLMs) to perform various Vision--Language (VL) tasks, such as image captioning, visual question answering, and text-to-image generation.
MLLMs, such as Emu~\citep{sun2023generative}, SEED-LLaMA~\citep{ge2023making} and LaVIT~\citep{jin2023unified}, follow a similar framework that integrates the capabilities of LLMs and VFMs under an autoregressive objective of predicting the next visual or textual token, showing impressive performance on VL tasks.

Despite their success, existing MLLMs typically do not make full use of \textit{concepts} in VL learning, relying on coarse-grained concept annotations (\eg{} image captions) but ignoring fine-grained concept annotations (\eg{} object labels and object regions).
This leads to superficial and incomplete understanding of concepts, limiting VL alignment.
Specifically, by ``concept'' we mean an abstraction and generalization of a group of things having common characteristics~\citep{goguen1969logic,carey2000origin,blouw2016concepts}.
Concepts can be categorized into concrete and abstract concepts by whether they can be sensed by five human senses~\citep{shevade2005collaborative,connell2018interoception}.
Concrete concepts, such as objects, attributes, and relationships, are not only easy to collect and annotate, but also semantically consistent when expressed across modalities~\citep{chen2019cross,xu2020cross,xie2020multi}.\looseness=-1

Hence, many traditional Vision--Language Models (VLMs) combine coarse- and fine-grained concept annotations to better learn cross-modal consistent concrete concepts, thus improving VL alignment~\citep{li2020oscar,zeng2021multi,shen2022k,menon2023visual}.
However, they rely on additional components and loss functions to leverage different-grained concept annotations (\eg{} bounding box prediction), and optimize the multimodal comprehension ability of VLMs at different granularities \textbf{separately} through multitask learning.
Moreover, they \textbf{seldom} explore the potential of multi-grained concept annotations in multimodal generation tasks, such as text-to-image generation, let alone exploring both multimodal comprehension and generation tasks under the same framework.\looseness=-1

As prior work demonstrate that concepts are crucial for VL alignment and combining coarse- and fine-grained concept annotations can better learn concepts, 
we argue that existing MLLMs should make better use of concepts by incorporating multi-grained concept annotations into their training.
To explore the potential of multi-grained concept annotations, we first construct a new multimodal dataset, \datasetname{}, and introduce a general MLLM framework, both of which can be the foundation for the general and comprehensive exploration.
Different from previous VLM work,  we
1) provide multimodal annotations for images, including both textual forms (caption, labels and label descriptions) and visual form (object regions), to \textbf{enrich} multi-grained concept annotations;
2) design a structured template to integrate multimodal multi-grained concept annotations into image--text interleaved documents, to \textbf{leverage} the complex context processing capability of MLLMs;
3) instead of additional components or loss functions used in VLMs, train MLLMs with an autoregressive discrete framework and predict the next visual or textual token in a multimodal discrete token sequence, to \textbf{improve} multimodal comprehension and generation ability \textbf{across} multiple granularities \textbf{simultaneously}.
This can reuse existing LLM training regimes and \textbf{ensure} the generality and applicability of our \textbf{exploration} based on \datasetname{} across different MLLM frameworks.
The key findings of our exploration are as follows:\looseness=-1
\begin{itemize}[leftmargin=*]
    \item We introduce \datasetname{}, a new dataset with multimodal multi-grained concept annotations (\sref{sec:dataset}). Under a general MLLM framework (\sref{sec:method}), we show the potential of \datasetname{} to help MLLMs better locate and learn concepts, aligning vision and language across multiple granularities (\sref{sec:experiment}).
    \item We explore different data recipes for multi-grained concept annotations (\sref{sec:experiment:data_recipes} \& \ref{sec:experiment:meso}). Our analyses show that multi-grained annotations can integrate and complement each other to help MLLMs ground concepts in the textual annotations to corresponding regions in images, thus improving the ability to understand and generate concepts.
    \item Through evaluation on $12$ multimodal comprehension and generation benchmarks in pre-training (\sref{sec:experiment:scalable}) and supervised fine-tuning stages (\sref{sec:experiment:sft}), we explore the fair comparison and effective collaboration between \datasetname{} and coarse-grained image--caption data. 
    We find that they each have their own strengths in depth and breadth of concept representation, and that appropriate curriculum learning strategies can effectively combine their strengths to further improve performance.
\end{itemize}
\section{\datasetname{} Dataset}
\label{sec:dataset}
To be clear, our goal is \textbf{not} to supplant existing image--caption datasets, but to build a multimodal dataset with multi-grained concept annotations to address the lack of such datasets, and \textbf{explore} its potential in MLLMs.
We now introduce its collection, pre-processing, complement and construction.\looseness=-1

\subsection{Collection and Pre-processing}
\label{sec:dataset:collection}
In this work, we focus on concrete concepts, especially objects, attributes of objects, and relationships between objects. 
They are fundamental elements in VL learning and widely annotated in object detection datasets.
Therefore, we collect four public large-scale human-annotated object detection datasets, including Open Images~\citep{OpenImages}, Objects365~\citep{shao2019objects365}, V3Det~\citep{wang2023v3det}, and Visual Genome~\citep{krishna2017visual}.
Images in these datasets are uploaded to Flickr by real-world users and collected by dataset providers.
They typically show complex scenes with multiple objects, and are annotated with fine-grained category labels and object bounding boxes.
Comparing with widely-used coarse-grained image--caption datasets, fine-grained object annotations provided in these datasets can help MLLMs locate and learn concepts in images.

\begin{figure}[t]
	\centering
	\includegraphics[width=1.0\textwidth]{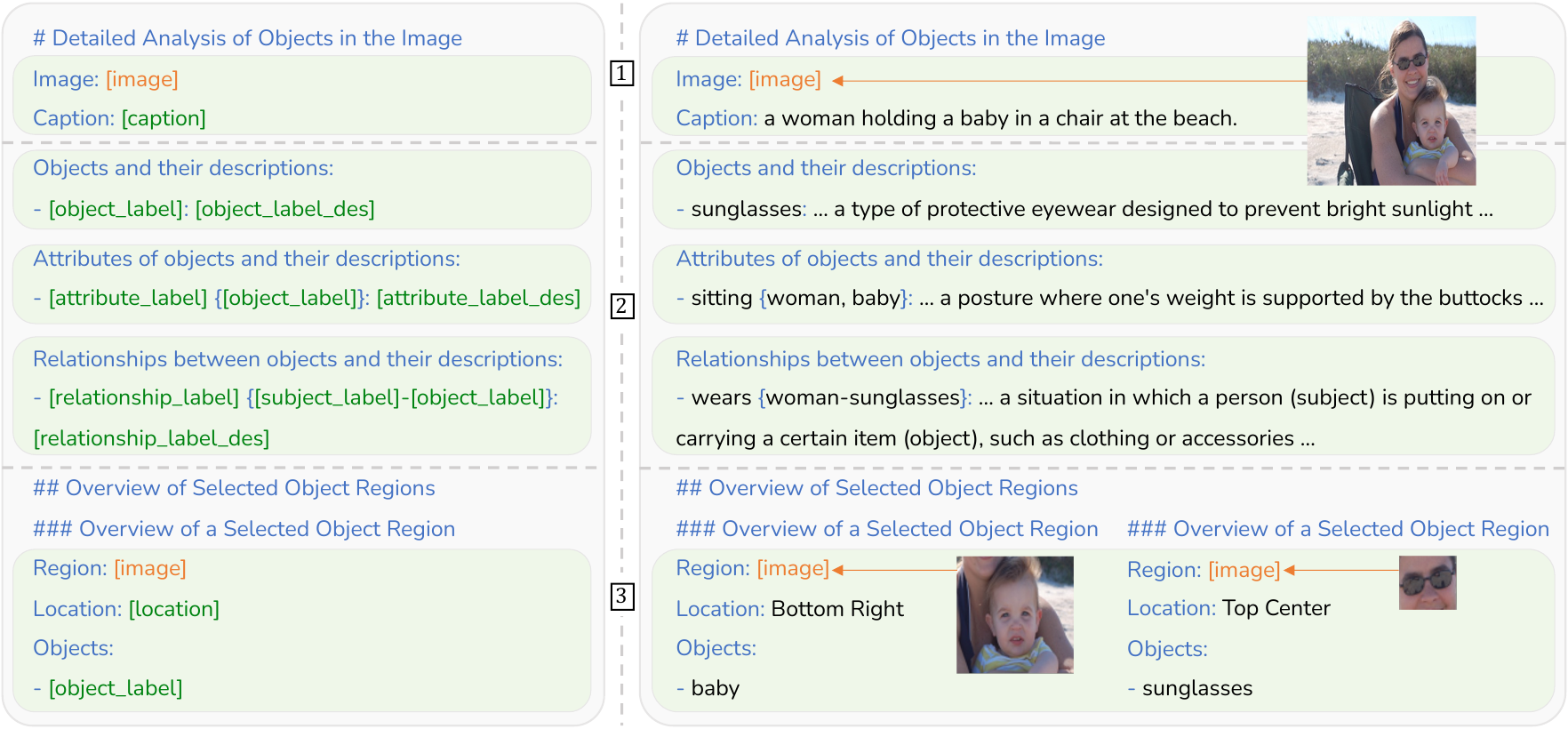}
	\caption{
		Structured template \figleft{} and data example \figright{} of \datasetname{}.
        Different colored text indicates \textcolor{templatecolor}{template text}, \textcolor{imageplaceholdercolor}{image placeholders}, \textcolor{annotationplaceholdercolor}{annotation placeholders} and multi-grained concept annotations, respectively. 
		Each image--text interleaved data sample will be tokenized into discrete tokens.\looseness=-1
	}
	\label{fig:data}
\end{figure}

\paragraph{Object Annotation Pre-processing.}
Fine-grained object annotations includes category labels and bounding box coordinates for each object region.
To accommodate varying aspect ratios of bounding boxes and the requirement for a square input image, 
we crop a new larger square region $S_i$ that contains the original object region $R_i, R_i \subseteq S_i$, with their centers aligned as closely as possible.
We then update the annotations of $S_i$ by integrating the category label of $R_i$ with the category labels of surrounding object regions.
Notably, instead of transforming bounding box coordinates into tokens in the text~\citep{chen2021pix2seq,liu2023visual,peng2023kosmos} or visual markers in the image~\citep{shtedritski2023does,yang2023set,yao2024cpt}, 
for each object, \datasetname{} directly provides visual tokens of the cropped region $S_i$ and textual tokens of fine-grained category labels and location descriptions (\Fref{fig:data} {\scriptsize $\boxed{3}$}~).
Fine-grained cropped regions can help MLLMs locate and align concepts in images and in annotations at a detailed level.

\subsection{Multi-Grained Concept Annotation Complement}
\label{sec:dataset:annotation}
We follow LAION-COCO~\citep{laion2023laioncoco} to automatically synthesize captions for all images (\Fref{fig:data} {\scriptsize $\boxed{1}$}~) by BLIP-2~\citep{li2023blip} and CLIP~\citep{radford2021learning}, since partial images are not annotated with captions.
We do not use or synthesize captions for object regions to avoid introducing hallucinations.\looseness=-1

\paragraph{Label Description Generation.}
Label descriptions are corresponding concept descriptions of concrete concepts in the image, which convey understanding about a concept by visually observed details and relevant knowledge.
Previous works~\citep{shen2022k,yao2022detclip,menon2023visual} successfully improve concept understanding by introducing label descriptions from WordNet~\citep{miller1995wordnet} and LLMs~\citep{brown2020language}.
Inspired by them, we design prompt templates and several human-annotated examples for each type of category labels (object, object attribute, and relationship between objects), and then generate label-description pairs (\Fref{fig:data} {\scriptsize $\boxed{2}$}~) by GPT-4~\citep{achiam2023gpt}.
We manually check them to ensure quality.
Moreover, for better differentiation of polysemous category labels, we manually check them based on the definitions in WordNet, and update them based on the specific data samples, \eg{} ``batter'' $\rightarrow$ ``batter (ballplayer)'' or ``batter (cooking)''.\looseness=-1

\subsection{Construction}
\label{sec:dataset:construction}
Above steps transform four object detection datasets into \datasetname{}, a multimodal dataset with multi-grained concept annotations.
Furthermore, we carefully design a structured template to integrate multi-grained concept annotations into an image--text interleaved document.
As shown in \Fref{fig:data} \figleft{}, the structured template consists of:
{\scriptsize $\boxed{1}$} coarse-grained image-annotation part: each image is annotated with a short and general description of the whole image; 
{\scriptsize $\boxed{2}$} fine-grained image-annotation part: concrete concepts (objects, attributes and relationships) present in the image are annotated with corresponding category labels and label descriptions; 
{\scriptsize $\boxed{3}$} fine-grained object-annotation part: each object in the image is annotated with a cropped object region, object labels in the region, and a location description.
A data example of \datasetname{} is shown in \Fref{fig:data} \figright{}.

Different from previous VLM work that provide multiple \textbf{isolated} annotations for each image, \datasetname{} provides richer multimodal multi-grained concept annotations in both textual forms (caption, labels and label descriptions) and visual form (object regions) for each image, and integrates them into an image--text interleaved document by our structured template.
This can \textbf{leverage} MLLMs' complex context processing capability to facilitate VL alignment \textbf{across} multiple granularities \textbf{simultaneously} under our MLLM framework.
In a nutshell, \datasetname{} fills the gap in the MLLM field for datasets with multi-grained concept annotations.
It contains $3.5$M unique images, $23.9$M unique object regions, and $61.8$M category label--description pairs.
Based on \datasetname{}, we explore and analyse different data recipes for multi-grained concept annotations (\sref{sec:experiment:data_recipes} \& \ref{sec:experiment:meso}), and further compare \datasetname{} with image--caption data (\sref{sec:experiment:scalable} \& \ref{sec:experiment:sft}).
More data details are shown in \Apref{sec:appendix:dataset}.\looseness=-1

\section{Framework}
\label{sec:method}
We introduce a general MLLM framework and its two training stages.
Our goal is not to develop new frameworks, training objectives or benchmark SOTAs, but to explore the potential of multi-grained concept annotations for MLLMs under the general framework.

\subsection{An Autoregressive Discrete MLLM Framework}
\label{sec:method:framework}
\begin{wrapfigure}{r}{0.45\textwidth}
  \vspace{-1.5em}
  \begin{center}
    \includegraphics[width=0.45\textwidth]{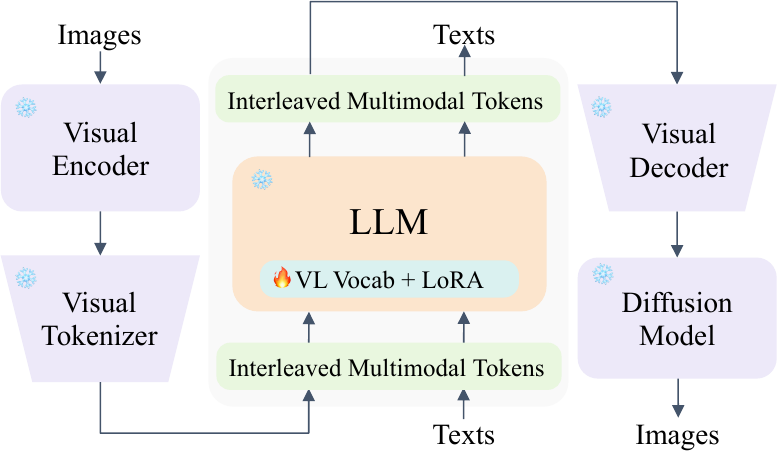}
  \end{center}
  \caption{Illustration of our general MLLM framework.
      Only the LLM are loaded and partially fine-tuned during training.
  }
  \label{fig:method}
\end{wrapfigure}

Based on previous works~\citep{ge2023making,jin2023unified}, we standardize a framework consisting of several visual modules and a LLM with an extended VL vocabulary (\fref{fig:method}).
It is trained with an autoregressive objective to generate predictions of the next token in a discrete sequence of image--text interleaved tokens, and can support our exploration in multimodal comprehension and generation.\looseness=-1

\paragraph{Visual Modules.}
Inherited from LaVIT~\citep{jin2023unified}, the visual modules consist of a visual encoder, a visual tokenizer, a visual decoder and a diffusion model.
The visual encoder is a pre-trained vision transformer~\citep{dosovitskiy2020image,sun2023eva}, which encodes an image into a sequence of visual embeddings.
The visual tokenizer quantizes these embeddings into a sequence of discrete visual tokens by a visual codebook~\citep{van2017neural}.
The visual decoder reconstructs predicted visual tokens into a sequence of visual embeddings, which are then taken as the condition of the diffusion model~\citep{sohl2015deep,ho2020denoising,rombach2022high} to progressively generate target image pixels from a Gaussian noise.
Overall, visual modules can tokenize an image into a sequence of discrete visual tokens as input and decode predicted visual tokens back into an image.\looseness=-1

\paragraph{LLM with an Extended VL Vocabulary.}
The LLM is inherited from LLaMA-2-7B~\citep{touvron2023llama2}, and its vocabulary is extended to support both textual and visual tokens.
Since visual tokens in the VL vocabulary correspond one-to-one with visual latent codes in the visual codebook, instead of initializing visual token embeddings with the distribution of original textual token embeddings~\citep{ge2023making,jin2023unified}, we directly initialize them with visual latent codes and a projection matrix (More details in \Apref{sec:appendix:discussion:framework}).
The LLM can process a sequence of interleaved visual and textual tokens, and predict the next visual or textual token in an autoregressive manner.

To pursue simplicity, efficiency, and scalability~\citep{ge2023making}, instead of additional components or loss functions used in VLMs, we train MLLMs with a single autoregressive objective.
This allows for a fair comparison between \datasetname{} and coarse-grained image--caption data under the same framework, and ensures the generality and applicability of \datasetname{} to different MLLM frameworks.
More importantly, our framework facilitates VL alignment \textbf{across} multiple granularities \textbf{simultaneously} through \datasetname{} and MLLMs' complex context processing capability.
In all training stages, we follow common practice~\citep{ge2023planting,zhu2023vl} to \textbf{freeze} most of the parameters and only tune partial parameters of the LLM: a VL vocabulary, additional LoRA modules~\citep{hu2021lora}, norm layers and a language modeling head layer, 
to greatly improve efficiency.
We pre-tokenize images into discrete visual token sequences and do not load all visual modules during training since they are frozen.\looseness=-1

\subsection{Training Stage 1: Pre-training (PT)}
\label{sec:method:pretraining}
Similar to LLMs that tokenize text-only documents into discrete textual token sequences, for each data sample in \datasetname{} shown in \fref{fig:data} \figright{} or an image--caption pair, our framework tokenize it into a discrete token sequence consisting of interleaved visual and textual token sequences.
To help the LLM distinguish between two types of sequences, following LaVIT, we add two special tokens \texttt{[IMG]} and \texttt{[/IMG]} to the vocabulary, and insert them before and after each visual token sequence, respectively.
Our framework is trained with an autoregressive objective to maximize the likelihood of predicting the next visual or textual token.
Details of training settings are provided in \Apref{sec:appendix:train_details}.

\subsection{Training Stage 2: Supervised Fine-tuning (SFT)}
\label{sec:method:fine-tuning}
To align pre-trained MLLMs with natural language instructions, following previous works~\citep{sun2023generative,sun2023generative2,liu2023visual,liu2023improvedllava,liu2024llavanext,zhu2023vl,hu2024large}, we collect $1.21$M samples from several public datasets for supervised fine-tuning, including multimodal instruction datasets~\citep{liu2023improvedllava,yu2023reformulating,zhang2023llavar,chen2023sharegpt4v,laiongpt4v,brooks2023instructpix2pix,zhang2024magicbrush} and text-only instruction datasets~\citep{alpaca,sharegpt}, and $1$M samples from an aesthetic image--caption dataset~\citep{laioncocoaesthetic}.
We also play back $1$M samples from \datasetname{} to avoid forgetting the knowledge learned in the pre-training stage.
Following LLaVA~\citep{liu2023visual,liu2023improvedllava,liu2024llavanext}, all datasets are transformed into a unified format, which consists of a general system message and multiple instruction--answer pairs. Only answer tokens are accounted in loss calculation.
Details of instruction data are provided in \Apref{sec:appendix:sft_data}.

\section{Experiment}
\label{sec:experiment}
Based on \datasetname{} and a general MLLM framework, we can \textbf{explore} the potential of multi-grained concept annotations for MLLMs on multimodal comprehension and generation benchmarks in pre-training and SFT stages.
Specifically, we follow previous works~\citep{ge2023making,liu2023visual,zhu2023vl} to focus on the zero-shot multimodal capabilities, including image captioning (COCO~\citep{chen2015microsoft}, NoCaps~\citep{agrawal2019nocaps}), text-to-image generation (COCO~\citep{chen2015microsoft}, VIST~\citep{huang-etal-2016-visual}), visual question answering (VQAv2~\citep{balanced_vqa_v2}, GQA~\citep{hudson2019gqa}, VizWiz~\citep{gurari2018vizwiz}), comprehensive multi-choice benchmarks (POPE~\citep{li-etal-2023-evaluating}, MME~\citep{yin2023survey}, MMBench~\citep{liu2023mmbench}, ScienceQA~\citep{lu2022learn}, SEED-Bench~\citep{li2023seed}).
More evaluation details are provided in \Apref{sec:appendix:eval_details}.\looseness=-1

\subsection{Data Recipes for Multi-Grained Concept Annotations}
\label{sec:experiment:data_recipes}

\begin{figure}[t]
	\centering
	\includegraphics[width=0.95\textwidth]{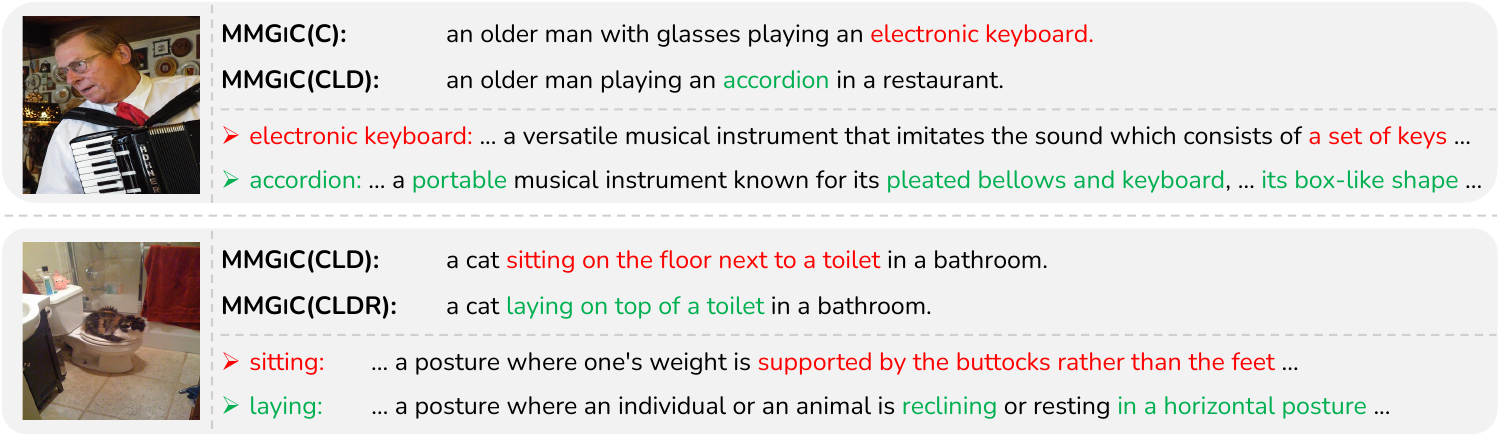}
	\caption{
		Comparison of generated captions by MLLMs pre-trained with different data recipes.
        \datasetnamecaptiononly{}, \datasetnamenoregion{} and \datasetnamefull{} denote the \textcolor{indexcolor}{\{$0,2,3$\}}-th data recipes in \Tref{tab:data_recipes}, respectively.
        The bottom right of each example shows associated label--description pairs from \datasetname{}.\looseness=-1
	}
	\label{fig:case_study_1}
\end{figure}

As mentioned in \sref{sec:dataset:construction} and \fref{fig:data}, multi-grained concept annotations include four components: coarse-grained image captions (C), fine-grained category labels (L), label descriptions (D) and object regions (R).
Hence, we design four data recipes to investigate the importance of each component on zero-shot image captioning and image generation tasks and find the best recipe.
As shown in \Tref{tab:data_recipes}:\looseness=-1
\begin{itemize}[leftmargin=2em, topsep=0em, itemsep=0em]
    \item \textcolor{indexcolor}{\{$0,1,2$\}}-th rows: simply appending category labels to image captions does not help and even hurts the performance on both tasks. 
    MLLMs may struggle to understand the association between images, captions and category labels, leading to confusion and treating category labels as additional noise.
    Whereas, label descriptions can strengthen the association between images and annotations, to help MLLMs align the concepts represented by labels with the concepts in the image, thus mitigating the performance drop.
    As shown in \Fref{fig:case_study_1} \figtop{}, \datasetnamecaptiononly{} incorrectly identifies ``accordion'' as ``electronic keyboard''. While for \datasetnamenoregion{}, label descriptions can help MLLMs better understand labels by \textbf{visual details} ``pleated bellows and keyboard, box-like'' and \textbf{relevant knowledge} ``portable''.
    \item \textcolor{indexcolor}{\{$2,3$\}}-th rows: object regions can further \textbf{complement} other annotations to help MLLMs better locate and align concepts in images and in annotations at a fine-grained level, thus significantly improving the performance on both tasks.
    As shown in \Fref{fig:case_study_1} \figbottom{}, \datasetnamenoregion{} incorrectly identifies ``laying'' as ``sitting'' and ``on top of a toilet'' as ``on the floor next to a toilet''.
    While for \datasetnamefull{}, object regions can help MLLMs \textbf{ground} concepts in the textual annotations to corresponding regions in the image, especially in terms of instance interaction and spatial relationship here, which is also consistent with meso analyses in \Sref{sec:experiment:meso}.
\end{itemize}

\begin{table}[t]
    \centering
    \tablestyle{4pt}{1.1}
    \caption{
        Zero-shot evaluation of different data recipes for \datasetname{}.
        C: image caption; L: category labels; D: label descriptions; R: object regions; CLIP-T/I: image--text or image-image similarity via CLIP; $\downarrow$: lower is better, otherwise higher is better.
        The best results are \textbf{bold}.
    }
    \vspace{1em}
    \label{tab:data_recipes}
    \scalebox{0.9}{
        \begin{tabular}{c|cccc|cc|rcccc}
            \toprule
            \multirow{2}{*}{} & \multicolumn{4}{c|}{\multirow{2}{*}{{Data Recipes}}} & \multicolumn{2}{c|}{{Image  Captioning}} & \multicolumn{5}{c}{{Image Generation}} \\ 
            & \multicolumn{4}{c|}{} & {COCO} & {NoCaps} & \multicolumn{3}{c}{{MS-COCO-30K}} & \multicolumn{2}{c}{{VIST}} \\ 
            \midrule
            {} & {C} & {L} & {D} & {R} & {CIDEr} & {CIDEr} & \multicolumn{1}{c}{FID ($\downarrow$)} & {CLIP-T} & {CLIP-I} & {FID} ($\downarrow$) & {CLIP-I} \\
            \midrule
            \textcolor{indexcolor}{0} & \cmark &  &  &  & 113.64 & { }{ }99.11 & \textbf{7.20} & 30.81 & 71.62 & 67.61 & 62.22 \\
            \textcolor{indexcolor}{1} & \cmark & \cmark &  &  & 113.67 & { }{ }96.80 & 10.52 & 30.43 & 71.09 & 71.67 & 61.96 \\
            \textcolor{indexcolor}{2} & \cmark & \cmark & \cmark &  & 116.02 & { }{ }98.61 & 8.92 & 30.89 & 71.60 & 51.89 & 64.30 \\
            \textcolor{indexcolor}{3} & \cmark & \cmark & \cmark & \cmark & \textbf{118.30} & \textbf{102.01} & 7.36 & \textbf{31.57} & \textbf{72.24} & \textbf{35.33} & \textbf{66.10} \\ 
            \bottomrule
        \end{tabular}
    }
\end{table}

\begin{figure}[t]
	\centering
	\includegraphics[width=\textwidth]{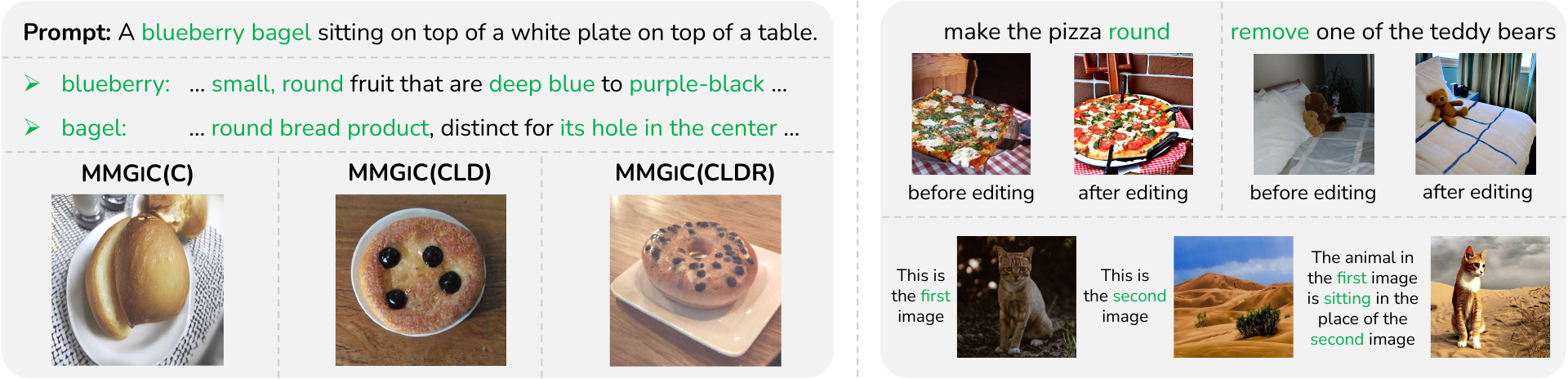}
	\caption{
		Comparison of generated images by MLLMs pre-trained with different data recipes \figleft{} and image editing and multimodal in-context image synthesis examples \figright{}.
	}
	\label{fig:case_study_2}
	\vspace{-1em}
\end{figure}

We further analyse generated images by MLLMs pre-trained with different data recipes in \Fref{fig:case_study_2} \figleft{}.
The input prompt and the label description of ``bagel'' are shown in the top and generated images are shown in the bottom.
We found that while \datasetnamecaptiononly{} could accurately generate "on top of a white plate on top of a table", it fails to correctly generate ``a bagel'', let alone ``a blueberry bagel''.
Label descriptions in \datasetnamenoregion{} can provide more \textbf{visual details} about both ``bagel'' and ``blueberry'', 
which helps MLLM better understand and generate the concept of ``bagel'' with the feature of ``round bread product'', and also add four "blueberries" to th  "bagel".
Furthermore, with the help of object regions in \datasetnamefull{}, MLLMs can correctly understand and generate a "bagel" with the feature of "its hole in the center".
We also show some emergent abilities of MLLMs pre-trained with \datasetname{} in \Fref{fig:case_study_2} \figright{}.
Notably, \datasetname{} does \textbf{not} contain any samples about image editing and in-context image synthesis.
Leveraging image--text interleaved documents with multi-grained concept annotations,
the top two examples show that MLLMs can precisely understand editing instructions and perform appropriate editing, 
while the bottom example shows that MLLMs can synthesize image precisely based on the image--text interleaved sequences.
More analyses are provided in \Apref{sec:appendix:image_editing}.\looseness=-1

In summary, multi-grained concept annotations are not \textbf{isolated} from each other.
With our structured template and MLLM framework, they can \textbf{integrate} into multimodal documents and \textbf{complement} each other to help MLLMs better learn concepts, thus improving the ability to understand and generate concepts.
We take the \textcolor{indexcolor}{$3$}-rd data recipe as the default data recipe and denote it simply as \datasetname{}.

\subsection{Comparison and Collaboration Between \datasetname{} and Image--Caption Data}
\label{sec:experiment:scalable}

\begin{table}[t]
    \centering
    \tablestyle{4pt}{1.1}
    \caption{
        Zero-shot evaluation of comparison and collaboration between \datasetname{} and \icdatasetname{}.
        \icdatasetnamepart{}: select the same number of samples as \datasetname{} from \icdatasetname{}; 
        \datasetname{}$+$\icdatasetname{}: joint training; 
        \datasetname{} $\rightarrow$ \icdatasetname{}: train on \datasetname{} first and then on \icdatasetname{}.
        The best results are \textbf{bold} and the second-best are \underline{underlined}.\looseness=-1
    }
    \vspace{1em}
    \label{tab:scalable}
    \scalebox{0.9}{
        \begin{tabular}{c|l|cc|rcccc}
            \toprule
            & \multirow{3}{*}{{Training Data}} & \multicolumn{2}{c|}{{Image  Captioning}} & \multicolumn{5}{c}{{Image Generation}} \\
            & {} & COCO & NoCaps & \multicolumn{3}{c}{MS-COCO-30K} & \multicolumn{2}{c}{VIST} \\
            \cmidrule{3-9}
            & {} & CIDEr & CIDEr & FID ($\downarrow$) & CLIP-T & CLIP-I & FID ($\downarrow$)& CLIP-I \\
            \midrule
            \textcolor{indexcolor}{0} & \icdatasetnamepart{} & { }{ }95.74 & { }{ }85.00 & 11.62 & 29.94 & 68.21 & 64.03 & 63.41 \\
            \textcolor{indexcolor}{1} & \icdatasetname{} & 104.15 & { }{ }92.24 & 7.65 & 31.40 & 70.27 & 41.65 & 65.06 \\
            \textcolor{indexcolor}{2} & \datasetname{} & 118.30 & 102.01 & 7.36 & 31.57 & \textbf{72.24} & \textbf{35.33} & \textbf{66.10} \\
            \hdashline
            \textcolor{indexcolor}{3} & \datasetname{}$+$\icdatasetname{} & 106.45 & { }{ }92.98 & \textbf{7.11} & 31.65 & 70.93 & 36.54 & \underline{65.89} \\
            \textcolor{indexcolor}{4} & \datasetname{} $\rightarrow$ \icdatasetname{} & 105.62 & { }{ }93.77 & 7.29 & 31.57 & 70.26 & 37.45 & 65.88 \\
            \textcolor{indexcolor}{5} & \icdatasetname{} $\rightarrow$ \datasetname{} & \underline{120.84} & \textbf{105.59} & \underline{7.13} & \textbf{31.96} & 71.54 & 37.13 & 65.62 \\
            \textcolor{indexcolor}{6} & {\datasetname{}$+$\icdatasetname{} $\rightarrow$ \datasetname{}} & \textbf{121.22} & \underline{105.33} & 7.22 & \underline{31.91} & \underline{71.75} & \underline{36.23} & 65.79 \\
            \bottomrule
        \end{tabular}
    }
\end{table}

\datasetname{} is constructed from public object detection datasets mainly covers common concepts.
Widely-used coarse-grained image--caption data, \eg{} Conceptual Captions~\citep{sharma-etal-2018-conceptual} and LAION-5B~\citep{schuhmann2022laion}, typically cover diverse, scalable but noisy concepts.
To further explore the potential of \datasetname{}, we then investigate the \textbf{comparison} and \textbf{collaboration} between \datasetname{} with image--caption data.
To strike a balance between increasing concept breadth, reducing data noise and improving efficiency, 
we collect several large-scale public image--caption datasets~\citep{ordonez2011sbu,sharma-etal-2018-conceptual,changpinyo2021conceptual,schuhmann2022laion,laioncoco,sun2024journeydb}, and follow LLaVA~\citep{liu2023visual} to first filter them by the frequency of noun-phrases extracted by spaCy from their given synthesized captions, and then automatically synthesize captions same as \datasetname{}.
We name this dataset as \icdatasetname{}, where $52$M unique images are collected and selected, almost $15$ times more than \datasetname{}.
More details are provided in \Apref{sec:appendix:ic_dataset}. 
As shown in \Tref{tab:scalable}:\looseness=-1
\begin{itemize}[leftmargin=2em, topsep=0em, itemsep=0em]
    \item \textcolor{indexcolor}{\{$0,1,2$\}}-th rows: comparing with \icdatasetname{}, \datasetname{} can help MLLMs achieve \textbf{significantly} better performance on both tasks even with a much \textbf{smaller} number of samples, which demonstrates the effectiveness of multi-grained concept annotations in concept understanding and generation.
    \item \textcolor{indexcolor}{\{$3,4,5,6$\}}-th rows: 
    naturally, we try to improve the performance and concept breadth of MLLMs by exploring the collaboration between \datasetname{} and \icdatasetname{}.
    However, simply joint training \datasetname{} and \icdatasetname{} achieves better performance than \icdatasetname{}, but significantly worse than \datasetname{}, especially on image captioning and VIST.
    We then follow the curriculum learning strategy~\citep{mccann2019the} to train them in different orders. 
    Interestingly, training on \icdatasetname{} first and then on \datasetname{} achieves significantly better performance than all the above strategies on image captioning and partial metrics of image generation.
    This is consistent with recent findings~\citep{hu2024minicpm,li2024llavanext-ablations} that training with \textbf{high-quality} data \textbf{later} in the pre-training phase leads to better performance. 
    Moreover, considering that the noise in \icdatasetname{} still cause a slight performance drop on VIST (\textcolor{indexcolor}{\{$2,5$\}}-th rows),
    we first jointly train on \datasetname{} and \icdatasetname{} to alleviate the effect of noise in \icdatasetname{}, and then on \datasetname{}, eventually achieving the best average performance (\textcolor{indexcolor}{$6$}-th row).\looseness=-1
\end{itemize}

Overall, by exploring comparison and collaboration between \datasetname{} with large-scale coarse-grained image--caption data in the pre-training stage, we demonstrate that \datasetname{} achieves better performance than \icdatasetname{} on both tasks, and their appropriate collaboration can further \textbf{improve} the average performance. \looseness=-1
We present \textcolor{indexcolor}{\{$1,2,6$\}}-th rows as three baselines, and denote them as \methodnameall{}.
More discussions about collaboration strategies are provided in \Apref{sec:appendix:limitation:collaboration}.

\subsection{Evaluation on Downstream Vision--Language Benchmarks After SFT} 
\label{sec:experiment:sft}
To further explore the potential of \datasetname{} dataset on various downstream VL benchmarks, we then perform SFT (\Sref{sec:method:fine-tuning}) on our three baselines.
Technically, different training datasets, training settings, framework, etc., lead to non-comparable and unfair comparisons of our baselines with existing MLLMs.
Hence, we show SOTA MLLMs in \demph{gray} as upper bound references, and their computation and data resources are extremely expensive and large (well over $10$ times that of our work).\looseness=-1

\begin{table}[t]
    \centering
    \tablestyle{3pt}{1.1}
    \caption{
        Zero-shot evaluation on multimodal comprehension benchmarks after SFT.
        MLLMs in Group \textcolor{indexcolor}{a} are for comprehension only, while MLLMs in Group \textcolor{indexcolor}{b} are for both comprehension and generation.
        MMB: MMBench; SQA$^\mathrm{I}$: ScienceQA-IMG; SEED$^\mathrm{I}$: SEED-Bench-IMG; $^{*}$: w/o SFT.\looseness=-1
    }
    \vspace{1em}
    \label{tab:sft_understanding}
    \scalebox{0.88}{
        \begin{tabular}{c|l|cc|ccc|ccccc}
            \toprule
            & \multirow{2}{*}{Model} & \multicolumn{2}{c|}{Image Captioning} & \multicolumn{3}{c|}{VQA} & \multicolumn{5}{c}{Multi-Choice Benchmark} \\
            &  & COCO & NoCaps & VQAv2 & GQA & VizWiz & POPE & MME & MMB & SQA$^\mathrm{I}$ & SEED$^\mathrm{I}$ \\
            \midrule
            \multicolumn{12}{l}{\demph{\it{SOTA MLLMs as upper bound references, with more training data or trainable parameters, not comparable}}}\\
            \midrule
            \multirow{3}{*}{\textcolor{indexcolor}{a}} & \demph{LLaVA-1.5-7B} &  &  & \demph{78.50} & \demph{62.00} & \demph{50.00} & \demph{85.90} & \demph{1826.80} & \demph{65.20} & \demph{66.80} & \demph{65.80} \\
            & \demph{Emu-I-14B} & \demph{120.40} & \demph{108.80} & \demph{62.00} & \demph{46.00} & \demph{38.30} &  &  &  &  & \demph{58.00} \\
            & \demph{Emu2-Chat-37B} &  &  & \demph{84.90} & \demph{65.10} & \demph{54.90} &  &  & \demph{62.40} &  & \demph{68.90} \\
            \hdashline
            \multirow{3}{*}{\textcolor{indexcolor}{b}} & \demph{VL-GPT-I-7B} & \demph{133.70} &  & \demph{67.20} & \demph{51.50} & \demph{38.90} &  &  &  &  &  \\
            & \demph{LaVIT-v2-7B$^{*}$} & \demph{133.30} & \demph{112.00} & \demph{68.30} & \demph{47.90} & \demph{41.00} &  &  &  &  &  \\
            & \demph{SEED-LLaMA-I-8B} & \demph{124.50} & \demph{{ }{ }97.78} & \demph{66.20} & \demph{52.24} & \demph{55.10} & \demph{79.92} & \demph{1497.53} & \demph{52.58} & \demph{60.24} & \demph{51.50} \\
            \midrule
            \multicolumn{12}{l}{\it{Our comparable baselines}}\\
            \midrule
            \textcolor{indexcolor}{0} & \methodnameic{} & 108.13 & { }{ }92.71 & \underline{70.28} & 56.02 & \underline{52.62} & 81.14 & \underline{1646.71} & \underline{59.54} & \underline{65.94} & 58.41 \\
            \textcolor{indexcolor}{1} & \methodname{} & \underline{119.35} & \underline{104.19} & 70.13 & \underline{56.84} & 51.14 & \underline{83.25} & 1636.47 & 58.51 & 65.79 & \underline{60.03} \\
            \textcolor{indexcolor}{2} & \methodnamewithic{} & \textbf{122.31} & \textbf{106.97} & \textbf{70.57} & \textbf{56.97} & \textbf{52.66} & \textbf{85.09} & \textbf{1668.19} & \textbf{59.88} & \textbf{66.24} & \textbf{60.75} \\
            \bottomrule
        \end{tabular}
    }
    \vspace{-1em}
\end{table}

\begin{table}[t]
    \centering
    \tablestyle{4pt}{1.1}
    \caption{
        Zero-shot evaluation on multimodal generation benchmarks after SFT.
        MLLMs in Group \textcolor{indexcolor}{a} are for generation only, while MLLMs in Group \textcolor{indexcolor}{b} are for both comprehension and generation.
    }
    \label{tab:sft_generation_1}
    \vspace{1em}
    \scalebox{0.85}{
        \begin{tabular}{c|l|ccc|cc}
            \toprule
            & \multirow{2}{*}{Model} & \multicolumn{3}{c|}{MS-COCO-30K} & \multicolumn{2}{c}{VIST} \\
            &  & {FID} ($\downarrow$) & CLIP-T & CLIP-I & {FID} ($\downarrow$) & CLIP-I \\
            \midrule
            \multicolumn{7}{l}{\demph{\it{SOTA MLLMs as upper bound references, not comparable}}}\\
            \midrule
            \multirow{3}{*}{\textcolor{indexcolor}{a}} & \demph{\textsc{Kosmos-G}}~\citep{pan2023kosmosg} & \demph{10.99} &  &  &  &  \\
            & \demph{GILL}~\citep{koh2024generating} & \demph{12.20} &  &  &  & \demph{64.10}  \\
            & \demph{Emu2-Gen-37B}~\citep{sun2023generative2} &  & \demph{29.70} & \demph{68.60} &  &  \\
            \hdashline
            \multirow{3}{*}{\textcolor{indexcolor}{b}} & \demph{VL-GPT-I-7B}~\citep{zhu2023vl} & \demph{11.53} &  &  &  &  \\
            & \demph{LaVIT-v2-7B$^{*}$}~\citep{jin2023unified} & \demph{{ }{ }7.10} & \demph{31.93} & \demph{71.06} & \demph{34.76} & \demph{68.41}  \\
            & \demph{SEED-LLaMA-I-8B}~\citep{ge2023making} & \demph{16.66} & \demph{29.52} & \demph{69.22} & \demph{43.69} & \demph{65.21}  \\
            \midrule
            \multicolumn{7}{l}{\it{Our comparable baselines}}\\
            \midrule
            \textcolor{indexcolor}{0} & \methodnameic{} & { }{ }8.11 & 30.90 & 70.72 & 38.19 & 65.37 \\
            \textcolor{indexcolor}{1} & \methodname{} & { }{ }\textbf{6.79} & \textbf{31.63} & \textbf{72.44} & \textbf{34.32} & \textbf{67.66} \\
            \textcolor{indexcolor}{2} & \methodnamewithic{} & { }{ }\underline{7.29} & \underline{31.54} & \underline{72.03} & \underline{34.39} & \underline{67.19} \\
            \bottomrule
        \end{tabular}
    }
\end{table}

\paragraph{Zero-shot Multimodal Comprehension.}
As shown in \Tref{tab:sft_understanding}, \methodnamewithic{} achieves the best performance on all $10$ benchmarks compared to the other two baselines, and even outperforms some SOTA MLLMs with more training data or full-param training or larger LLMs on some benchmarks.
Moreover, even with less than $4$M pre-training data compared to \methodnameic{} with $52$M pre-training data, \methodname{} significantly outperforms \methodnameic{} on the benchmarks that inspect \textbf{in-depth} understanding of common concrete concepts, \eg{} COCO, NoCaps, POPE, SEED-Bench.
In contrast, \methodnameic{} outperforms \methodname{} on benchmarks that require a \textbf{broader} understanding of concrete concepts, \eg{} VizWiz, MME, MMBench.
More importantly, \methodnamewithic{} can effectively \textbf{combine} the strengths of both in terms of \textbf{depth} and \textbf{breadth} of concept representation and further improve performance, \eg{} $3.95\%$ and $2.34\%$ absolute improvements over \methodnameic{} on POPE and SEED-Bench.
We further analyse in terms of dataset statistics and concept overlap in \apref{sec:appendix:concept_overlap}.\looseness=-1

\paragraph{Zero-shot Multimodal Generation.}
As shown in \Tref{tab:sft_generation_1}, for two text-to-image generation benchmarks that focus on common concrete concepts, \methodname{} achieves the best performance, and matches or even outperforms some SOTA MLLMs on some metrics. 
This demonstrates that fine-grained category labels, label descriptions and object regions can help MLLMs better \textbf{learn} and \textbf{generate} concepts.
Besides, the noise introduced by \icdatasetname{} (discussed in \Sref{sec:experiment:scalable}) in the pre-training stage may not be well alleviated by SFT, thus \methodnamewithic{} achieves the second-best performance.
More results and analyses on image editing are provided in \Apref{sec:appendix:image_editing}.

\begin{figure}[t]
	\centering
    \includegraphics[width=\textwidth]{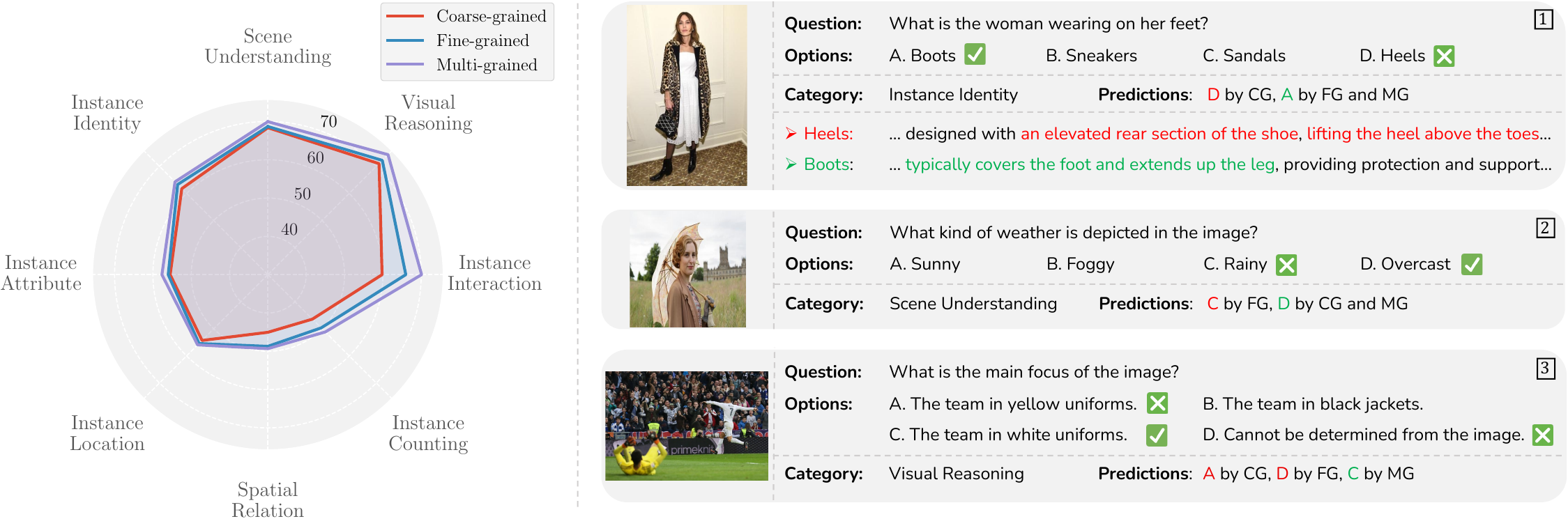}
	\caption{
        Analysis on $8$ dimensions of SEED-Bench-IMG.
        Left: the performance of \methodname{} trained with different-grained concept annotations from \datasetname{}.
        Right: corresponding case studies. 
		CG, FG, and MG denote MLLMs trained with coarse-, fine-, and multi-grained concept annotations from \datasetname{}, respectively.
        ✅ denote the ground truth; ❎ denote incorrect prediction(s).\looseness=-1
	}
	\label{fig:meso_seed_all}
\end{figure}

\subsection{The Impact of Different-Grained Concept Annotations}
\label{sec:experiment:meso}

To delve into the impact of different combination of multi-grained concept annotations in \datasetname{} for MLLMs,
we conduct meso analyses on different capability dimensions of SEED-Bench~\citep{li2023seed}, a large-scale multi-choice benchmark for multimodal comprehension.
For briefly, ``Coarse-grained'' (CG) means that only image captions from \datasetname{} are used.
``Fine-grained'' (FG) uses category labels, label descriptions and object regions from \datasetname{}.
``Multi-grained'' (MG) means that both above are used.
We follow common practice~\citep{ge2023making} to select $9$ image evaluation dimensions, \ie{} about $14$K questions, and denote it as SEED-Bench-IMG.\footnote{We ignore the ``\textit{Text Recognition}'' dimension as its noisy data can lead to result fluctuations.}
``\textit{Scene Understanding, Visual Reasoning}'' dimensions focus on the holistic understanding and cross-modal reasoning of the image, while the other $6$ dimensions focus on in-depth understanding of concrete concepts in the image.\looseness=-1

\paragraph{Quantitative Analysis.}
We first explore the impact of different-grained concept annotations from \datasetname{} on each evaluation dimension of SEED-Bench-IMG in \Fref{fig:meso_seed_all} \figleft{}.
For overall accuracy, FG significantly improves performance over CG by $1.39$ points, and multi-grained can further improve by $1.4$ points.
For each evaluation dimension:
\begin{itemize}[leftmargin=2em, topsep=0em, itemsep=0em]
    \item FG provides \textbf{deeper} understanding of concepts than CG.
    Compared with coarse-grained image captions, fine-grained concept annotations can help MLLMs better \textbf{understand} and \textbf{locate} concepts in images, and \textbf{recognize} relationships between concepts, especially on the ``\textit{Instance Identity, Spatial Relation, Instance Counting, Instance Interaction, Visual Reasoning}'' dimensions.\looseness=-1
    \item MG can facilitate \textbf{collaboration} between concept annotations of different granularities, thus fully \textbf{integrating} each other's strengths and achieving further improvements in all dimensions, especially on the ``\textit{Scene Understanding, Instance Attribute, Instance Counting, Instance Interaction, Visual Reasoning}'' dimensions.
    This demonstrates that our structured template for \datasetname{} can help MLLMs better utilize multi-grained concept annotations to learn concepts and thus promote vision--language alignment \textbf{across} multiple granularities \textbf{simultaneously}.\looseness=-1
\end{itemize}

\paragraph{Qualitative Analysis.}
To better analyse the advantages of different-grained concept annotations in \datasetname{} for MLLMs, we provide corresponding qualitative analysis in \Fref{fig:meso_seed_all} \figright{}.
More case studies are provided in \Apref{sec:appendix:seed_case_study}.
\begin{itemize}[leftmargin=2em, topsep=0em, itemsep=0.07em]
	\item Example {\scriptsize $\boxed{1}$} ``Instance Identity'': 
    FG provides \textbf{deeper} understanding of concepts than CG.
	While CG provides a holistic description of the image, it cannot help MLLMs distinguish between ``Heels'' and ``Boots''.
	However, FG provides visual details ``lifting the heel above the toes'' and ``covers the foot and extends up the leg'' by label--description pairs, which help MLLMs distinguish between similar concepts.
	Object region-annotation pairs further help MLLMs better locate and learn these concepts.
	Hence, FG helps \textbf{capture} visual details and \textbf{identify} concepts correctly.\looseness=-1
	\item Example {\scriptsize $\boxed{2}$} ``Scene Understanding'':
    CG provides more \textbf{holistic} understanding of the image than FG.
	While the lady in the image is holding an ``umbrella'', her surroundings show that the sky is covered with clouds, and it is not raining.
	MLLMs trained with FG seem to be too focused on the ``umbrella'' and ignore the overall scene of the image, while CG can help MLLMs predict the correct answer by better understanding the \textbf{global context}.
	\item Example {\scriptsize $\boxed{3}$} ``Visual Reasoning'':
    MG can \textbf{combine} the advantages of CG and FG to improve visual reasoning.
	This image shows a football match with a player in ``yellow uniforms'' lying on the ground and a player in ``white uniforms'' celebrating as the audience cheers him on.
	Rich objects and details confuse MLLMs trained with FG to confirm the main focus of the image, while MLLMs trained with CG perceive the visually more prominent ``yellow'' player as the main focus.
	However, with our structured template and MLLM framework, MG enables different-grained concept annotations to \textbf{integrate} and \textbf{complement} each other, thus better understanding and reasoning about the image from both \textbf{global context} and \textbf{local details}.
	MLLMs trained with MG correctly identify the ``white'' player as the main focus of the image.
\end{itemize}

In addition, we not only provide qualitative analysis in~\Fref{fig:case_study_1},~\ref{fig:case_study_2},~\ref{fig:meso_seed_all}, but also more exploratory experiments and analysis in \Apref{sec:appendix:experiment}, including concept overlap analysis, results and analysis on image editing tasks, self-generated multi-grained concept annotations in evaluation, the effectiveness of directly using \datasetname{} as SFT data, the impact of the nature of image--text interleaved, etc.
\section{Related Work}
\label{sec:related}
MLLMs have emerged recently including ones~\citep{sun2023generative,ge2023making,jin2023unified,zhu2023vl,dong2024dreamllm} that propose multimodal generalists capable of both multimodal comprehension and generation.
However, existing MLLMs often overlook the importance of concepts and only utilize coarse-grained concept annotations, \eg{} image captions, which may limit vision--language alignment.
Factually, many traditional VLMs recognized the importance of concepts in vision--language learning.
To better utilize concepts in vision--language learning and improve performance, they incorporated fine-grained concept annotations into coarse-grained image captions.
For example, 
object labels~\citep{li2020oscar,zhang2021vinvl}, label descriptions~\citep{shen2022k,yao2022detclip,menon2023visual,li2024desco}, region descriptions~\citep{zeng2021multi} and object regions~\citep{zeng2021multi,li2022grounded}.
Especially, Oscar~\citep{li2020oscar} appends fine-grained object labels detected in the image after the coarse-grained image caption to simplify semantic alignment between vision and language.
X-VLM~\citep{zeng2021multi} proposes to align visual concepts (images and object regions) with coarse-grained image captions and fine-grained object labels and object region descriptions in multi-granularity.
In this paper, different from previous VLM work, we collect and construct rich multimodal multi-grained concept annotations in \datasetname{} dataset.
Without additional components or loss functions, we design a structured template to leverage the advantages of MLLMs under an autoregressive discrete framework.
Through evaluation of $12$ multimodal comprehension and generation benchmarks, as well as the comparison, combination, and analysis of \datasetname{} and image--caption data, for the first time, we explore and demonstrate the potential of \datasetname{} in MLLMs.\looseness=-1
\section{Conclusion and Future Work}
\label{sec:conclusion}

To explore the potential of multi-grained concept annotations for MLLMs, we introduce a new multimodal dataset, \datasetname{}, and a general MLLM framework, which can be the foundation for the general and comprehensive exploration.
With our structured template and autoregressive discrete framework, multi-grained concept annotations can integrate and complement each other to help MLLMs better locate and learn concepts, thereby aligning vision and language across multiple granularities simultaneously.
Furthermore, \datasetname{} and coarse-grained image--caption data each have their own strengths in depth and breadth of concept representation, and appropriately combining them can effectively integrate their strengths to further improve performance.
We hope to inspire future research to further explore the potential of multi-grained concept annotations by incorporating more different types of annotations, scaling up data by automatic synthesis, scaling up (concrete and even abstract) concepts, and other VL tasks.\looseness=-1

\section*{Acknowledgements}
We gratefully acknowledge the support of the National Natural Science Foundation of China (NSFC) via grant 62236004, 62441603 and 62476073.

\bibliography{reference}

\begin{thebibliography}{135}
\providecommand{\natexlab}[1]{#1}
\providecommand{\url}[1]{\texttt{#1}}
\expandafter\ifx\csname urlstyle\endcsname\relax
  \providecommand{\doi}[1]{doi: #1}\else
  \providecommand{\doi}{doi: \begingroup \urlstyle{rm}\Url}\fi

\bibitem[Achiam et~al.(2023)Achiam, Adler, Agarwal, Ahmad, Akkaya, Aleman, Almeida, Altenschmidt, Altman, Anadkat, et~al.]{achiam2023gpt}
Josh Achiam, Steven Adler, Sandhini Agarwal, Lama Ahmad, Ilge Akkaya, Florencia~Leoni Aleman, Diogo Almeida, Janko Altenschmidt, Sam Altman, Shyamal Anadkat, et~al.
\newblock Gpt-4 technical report.
\newblock \emph{ArXiv preprint}, abs/2303.08774, 2023.
\newblock URL \url{https://arxiv.org/abs/2303.08774}.

\bibitem[Agrawal et~al.(2019)Agrawal, Anderson, Desai, Wang, Chen, Jain, Johnson, Batra, Parikh, and Lee]{agrawal2019nocaps}
Harsh Agrawal, Peter Anderson, Karan Desai, Yufei Wang, Xinlei Chen, Rishabh Jain, Mark Johnson, Dhruv Batra, Devi Parikh, and Stefan Lee.
\newblock nocaps: novel object captioning at scale.
\newblock In \emph{2019 {IEEE/CVF} International Conference on Computer Vision, {ICCV} 2019, Seoul, Korea (South), October 27 - November 2, 2019}, pp.\  8947--8956. {IEEE}, 2019.
\newblock \doi{10.1109/ICCV.2019.00904}.
\newblock URL \url{https://doi.org/10.1109/ICCV.2019.00904}.

\bibitem[Alayrac et~al.(2022)Alayrac, Donahue, Luc, Miech, Barr, Hasson, Lenc, Mensch, Millican, Reynolds, Ring, Rutherford, Cabi, Han, Gong, Samangooei, Monteiro, Menick, Borgeaud, Brock, Nematzadeh, Sharifzadeh, Binkowski, Barreira, Vinyals, Zisserman, and Simonyan]{alayrac2022flamingo}
Jean{-}Baptiste Alayrac, Jeff Donahue, Pauline Luc, Antoine Miech, Iain Barr, Yana Hasson, Karel Lenc, Arthur Mensch, Katherine Millican, Malcolm Reynolds, Roman Ring, Eliza Rutherford, Serkan Cabi, Tengda Han, Zhitao Gong, Sina Samangooei, Marianne Monteiro, Jacob~L. Menick, Sebastian Borgeaud, Andy Brock, Aida Nematzadeh, Sahand Sharifzadeh, Mikolaj Binkowski, Ricardo Barreira, Oriol Vinyals, Andrew Zisserman, and Kar{\'{e}}n Simonyan.
\newblock Flamingo: a visual language model for few-shot learning.
\newblock In Sanmi Koyejo, S.~Mohamed, A.~Agarwal, Danielle Belgrave, K.~Cho, and A.~Oh (eds.), \emph{Advances in Neural Information Processing Systems 35: Annual Conference on Neural Information Processing Systems 2022, NeurIPS 2022, New Orleans, LA, USA, November 28 - December 9, 2022}, 2022.
\newblock URL \url{http://papers.nips.cc/paper\_files/paper/2022/hash/960a172bc7fbf0177ccccbb411a7d800-Abstract-Conference.html}.

\bibitem[Awadalla et~al.(2024)Awadalla, Xue, Lo, Shu, Lee, Guha, Jordan, Shen, Awadalla, Savarese, et~al.]{awadalla2024mint}
Anas Awadalla, Le~Xue, Oscar Lo, Manli Shu, Hannah Lee, Etash~Kumar Guha, Matt Jordan, Sheng Shen, Mohamed Awadalla, Silvio Savarese, et~al.
\newblock Mint-1t: Scaling open-source multimodal data by 10x: A multimodal dataset with one trillion tokens.
\newblock \emph{ArXiv preprint}, abs/2406.11271, 2024.
\newblock URL \url{https://arxiv.org/abs/2406.11271}.

\bibitem[Blouw et~al.(2016)Blouw, Solodkin, Thagard, and Eliasmith]{blouw2016concepts}
Peter Blouw, Eugene Solodkin, Paul Thagard, and Chris Eliasmith.
\newblock Concepts as semantic pointers: A framework and computational model.
\newblock \emph{Cognitive science}, 40\penalty0 (5):\penalty0 1128--1162, 2016.

\bibitem[Brooks et~al.(2023)Brooks, Holynski, and Efros]{brooks2023instructpix2pix}
Tim Brooks, Aleksander Holynski, and Alexei~A. Efros.
\newblock Instructpix2pix: Learning to follow image editing instructions.
\newblock In \emph{{IEEE/CVF} Conference on Computer Vision and Pattern Recognition, {CVPR} 2023, Vancouver, BC, Canada, June 17-24, 2023}, pp.\  18392--18402. {IEEE}, 2023.
\newblock \doi{10.1109/CVPR52729.2023.01764}.
\newblock URL \url{https://doi.org/10.1109/CVPR52729.2023.01764}.

\bibitem[Brown et~al.(2020)Brown, Mann, Ryder, Subbiah, Kaplan, Dhariwal, Neelakantan, Shyam, Sastry, Askell, Agarwal, Herbert{-}Voss, Krueger, Henighan, Child, Ramesh, Ziegler, Wu, Winter, Hesse, Chen, Sigler, Litwin, Gray, Chess, Clark, Berner, McCandlish, Radford, Sutskever, and Amodei]{brown2020language}
Tom~B. Brown, Benjamin Mann, Nick Ryder, Melanie Subbiah, Jared Kaplan, Prafulla Dhariwal, Arvind Neelakantan, Pranav Shyam, Girish Sastry, Amanda Askell, Sandhini Agarwal, Ariel Herbert{-}Voss, Gretchen Krueger, Tom Henighan, Rewon Child, Aditya Ramesh, Daniel~M. Ziegler, Jeffrey Wu, Clemens Winter, Christopher Hesse, Mark Chen, Eric Sigler, Mateusz Litwin, Scott Gray, Benjamin Chess, Jack Clark, Christopher Berner, Sam McCandlish, Alec Radford, Ilya Sutskever, and Dario Amodei.
\newblock Language models are few-shot learners.
\newblock In Hugo Larochelle, Marc'Aurelio Ranzato, Raia Hadsell, Maria{-}Florina Balcan, and Hsuan{-}Tien Lin (eds.), \emph{Advances in Neural Information Processing Systems 33: Annual Conference on Neural Information Processing Systems 2020, NeurIPS 2020, December 6-12, 2020, virtual}, 2020.
\newblock URL \url{https://proceedings.neurips.cc/paper/2020/hash/1457c0d6bfcb4967418bfb8ac142f64a-Abstract.html}.

\bibitem[Cai et~al.(2022)Cai, Zhang, Zhu, Zhang, Li, and Xue]{cai2022bigdetection}
Likun Cai, Zhi Zhang, Yi~Zhu, Li~Zhang, Mu~Li, and Xiangyang Xue.
\newblock Bigdetection: {A} large-scale benchmark for improved object detector pre-training.
\newblock In \emph{{IEEE/CVF} Conference on Computer Vision and Pattern Recognition Workshops, {CVPR} Workshops 2022, New Orleans, LA, USA, June 19-20, 2022}, pp.\  4776--4786. {IEEE}, 2022.
\newblock \doi{10.1109/CVPRW56347.2022.00524}.
\newblock URL \url{https://doi.org/10.1109/CVPRW56347.2022.00524}.

\bibitem[Carey(2000)]{carey2000origin}
Susan Carey.
\newblock The origin of concepts.
\newblock \emph{Journal of Cognition and Development}, 1\penalty0 (1):\penalty0 37--41, 2000.

\bibitem[Caron et~al.(2021)Caron, Touvron, Misra, J{\'{e}}gou, Mairal, Bojanowski, and Joulin]{caron2021emerging}
Mathilde Caron, Hugo Touvron, Ishan Misra, Herv{\'{e}} J{\'{e}}gou, Julien Mairal, Piotr Bojanowski, and Armand Joulin.
\newblock Emerging properties in self-supervised vision transformers.
\newblock In \emph{2021 {IEEE/CVF} International Conference on Computer Vision, {ICCV} 2021, Montreal, QC, Canada, October 10-17, 2021}, pp.\  9630--9640. {IEEE}, 2021.
\newblock \doi{10.1109/ICCV48922.2021.00951}.
\newblock URL \url{https://doi.org/10.1109/ICCV48922.2021.00951}.

\bibitem[Changpinyo et~al.(2021)Changpinyo, Sharma, Ding, and Soricut]{changpinyo2021conceptual}
Soravit Changpinyo, Piyush Sharma, Nan Ding, and Radu Soricut.
\newblock Conceptual 12m: Pushing web-scale image-text pre-training to recognize long-tail visual concepts.
\newblock In \emph{{IEEE} Conference on Computer Vision and Pattern Recognition, {CVPR} 2021, virtual, June 19-25, 2021}, pp.\  3558--3568. Computer Vision Foundation / {IEEE}, 2021.
\newblock \doi{10.1109/CVPR46437.2021.00356}.
\newblock URL \url{https://openaccess.thecvf.com/content/CVPR2021/html/Changpinyo\_Conceptual\_12M\_Pushing\_Web-Scale\_Image-Text\_Pre-Training\_To\_Recognize\_Long-Tail\_Visual\_CVPR\_2021\_paper.html}.

\bibitem[Chen et~al.(2019)Chen, Ding, Lin, Zhao, and Han]{chen2019cross}
Hui Chen, Guiguang Ding, Zijia Lin, Sicheng Zhao, and Jungong Han.
\newblock Cross-modal image-text retrieval with semantic consistency.
\newblock In \emph{Proceedings of the 27th {ACM} International Conference on Multimedia, {MM} 2019, Nice, France, October 21-25, 2019}, pp.\  1749--1757, 2019.
\newblock \doi{10.1145/3343031.3351055}.
\newblock URL \url{https://doi.org/10.1145/3343031.3351055}.

\bibitem[Chen et~al.(2023)Chen, Li, Dong, Zhang, He, Wang, Zhao, and Lin]{chen2023sharegpt4v}
Lin Chen, Jisong Li, Xiaoyi Dong, Pan Zhang, Conghui He, Jiaqi Wang, Feng Zhao, and Dahua Lin.
\newblock Sharegpt4v: Improving large multimodal models with better captions.
\newblock \emph{ArXiv preprint}, abs/2311.12793, 2023.
\newblock URL \url{https://arxiv.org/abs/2311.12793}.

\bibitem[Chen et~al.(2024)Chen, Qin, Zhang, Chen, Xu, and Che]{chen-etal-2024-m3cot}
Qiguang Chen, Libo Qin, Jin Zhang, Zhi Chen, Xiao Xu, and Wanxiang Che.
\newblock {M}$^3${C}o{T}: A novel benchmark for multi-domain multi-step multi-modal chain-of-thought.
\newblock In Lun-Wei Ku, Andre Martins, and Vivek Srikumar (eds.), \emph{Proceedings of the 62nd Annual Meeting of the Association for Computational Linguistics (Volume 1: Long Papers)}, pp.\  8199--8221, Bangkok, Thailand, 2024. Association for Computational Linguistics.
\newblock \doi{10.18653/v1/2024.acl-long.446}.
\newblock URL \url{https://aclanthology.org/2024.acl-long.446}.

\bibitem[Chen et~al.(2022)Chen, Saxena, Li, Fleet, and Hinton]{chen2021pix2seq}
Ting Chen, Saurabh Saxena, Lala Li, David~J. Fleet, and Geoffrey~E. Hinton.
\newblock Pix2seq: {A} language modeling framework for object detection.
\newblock In \emph{The Tenth International Conference on Learning Representations, {ICLR} 2022, Virtual Event, April 25-29, 2022}. OpenReview.net, 2022.
\newblock URL \url{https://openreview.net/forum?id=e42KbIw6Wb}.

\bibitem[Chen et~al.(2015)Chen, Fang, Lin, Vedantam, Gupta, Doll{\'a}r, and Zitnick]{chen2015microsoft}
Xinlei Chen, Hao Fang, Tsung-Yi Lin, Ramakrishna Vedantam, Saurabh Gupta, Piotr Doll{\'a}r, and C~Lawrence Zitnick.
\newblock Microsoft coco captions: Data collection and evaluation server.
\newblock \emph{ArXiv preprint}, abs/1504.00325, 2015.
\newblock URL \url{https://arxiv.org/abs/1504.00325}.

\bibitem[Chowdhery et~al.(2023)Chowdhery, Narang, Devlin, Bosma, Mishra, Roberts, Barham, Chung, Sutton, Gehrmann, et~al.]{chowdhery2023palm}
Aakanksha Chowdhery, Sharan Narang, Jacob Devlin, Maarten Bosma, Gaurav Mishra, Adam Roberts, Paul Barham, Hyung~Won Chung, Charles Sutton, Sebastian Gehrmann, et~al.
\newblock Palm: Scaling language modeling with pathways.
\newblock \emph{Journal of Machine Learning Research}, 24\penalty0 (240):\penalty0 1--113, 2023.

\bibitem[Connell et~al.(2018)Connell, Lynott, and Banks]{connell2018interoception}
Louise Connell, Dermot Lynott, and Briony Banks.
\newblock Interoception: the forgotten modality in perceptual grounding of abstract and concrete concepts.
\newblock \emph{Philosophical Transactions of the Royal Society B: Biological Sciences}, 373\penalty0 (1752):\penalty0 20170143, 2018.

\bibitem[Dai et~al.(2024)Dai, Lee, Wang, Yang, Liu, Barker, Rintamaki, Shoeybi, Catanzaro, and Ping]{dai2024nvlm}
Wenliang Dai, Nayeon Lee, Boxin Wang, Zhuoling Yang, Zihan Liu, Jon Barker, Tuomas Rintamaki, Mohammad Shoeybi, Bryan Catanzaro, and Wei Ping.
\newblock Nvlm: Open frontier-class multimodal llms.
\newblock \emph{ArXiv preprint}, abs/2409.11402, 2024.
\newblock URL \url{https://arxiv.org/abs/2409.11402}.

\bibitem[Dehghani et~al.(2023)Dehghani, Djolonga, Mustafa, Padlewski, Heek, Gilmer, Steiner, Caron, Geirhos, Alabdulmohsin, Jenatton, Beyer, Tschannen, Arnab, Wang, Ruiz, Minderer, Puigcerver, Evci, Kumar, van Steenkiste, Elsayed, Mahendran, Yu, Oliver, Huot, Bastings, Collier, Gritsenko, Birodkar, Vasconcelos, Tay, Mensink, Kolesnikov, Pavetic, Tran, Kipf, Lucic, Zhai, Keysers, Harmsen, and Houlsby]{dehghani2023scaling}
Mostafa Dehghani, Josip Djolonga, Basil Mustafa, Piotr Padlewski, Jonathan Heek, Justin Gilmer, Andreas~Peter Steiner, Mathilde Caron, Robert Geirhos, Ibrahim Alabdulmohsin, Rodolphe Jenatton, Lucas Beyer, Michael Tschannen, Anurag Arnab, Xiao Wang, Carlos~Riquelme Ruiz, Matthias Minderer, Joan Puigcerver, Utku Evci, Manoj Kumar, Sjoerd van Steenkiste, Gamaleldin~Fathy Elsayed, Aravindh Mahendran, Fisher Yu, Avital Oliver, Fantine Huot, Jasmijn Bastings, Mark Collier, Alexey~A. Gritsenko, Vighnesh Birodkar, Cristina~Nader Vasconcelos, Yi~Tay, Thomas Mensink, Alexander Kolesnikov, Filip Pavetic, Dustin Tran, Thomas Kipf, Mario Lucic, Xiaohua Zhai, Daniel Keysers, Jeremiah~J. Harmsen, and Neil Houlsby.
\newblock Scaling vision transformers to 22 billion parameters.
\newblock In Andreas Krause, Emma Brunskill, Kyunghyun Cho, Barbara Engelhardt, Sivan Sabato, and Jonathan Scarlett (eds.), \emph{International Conference on Machine Learning, {ICML} 2023, 23-29 July 2023, Honolulu, Hawaii, {USA}}, volume 202 of \emph{Proceedings of Machine Learning Research}, pp.\  7480--7512. {PMLR}, 2023.
\newblock URL \url{https://proceedings.mlr.press/v202/dehghani23a.html}.

\bibitem[Dong et~al.(2024)Dong, Han, Peng, Qi, Ge, Yang, Zhao, Sun, Zhou, Wei, Kong, Zhang, Ma, and Yi]{dong2024dreamllm}
Runpei Dong, Chunrui Han, Yuang Peng, Zekun Qi, Zheng Ge, Jinrong Yang, Liang Zhao, Jianjian Sun, Hongyu Zhou, Haoran Wei, Xiangwen Kong, Xiangyu Zhang, Kaisheng Ma, and Li~Yi.
\newblock Dream{LLM}: Synergistic multimodal comprehension and creation.
\newblock In \emph{The Twelfth International Conference on Learning Representations}, 2024.
\newblock URL \url{https://openreview.net/forum?id=y01KGvd9Bw}.

\bibitem[Dosovitskiy et~al.(2021)Dosovitskiy, Beyer, Kolesnikov, Weissenborn, Zhai, Unterthiner, Dehghani, Minderer, Heigold, Gelly, Uszkoreit, and Houlsby]{dosovitskiy2020image}
Alexey Dosovitskiy, Lucas Beyer, Alexander Kolesnikov, Dirk Weissenborn, Xiaohua Zhai, Thomas Unterthiner, Mostafa Dehghani, Matthias Minderer, Georg Heigold, Sylvain Gelly, Jakob Uszkoreit, and Neil Houlsby.
\newblock An image is worth 16x16 words: Transformers for image recognition at scale.
\newblock In \emph{9th International Conference on Learning Representations, {ICLR} 2021, Virtual Event, Austria, May 3-7, 2021}. OpenReview.net, 2021.
\newblock URL \url{https://openreview.net/forum?id=YicbFdNTTy}.

\bibitem[Ge et~al.(2023{\natexlab{a}})Ge, Ge, Zeng, Wang, and Shan]{ge2023planting}
Yuying Ge, Yixiao Ge, Ziyun Zeng, Xintao Wang, and Ying Shan.
\newblock Planting a seed of vision in large language model.
\newblock \emph{ArXiv preprint}, abs/2307.08041, 2023{\natexlab{a}}.
\newblock URL \url{https://arxiv.org/abs/2307.08041}.

\bibitem[Ge et~al.(2023{\natexlab{b}})Ge, Zhao, Zeng, Ge, Li, Wang, and Shan]{ge2023making}
Yuying Ge, Sijie Zhao, Ziyun Zeng, Yixiao Ge, Chen Li, Xintao Wang, and Ying Shan.
\newblock Making llama see and draw with seed tokenizer.
\newblock \emph{ArXiv preprint}, abs/2310.01218, 2023{\natexlab{b}}.
\newblock URL \url{https://arxiv.org/abs/2310.01218}.

\bibitem[Goguen(1969)]{goguen1969logic}
Joseph~A Goguen.
\newblock The logic of inexact concepts.
\newblock \emph{Synthese}, 19\penalty0 (3/4):\penalty0 325--373, 1969.

\bibitem[Goyal et~al.(2017)Goyal, Khot, Summers{-}Stay, Batra, and Parikh]{balanced_vqa_v2}
Yash Goyal, Tejas Khot, Douglas Summers{-}Stay, Dhruv Batra, and Devi Parikh.
\newblock Making the {V} in {VQA} matter: Elevating the role of image understanding in visual question answering.
\newblock In \emph{2017 {IEEE} Conference on Computer Vision and Pattern Recognition, {CVPR} 2017, Honolulu, HI, USA, July 21-26, 2017}, pp.\  6325--6334. {IEEE} Computer Society, 2017.
\newblock \doi{10.1109/CVPR.2017.670}.
\newblock URL \url{https://doi.org/10.1109/CVPR.2017.670}.

\bibitem[Gupta et~al.(2019)Gupta, Doll{\'{a}}r, and Girshick]{gupta2019lvis}
Agrim Gupta, Piotr Doll{\'{a}}r, and Ross~B. Girshick.
\newblock {LVIS:} {A} dataset for large vocabulary instance segmentation.
\newblock In \emph{{IEEE} Conference on Computer Vision and Pattern Recognition, {CVPR} 2019, Long Beach, CA, USA, June 16-20, 2019}, pp.\  5356--5364. Computer Vision Foundation / {IEEE}, 2019.
\newblock \doi{10.1109/CVPR.2019.00550}.
\newblock URL \url{http://openaccess.thecvf.com/content\_CVPR\_2019/html/Gupta\_LVIS\_A\_Dataset\_for\_Large\_Vocabulary\_Instance\_Segmentation\_CVPR\_2019\_paper.html}.

\bibitem[Gurari et~al.(2018)Gurari, Li, Stangl, Guo, Lin, Grauman, Luo, and Bigham]{gurari2018vizwiz}
Danna Gurari, Qing Li, Abigale~J. Stangl, Anhong Guo, Chi Lin, Kristen Grauman, Jiebo Luo, and Jeffrey~P. Bigham.
\newblock Vizwiz grand challenge: Answering visual questions from blind people.
\newblock In \emph{2018 {IEEE} Conference on Computer Vision and Pattern Recognition, {CVPR} 2018, Salt Lake City, UT, USA, June 18-22, 2018}, pp.\  3608--3617. {IEEE} Computer Society, 2018.
\newblock \doi{10.1109/CVPR.2018.00380}.
\newblock URL \url{http://openaccess.thecvf.com/content\_cvpr\_2018/html/Gurari\_VizWiz\_Grand\_Challenge\_CVPR\_2018\_paper.html}.

\bibitem[Hendrycks et~al.(2021)Hendrycks, Burns, Basart, Zou, Mazeika, Song, and Steinhardt]{hendrycks2021measuring}
Dan Hendrycks, Collin Burns, Steven Basart, Andy Zou, Mantas Mazeika, Dawn Song, and Jacob Steinhardt.
\newblock Measuring massive multitask language understanding.
\newblock In \emph{9th International Conference on Learning Representations, {ICLR} 2021, Virtual Event, Austria, May 3-7, 2021}. OpenReview.net, 2021.
\newblock URL \url{https://openreview.net/forum?id=d7KBjmI3GmQ}.

\bibitem[Ho et~al.(2020)Ho, Jain, and Abbeel]{ho2020denoising}
Jonathan Ho, Ajay Jain, and Pieter Abbeel.
\newblock Denoising diffusion probabilistic models.
\newblock In Hugo Larochelle, Marc'Aurelio Ranzato, Raia Hadsell, Maria{-}Florina Balcan, and Hsuan{-}Tien Lin (eds.), \emph{Advances in Neural Information Processing Systems 33: Annual Conference on Neural Information Processing Systems 2020, NeurIPS 2020, December 6-12, 2020, virtual}, 2020.
\newblock URL \url{https://proceedings.neurips.cc/paper/2020/hash/4c5bcfec8584af0d967f1ab10179ca4b-Abstract.html}.

\bibitem[Hu et~al.(2022)Hu, Shen, Wallis, Allen{-}Zhu, Li, Wang, Wang, and Chen]{hu2021lora}
Edward~J. Hu, Yelong Shen, Phillip Wallis, Zeyuan Allen{-}Zhu, Yuanzhi Li, Shean Wang, Lu~Wang, and Weizhu Chen.
\newblock Lora: Low-rank adaptation of large language models.
\newblock In \emph{The Tenth International Conference on Learning Representations, {ICLR} 2022, Virtual Event, April 25-29, 2022}. OpenReview.net, 2022.
\newblock URL \url{https://openreview.net/forum?id=nZeVKeeFYf9}.

\bibitem[Hu et~al.(2024{\natexlab{a}})Hu, Yao, Wang, WANG, Pan, Chen, Yu, Wu, Zhao, Zhang, Han, Lin, Xue, dahai li, Liu, and Sun]{hu2024large}
Jinyi Hu, Yuan Yao, Chongyi Wang, SHAN WANG, Yinxu Pan, Qianyu Chen, Tianyu Yu, Hanghao Wu, Yue Zhao, Haoye Zhang, Xu~Han, Yankai Lin, Jiao Xue, dahai li, Zhiyuan Liu, and Maosong Sun.
\newblock Large multilingual models pivot zero-shot multimodal learning across languages.
\newblock In \emph{The Twelfth International Conference on Learning Representations}, 2024{\natexlab{a}}.
\newblock URL \url{https://openreview.net/forum?id=Kuh5qgCGCp}.

\bibitem[Hu et~al.(2024{\natexlab{b}})Hu, Tu, Han, He, Cui, Long, Zheng, Fang, Huang, Zhao, et~al.]{hu2024minicpm}
Shengding Hu, Yuge Tu, Xu~Han, Chaoqun He, Ganqu Cui, Xiang Long, Zhi Zheng, Yewei Fang, Yuxiang Huang, Weilin Zhao, et~al.
\newblock Minicpm: Unveiling the potential of small language models with scalable training strategies.
\newblock \emph{ArXiv preprint}, abs/2404.06395, 2024{\natexlab{b}}.
\newblock URL \url{https://arxiv.org/abs/2404.06395}.

\bibitem[Huang et~al.(2016)Huang, Ferraro, Mostafazadeh, Misra, Agrawal, Devlin, Girshick, He, Kohli, Batra, Zitnick, Parikh, Vanderwende, Galley, and Mitchell]{huang-etal-2016-visual}
Ting-Hao~Kenneth Huang, Francis Ferraro, Nasrin Mostafazadeh, Ishan Misra, Aishwarya Agrawal, Jacob Devlin, Ross Girshick, Xiaodong He, Pushmeet Kohli, Dhruv Batra, C.~Lawrence Zitnick, Devi Parikh, Lucy Vanderwende, Michel Galley, and Margaret Mitchell.
\newblock Visual storytelling.
\newblock In Kevin Knight, Ani Nenkova, and Owen Rambow (eds.), \emph{Proceedings of the 2016 Conference of the North {A}merican Chapter of the Association for Computational Linguistics: Human Language Technologies}, pp.\  1233--1239, San Diego, California, 2016. Association for Computational Linguistics.
\newblock \doi{10.18653/v1/N16-1147}.
\newblock URL \url{https://aclanthology.org/N16-1147}.

\bibitem[Hudson \& Manning(2019)Hudson and Manning]{hudson2019gqa}
Drew~A. Hudson and Christopher~D. Manning.
\newblock {GQA:} {A} new dataset for real-world visual reasoning and compositional question answering.
\newblock In \emph{{IEEE} Conference on Computer Vision and Pattern Recognition, {CVPR} 2019, Long Beach, CA, USA, June 16-20, 2019}, pp.\  6700--6709. Computer Vision Foundation / {IEEE}, 2019.
\newblock \doi{10.1109/CVPR.2019.00686}.
\newblock URL \url{http://openaccess.thecvf.com/content\_CVPR\_2019/html/Hudson\_GQA\_A\_New\_Dataset\_for\_Real-World\_Visual\_Reasoning\_and\_Compositional\_CVPR\_2019\_paper.html}.

\bibitem[Jin et~al.(2023)Jin, Xu, Chen, Liao, Tan, Chen, Lei, Liu, Song, Lei, et~al.]{jin2023unified}
Yang Jin, Kun Xu, Liwei Chen, Chao Liao, Jianchao Tan, Bin Chen, Chenyi Lei, An~Liu, Chengru Song, Xiaoqiang Lei, et~al.
\newblock Unified language-vision pretraining with dynamic discrete visual tokenization.
\newblock \emph{ArXiv preprint}, abs/2309.04669, 2023.
\newblock URL \url{https://arxiv.org/abs/2309.04669}.

\bibitem[Kirillov et~al.(2023)Kirillov, Mintun, Ravi, Mao, Rolland, Gustafson, Xiao, Whitehead, Berg, Lo, Doll{\'{a}}r, and Girshick]{kirillov2023segment}
Alexander Kirillov, Eric Mintun, Nikhila Ravi, Hanzi Mao, Chlo{\'{e}} Rolland, Laura Gustafson, Tete Xiao, Spencer Whitehead, Alexander~C. Berg, Wan{-}Yen Lo, Piotr Doll{\'{a}}r, and Ross~B. Girshick.
\newblock Segment anything.
\newblock In \emph{{IEEE/CVF} International Conference on Computer Vision, {ICCV} 2023, Paris, France, October 1-6, 2023}, pp.\  3992--4003. {IEEE}, 2023.
\newblock \doi{10.1109/ICCV51070.2023.00371}.
\newblock URL \url{https://doi.org/10.1109/ICCV51070.2023.00371}.

\bibitem[Koh et~al.(2023)Koh, Fried, and Salakhutdinov]{koh2024generating}
Jing~Yu Koh, Daniel Fried, and Russ Salakhutdinov.
\newblock Generating images with multimodal language models.
\newblock In Alice Oh, Tristan Naumann, Amir Globerson, Kate Saenko, Moritz Hardt, and Sergey Levine (eds.), \emph{Advances in Neural Information Processing Systems 36: Annual Conference on Neural Information Processing Systems 2023, NeurIPS 2023, New Orleans, LA, USA, December 10 - 16, 2023}, 2023.
\newblock URL \url{http://papers.nips.cc/paper\_files/paper/2023/hash/43a69d143273bd8215578bde887bb552-Abstract-Conference.html}.

\bibitem[Krishna et~al.(2017)Krishna, Zhu, Groth, Johnson, Hata, Kravitz, Chen, Kalantidis, Li, Shamma, et~al.]{krishna2017visual}
Ranjay Krishna, Yuke Zhu, Oliver Groth, Justin Johnson, Kenji Hata, Joshua Kravitz, Stephanie Chen, Yannis Kalantidis, Li-Jia Li, David~A Shamma, et~al.
\newblock Visual genome: Connecting language and vision using crowdsourced dense image annotations.
\newblock \emph{International journal of computer vision}, 123\penalty0 (1):\penalty0 32--73, 2017.

\bibitem[Kuznetsova et~al.(2020)Kuznetsova, Rom, Alldrin, Uijlings, Krasin, Pont-Tuset, Kamali, Popov, Malloci, Kolesnikov, Duerig, and Ferrari]{OpenImages}
Alina Kuznetsova, Hassan Rom, Neil Alldrin, Jasper Uijlings, Ivan Krasin, Jordi Pont-Tuset, Shahab Kamali, Stefan Popov, Matteo Malloci, Alexander Kolesnikov, Tom Duerig, and Vittorio Ferrari.
\newblock The open images dataset v4: Unified image classification, object detection, and visual relationship detection at scale.
\newblock \emph{IJCV}, 2020.

\bibitem[LAION(2024{\natexlab{a}})]{laioncocoaesthetic}
LAION.
\newblock laion-coco-aesthetic.
\newblock \url{https://huggingface.co/datasets/guangyil/laion-coco-aesthetic}, 2024{\natexlab{a}}.

\bibitem[LAION(2024{\natexlab{b}})]{laiongpt4v}
GPT-4V LAION.
\newblock Laion-gpt4v.
\newblock \url{https://huggingface.co/datasets/laion/gpt4v-dataset}, 2024{\natexlab{b}}.

\bibitem[Lan et~al.(2020)Lan, Chen, Goodman, Gimpel, Sharma, and Soricut]{lan2020albert}
Zhenzhong Lan, Mingda Chen, Sebastian Goodman, Kevin Gimpel, Piyush Sharma, and Radu Soricut.
\newblock {ALBERT:} {A} lite {BERT} for self-supervised learning of language representations.
\newblock In \emph{8th International Conference on Learning Representations, {ICLR} 2020, Addis Ababa, Ethiopia, April 26-30, 2020}. OpenReview.net, 2020.
\newblock URL \url{https://openreview.net/forum?id=H1eA7AEtvS}.

\bibitem[Lauren{\c{c}}on et~al.(2023)Lauren{\c{c}}on, Saulnier, Tronchon, Bekman, Singh, Lozhkov, Wang, Karamcheti, Rush, Kiela, Cord, and Sanh]{laurenccon2024obelics}
Hugo Lauren{\c{c}}on, Lucile Saulnier, L{\'{e}}o Tronchon, Stas Bekman, Amanpreet Singh, Anton Lozhkov, Thomas Wang, Siddharth Karamcheti, Alexander~M. Rush, Douwe Kiela, Matthieu Cord, and Victor Sanh.
\newblock {OBELICS:} an open web-scale filtered dataset of interleaved image-text documents.
\newblock In Alice Oh, Tristan Naumann, Amir Globerson, Kate Saenko, Moritz Hardt, and Sergey Levine (eds.), \emph{Advances in Neural Information Processing Systems 36: Annual Conference on Neural Information Processing Systems 2023, NeurIPS 2023, New Orleans, LA, USA, December 10 - 16, 2023}, 2023.
\newblock URL \url{http://papers.nips.cc/paper\_files/paper/2023/hash/e2cfb719f58585f779d0a4f9f07bd618-Abstract-Datasets\_and\_Benchmarks.html}.

\bibitem[Li et~al.(2024{\natexlab{a}})Li, Zhang, Zhang, Guo, Zhang, Zhang, Li, Liu, and Li]{li2024llavanext-ablations}
Bo~Li, Hao Zhang, Kaichen Zhang, Dong Guo, Yuanhan Zhang, Renrui Zhang, Feng Li, Ziwei Liu, and Chunyuan Li.
\newblock Llava-next: What else influences visual instruction tuning beyond data?, 2024{\natexlab{a}}.
\newblock URL \url{https://llava-vl.github.io/blog/2024-05-25-llava-next-ablations/}.

\bibitem[Li et~al.(2023{\natexlab{a}})Li, Wang, Wang, Ge, Ge, and Shan]{li2023seed}
Bohao Li, Rui Wang, Guangzhi Wang, Yuying Ge, Yixiao Ge, and Ying Shan.
\newblock Seed-bench: Benchmarking multimodal llms with generative comprehension.
\newblock \emph{ArXiv preprint}, abs/2307.16125, 2023{\natexlab{a}}.
\newblock URL \url{https://arxiv.org/abs/2307.16125}.

\bibitem[Li et~al.(2022{\natexlab{a}})Li, Li, Xiong, and Hoi]{li2022blip}
Junnan Li, Dongxu Li, Caiming Xiong, and Steven C.~H. Hoi.
\newblock {BLIP:} bootstrapping language-image pre-training for unified vision-language understanding and generation.
\newblock In Kamalika Chaudhuri, Stefanie Jegelka, Le~Song, Csaba Szepesv{\'{a}}ri, Gang Niu, and Sivan Sabato (eds.), \emph{International Conference on Machine Learning, {ICML} 2022, 17-23 July 2022, Baltimore, Maryland, {USA}}, volume 162 of \emph{Proceedings of Machine Learning Research}, pp.\  12888--12900. {PMLR}, 2022{\natexlab{a}}.
\newblock URL \url{https://proceedings.mlr.press/v162/li22n.html}.

\bibitem[Li et~al.(2023{\natexlab{b}})Li, Li, Savarese, and Hoi]{li2023blip}
Junnan Li, Dongxu Li, Silvio Savarese, and Steven C.~H. Hoi.
\newblock {BLIP-2:} bootstrapping language-image pre-training with frozen image encoders and large language models.
\newblock In Andreas Krause, Emma Brunskill, Kyunghyun Cho, Barbara Engelhardt, Sivan Sabato, and Jonathan Scarlett (eds.), \emph{International Conference on Machine Learning, {ICML} 2023, 23-29 July 2023, Honolulu, Hawaii, {USA}}, volume 202 of \emph{Proceedings of Machine Learning Research}, pp.\  19730--19742. {PMLR}, 2023{\natexlab{b}}.
\newblock URL \url{https://proceedings.mlr.press/v202/li23q.html}.

\bibitem[Li et~al.(2022{\natexlab{b}})Li, Zhang, Zhang, Yang, Li, Zhong, Wang, Yuan, Zhang, Hwang, Chang, and Gao]{li2022grounded}
Liunian~Harold Li, Pengchuan Zhang, Haotian Zhang, Jianwei Yang, Chunyuan Li, Yiwu Zhong, Lijuan Wang, Lu~Yuan, Lei Zhang, Jenq{-}Neng Hwang, Kai{-}Wei Chang, and Jianfeng Gao.
\newblock Grounded language-image pre-training.
\newblock In \emph{{IEEE/CVF} Conference on Computer Vision and Pattern Recognition, {CVPR} 2022, New Orleans, LA, USA, June 18-24, 2022}, pp.\  10955--10965. {IEEE}, 2022{\natexlab{b}}.
\newblock \doi{10.1109/CVPR52688.2022.01069}.
\newblock URL \url{https://doi.org/10.1109/CVPR52688.2022.01069}.

\bibitem[Li et~al.(2023{\natexlab{c}})Li, Dou, Peng, and Chang]{li2024desco}
Liunian~Harold Li, Zi{-}Yi Dou, Nanyun Peng, and Kai{-}Wei Chang.
\newblock Desco: Learning object recognition with rich language descriptions.
\newblock In Alice Oh, Tristan Naumann, Amir Globerson, Kate Saenko, Moritz Hardt, and Sergey Levine (eds.), \emph{Advances in Neural Information Processing Systems 36: Annual Conference on Neural Information Processing Systems 2023, NeurIPS 2023, New Orleans, LA, USA, December 10 - 16, 2023}, 2023{\natexlab{c}}.
\newblock URL \url{http://papers.nips.cc/paper\_files/paper/2023/hash/761c3284ee4859bff3c7e5d9299a45ee-Abstract-Conference.html}.

\bibitem[Li et~al.(2024{\natexlab{b}})Li, Zhang, Diao, Wang, Wang, and Duan]{li2024densefusion}
Xiaotong Li, Fan Zhang, Haiwen Diao, Yueze Wang, Xinlong Wang, and Ling-Yu Duan.
\newblock Densefusion-1m: Merging vision experts for comprehensive multimodal perception.
\newblock \emph{ArXiv preprint}, abs/2407.08303, 2024{\natexlab{b}}.
\newblock URL \url{https://arxiv.org/abs/2407.08303}.

\bibitem[Li et~al.(2020)Li, Yin, Li, Zhang, Hu, Zhang, Wang, Hu, Dong, Wei, et~al.]{li2020oscar}
Xiujun Li, Xi~Yin, Chunyuan Li, Pengchuan Zhang, Xiaowei Hu, Lei Zhang, Lijuan Wang, Houdong Hu, Li~Dong, Furu Wei, et~al.
\newblock Oscar: Object-semantics aligned pre-training for vision--language tasks.
\newblock In \emph{European Conference on Computer Vision}, pp.\  121--137. Springer, 2020.

\bibitem[Li et~al.(2023{\natexlab{d}})Li, Du, Zhou, Wang, Zhao, and Wen]{li-etal-2023-evaluating}
Yifan Li, Yifan Du, Kun Zhou, Jinpeng Wang, Xin Zhao, and Ji-Rong Wen.
\newblock Evaluating object hallucination in large vision-language models.
\newblock In Houda Bouamor, Juan Pino, and Kalika Bali (eds.), \emph{Proceedings of the 2023 Conference on Empirical Methods in Natural Language Processing}, pp.\  292--305, Singapore, 2023{\natexlab{d}}. Association for Computational Linguistics.
\newblock \doi{10.18653/v1/2023.emnlp-main.20}.
\newblock URL \url{https://aclanthology.org/2023.emnlp-main.20}.

\bibitem[Li et~al.(2024{\natexlab{c}})Li, Yang, Liu, Ma, Zhang, Yang, Sun, Liu, and Bai]{li2024monkey}
Zhang Li, Biao Yang, Qiang Liu, Zhiyin Ma, Shuo Zhang, Jingxu Yang, Yabo Sun, Yuliang Liu, and Xiang Bai.
\newblock Monkey: Image resolution and text label are important things for large multi-modal models.
\newblock In \emph{Proceedings of the IEEE/CVF Conference on Computer Vision and Pattern Recognition}, pp.\  26763--26773, 2024{\natexlab{c}}.

\bibitem[Lin et~al.(2023)Lin, Yin, Ping, Lu, Molchanov, Tao, Mao, Kautz, Shoeybi, and Han]{lin2023vila}
Ji~Lin, Hongxu Yin, Wei Ping, Yao Lu, Pavlo Molchanov, Andrew Tao, Huizi Mao, Jan Kautz, Mohammad Shoeybi, and Song Han.
\newblock Vila: On pre-training for visual language models.
\newblock \emph{ArXiv preprint}, abs/2312.07533, 2023.
\newblock URL \url{https://arxiv.org/abs/2312.07533}.

\bibitem[Lin et~al.(2014)Lin, Maire, Belongie, Hays, Perona, Ramanan, Doll{\'a}r, and Zitnick]{lin2014microsoft}
Tsung-Yi Lin, Michael Maire, Serge Belongie, James Hays, Pietro Perona, Deva Ramanan, Piotr Doll{\'a}r, and C~Lawrence Zitnick.
\newblock Microsoft coco: Common objects in context.
\newblock In \emph{European conference on computer vision}, pp.\  740--755. Springer, 2014.

\bibitem[Liu et~al.(2023{\natexlab{a}})Liu, Guan, Li, Chen, Yacoob, Manocha, and Zhou]{liu2023hallusionbench}
Fuxiao Liu, Tianrui Guan, Zongxia Li, Lichang Chen, Yaser Yacoob, Dinesh Manocha, and Tianyi Zhou.
\newblock Hallusionbench: You see what you think? or you think what you see? an image-context reasoning benchmark challenging for gpt-4v (ision), llava-1.5, and other multi-modality models.
\newblock \emph{ArXiv preprint}, abs/2310.14566, 2023{\natexlab{a}}.
\newblock URL \url{https://arxiv.org/abs/2310.14566}.

\bibitem[Liu et~al.(2023{\natexlab{b}})Liu, Li, Li, and Lee]{liu2023improvedllava}
Haotian Liu, Chunyuan Li, Yuheng Li, and Yong~Jae Lee.
\newblock Improved baselines with visual instruction tuning.
\newblock \emph{ArXiv preprint}, abs/2310.03744, 2023{\natexlab{b}}.
\newblock URL \url{https://arxiv.org/abs/2310.03744}.

\bibitem[Liu et~al.(2023{\natexlab{c}})Liu, Li, Wu, and Lee]{liu2023visual}
Haotian Liu, Chunyuan Li, Qingyang Wu, and Yong~Jae Lee.
\newblock Visual instruction tuning.
\newblock In Alice Oh, Tristan Naumann, Amir Globerson, Kate Saenko, Moritz Hardt, and Sergey Levine (eds.), \emph{Advances in Neural Information Processing Systems 36: Annual Conference on Neural Information Processing Systems 2023, NeurIPS 2023, New Orleans, LA, USA, December 10 - 16, 2023}, 2023{\natexlab{c}}.
\newblock URL \url{http://papers.nips.cc/paper\_files/paper/2023/hash/6dcf277ea32ce3288914faf369fe6de0-Abstract-Conference.html}.

\bibitem[Liu et~al.(2024{\natexlab{a}})Liu, Li, Li, and Lee]{liu2024improved}
Haotian Liu, Chunyuan Li, Yuheng Li, and Yong~Jae Lee.
\newblock Improved baselines with visual instruction tuning.
\newblock In \emph{Proceedings of the IEEE/CVF Conference on Computer Vision and Pattern Recognition}, pp.\  26296--26306, 2024{\natexlab{a}}.

\bibitem[Liu et~al.(2024{\natexlab{b}})Liu, Li, Li, Li, Zhang, Shen, and Lee]{liu2024llavanext}
Haotian Liu, Chunyuan Li, Yuheng Li, Bo~Li, Yuanhan Zhang, Sheng Shen, and Yong~Jae Lee.
\newblock Llava-next: Improved reasoning, ocr, and world knowledge, 2024{\natexlab{b}}.
\newblock URL \url{https://llava-vl.github.io/blog/2024-01-30-llava-next/}.

\bibitem[Liu et~al.(2024{\natexlab{c}})Liu, Zheng, Muennighoff, Zeng, Dou, Pang, Jiang, and Lin]{liu2024regmix}
Qian Liu, Xiaosen Zheng, Niklas Muennighoff, Guangtao Zeng, Longxu Dou, Tianyu Pang, Jing Jiang, and Min Lin.
\newblock Regmix: Data mixture as regression for language model pre-training.
\newblock \emph{ArXiv preprint}, abs/2407.01492, 2024{\natexlab{c}}.
\newblock URL \url{https://arxiv.org/abs/2407.01492}.

\bibitem[Liu et~al.(2023{\natexlab{d}})Liu, Duan, Zhang, Li, Zhang, Zhao, Yuan, Wang, He, Liu, et~al.]{liu2023mmbench}
Yuan Liu, Haodong Duan, Yuanhan Zhang, Bo~Li, Songyang Zhang, Wangbo Zhao, Yike Yuan, Jiaqi Wang, Conghui He, Ziwei Liu, et~al.
\newblock Mmbench: Is your multimodal model an all-around player?
\newblock \emph{ArXiv preprint}, abs/2307.06281, 2023{\natexlab{d}}.
\newblock URL \url{https://arxiv.org/abs/2307.06281}.

\bibitem[Lu et~al.(2023)Lu, Clark, Zellers, Mottaghi, and Kembhavi]{lu2022unified}
Jiasen Lu, Christopher Clark, Rowan Zellers, Roozbeh Mottaghi, and Aniruddha Kembhavi.
\newblock {UNIFIED-IO:} {A} unified model for vision, language, and multi-modal tasks.
\newblock In \emph{The Eleventh International Conference on Learning Representations, {ICLR} 2023, Kigali, Rwanda, May 1-5, 2023}. OpenReview.net, 2023.
\newblock URL \url{https://openreview.net/pdf?id=E01k9048soZ}.

\bibitem[Lu et~al.(2024{\natexlab{a}})Lu, Clark, Lee, Zhang, Khosla, Marten, Hoiem, and Kembhavi]{lu2024unified}
Jiasen Lu, Christopher Clark, Sangho Lee, Zichen Zhang, Savya Khosla, Ryan Marten, Derek Hoiem, and Aniruddha Kembhavi.
\newblock Unified-io 2: Scaling autoregressive multimodal models with vision language audio and action.
\newblock In \emph{Proceedings of the IEEE/CVF Conference on Computer Vision and Pattern Recognition}, pp.\  26439--26455, 2024{\natexlab{a}}.

\bibitem[Lu et~al.(2022)Lu, Mishra, Xia, Qiu, Chang, Zhu, Tafjord, Clark, and Kalyan]{lu2022learn}
Pan Lu, Swaroop Mishra, Tanglin Xia, Liang Qiu, Kai{-}Wei Chang, Song{-}Chun Zhu, Oyvind Tafjord, Peter Clark, and Ashwin Kalyan.
\newblock Learn to explain: Multimodal reasoning via thought chains for science question answering.
\newblock In Sanmi Koyejo, S.~Mohamed, A.~Agarwal, Danielle Belgrave, K.~Cho, and A.~Oh (eds.), \emph{Advances in Neural Information Processing Systems 35: Annual Conference on Neural Information Processing Systems 2022, NeurIPS 2022, New Orleans, LA, USA, November 28 - December 9, 2022}, 2022.
\newblock URL \url{http://papers.nips.cc/paper\_files/paper/2022/hash/11332b6b6cf4485b84afadb1352d3a9a-Abstract-Conference.html}.

\bibitem[Lu et~al.(2024{\natexlab{b}})Lu, Bansal, Xia, Liu, Li, Hajishirzi, Cheng, Chang, Galley, and Gao]{lu2024mathvista}
Pan Lu, Hritik Bansal, Tony Xia, Jiacheng Liu, Chunyuan Li, Hannaneh Hajishirzi, Hao Cheng, Kai-Wei Chang, Michel Galley, and Jianfeng Gao.
\newblock Mathvista: Evaluating mathematical reasoning of foundation models in visual contexts.
\newblock In \emph{The Twelfth International Conference on Learning Representations}, 2024{\natexlab{b}}.
\newblock URL \url{https://openreview.net/forum?id=KUNzEQMWU7}.

\bibitem[McCann et~al.(2019)McCann, Keskar, Xiong, and Socher]{mccann2019the}
Bryan McCann, Nitish~Shirish Keskar, Caiming Xiong, and Richard Socher.
\newblock The natural language decathlon: Multitask learning as question answering, 2019.
\newblock URL \url{https://openreview.net/forum?id=B1lfHhR9tm}.

\bibitem[McKinzie et~al.(2024)McKinzie, Gan, Fauconnier, Dodge, Zhang, Dufter, Shah, Du, Peng, Weers, et~al.]{mckinzie2024mm1}
Brandon McKinzie, Zhe Gan, Jean-Philippe Fauconnier, Sam Dodge, Bowen Zhang, Philipp Dufter, Dhruti Shah, Xianzhi Du, Futang Peng, Floris Weers, et~al.
\newblock Mm1: Methods, analysis \& insights from multimodal llm pre-training.
\newblock \emph{ArXiv preprint}, abs/2403.09611, 2024.
\newblock URL \url{https://arxiv.org/abs/2403.09611}.

\bibitem[Menon \& Vondrick(2023)Menon and Vondrick]{menon2023visual}
Sachit Menon and Carl Vondrick.
\newblock Visual classification via description from large language models.
\newblock In \emph{The Eleventh International Conference on Learning Representations, {ICLR} 2023, Kigali, Rwanda, May 1-5, 2023}. OpenReview.net, 2023.
\newblock URL \url{https://openreview.net/pdf?id=jlAjNL8z5cs}.

\bibitem[Meyer \& Gurevych(2012)Meyer and Gurevych]{meyer2012wiktionary}
Christian~M Meyer and Iryna Gurevych.
\newblock \emph{Wiktionary: A new rival for expert-built lexicons? Exploring the possibilities of collaborative lexicography}.
\newblock na, 2012.

\bibitem[Miller(1992)]{miller1995wordnet}
George~A. Miller.
\newblock {W}ord{N}et: A lexical database for {E}nglish.
\newblock In \emph{Speech and Natural Language: Proceedings of a Workshop Held at Harriman, New York, {F}ebruary 23-26, 1992}, 1992.
\newblock URL \url{https://aclanthology.org/H92-1116}.

\bibitem[Mitra et~al.(2023)Mitra, Huang, Darrell, and Herzig]{mitra2023compositional}
Chancharik Mitra, Brandon Huang, Trevor Darrell, and Roei Herzig.
\newblock Compositional chain-of-thought prompting for large multimodal models.
\newblock \emph{ArXiv preprint}, abs/2311.17076, 2023.
\newblock URL \url{https://arxiv.org/abs/2311.17076}.

\bibitem[Muennighoff et~al.(2023)Muennighoff, Rush, Barak, Scao, Tazi, Piktus, Pyysalo, Wolf, and Raffel]{muennighoff2024scaling}
Niklas Muennighoff, Alexander~M. Rush, Boaz Barak, Teven~Le Scao, Nouamane Tazi, Aleksandra Piktus, Sampo Pyysalo, Thomas Wolf, and Colin~A. Raffel.
\newblock Scaling data-constrained language models.
\newblock In Alice Oh, Tristan Naumann, Amir Globerson, Kate Saenko, Moritz Hardt, and Sergey Levine (eds.), \emph{Advances in Neural Information Processing Systems 36: Annual Conference on Neural Information Processing Systems 2023, NeurIPS 2023, New Orleans, LA, USA, December 10 - 16, 2023}, 2023.
\newblock URL \url{http://papers.nips.cc/paper\_files/paper/2023/hash/9d89448b63ce1e2e8dc7af72c984c196-Abstract-Conference.html}.

\bibitem[Ordonez et~al.(2011)Ordonez, Kulkarni, and Berg]{ordonez2011sbu}
Vicente Ordonez, Girish Kulkarni, and Tamara~L. Berg.
\newblock Im2text: Describing images using 1 million captioned photographs.
\newblock In John Shawe{-}Taylor, Richard~S. Zemel, Peter~L. Bartlett, Fernando C.~N. Pereira, and Kilian~Q. Weinberger (eds.), \emph{Advances in Neural Information Processing Systems 24: 25th Annual Conference on Neural Information Processing Systems 2011. Proceedings of a meeting held 12-14 December 2011, Granada, Spain}, pp.\  1143--1151, 2011.
\newblock URL \url{https://proceedings.neurips.cc/paper/2011/hash/5dd9db5e033da9c6fb5ba83c7a7ebea9-Abstract.html}.

\bibitem[Pan et~al.(2023)Pan, Dong, Huang, Peng, Chen, and Wei]{pan2023kosmosg}
Xichen Pan, Li~Dong, Shaohan Huang, Zhiliang Peng, Wenhu Chen, and Furu Wei.
\newblock Kosmos-g: Generating images in context with multimodal large language models.
\newblock \emph{ArXiv preprint}, abs/2310.02992, 2023.
\newblock URL \url{https://arxiv.org/abs/2310.02992}.

\bibitem[Peng et~al.(2022)Peng, Dong, Bao, Ye, and Wei]{peng2022beitv2}
Zhiliang Peng, Li~Dong, Hangbo Bao, Qixiang Ye, and Furu Wei.
\newblock Beit v2: Masked image modeling with vector-quantized visual tokenizers.
\newblock \emph{ArXiv preprint}, abs/2208.06366, 2022.
\newblock URL \url{https://arxiv.org/abs/2208.06366}.

\bibitem[Peng et~al.(2023)Peng, Wang, Dong, Hao, Huang, Ma, and Wei]{peng2023kosmos}
Zhiliang Peng, Wenhui Wang, Li~Dong, Yaru Hao, Shaohan Huang, Shuming Ma, and Furu Wei.
\newblock Kosmos-2: Grounding multimodal large language models to the world.
\newblock \emph{ArXiv preprint}, abs/2306.14824, 2023.
\newblock URL \url{https://arxiv.org/abs/2306.14824}.

\bibitem[Qin et~al.(2024{\natexlab{a}})Qin, Chen, Fei, Chen, Li, and Che]{qin2024factors}
Libo Qin, Qiguang Chen, Hao Fei, Zhi Chen, Min Li, and Wanxiang Che.
\newblock What factors affect multi-modal in-context learning? an in-depth exploration.
\newblock \emph{ArXiv preprint}, abs/2410.20482, 2024{\natexlab{a}}.
\newblock URL \url{https://arxiv.org/abs/2410.20482}.

\bibitem[Qin et~al.(2024{\natexlab{b}})Qin, Chen, Feng, Wu, Zhang, Li, Li, Che, and Yu]{qin2024large}
Libo Qin, Qiguang Chen, Xiachong Feng, Yang Wu, Yongheng Zhang, Yinghui Li, Min Li, Wanxiang Che, and Philip~S Yu.
\newblock Large language models meet nlp: A survey.
\newblock \emph{ArXiv preprint}, abs/2405.12819, 2024{\natexlab{b}}.
\newblock URL \url{https://arxiv.org/abs/2405.12819}.

\bibitem[Radford et~al.(2021)Radford, Kim, Hallacy, Ramesh, Goh, Agarwal, Sastry, Askell, Mishkin, Clark, Krueger, and Sutskever]{radford2021learning}
Alec Radford, Jong~Wook Kim, Chris Hallacy, Aditya Ramesh, Gabriel Goh, Sandhini Agarwal, Girish Sastry, Amanda Askell, Pamela Mishkin, Jack Clark, Gretchen Krueger, and Ilya Sutskever.
\newblock Learning transferable visual models from natural language supervision.
\newblock In Marina Meila and Tong Zhang (eds.), \emph{Proceedings of the 38th International Conference on Machine Learning, {ICML} 2021, 18-24 July 2021, Virtual Event}, volume 139 of \emph{Proceedings of Machine Learning Research}, pp.\  8748--8763. {PMLR}, 2021.
\newblock URL \url{http://proceedings.mlr.press/v139/radford21a.html}.

\bibitem[Rasheed et~al.(2023)Rasheed, Maaz, Shaji, Shaker, Khan, Cholakkal, Anwer, Xing, Yang, and Khan]{rasheed2023glamm}
Hanoona Rasheed, Muhammad Maaz, Sahal Shaji, Abdelrahman Shaker, Salman Khan, Hisham Cholakkal, Rao~M Anwer, Erix Xing, Ming-Hsuan Yang, and Fahad~S Khan.
\newblock Glamm: Pixel grounding large multimodal model.
\newblock \emph{ArXiv preprint}, abs/2311.03356, 2023.
\newblock URL \url{https://arxiv.org/abs/2311.03356}.

\bibitem[Rombach et~al.(2022)Rombach, Blattmann, Lorenz, Esser, and Ommer]{rombach2022high}
Robin Rombach, Andreas Blattmann, Dominik Lorenz, Patrick Esser, and Bj{\"o}rn Ommer.
\newblock High-resolution image synthesis with latent diffusion models.
\newblock In \emph{Proceedings of the IEEE/CVF conference on computer vision and pattern recognition}, pp.\  10684--10695, 2022.

\bibitem[Ruiz et~al.(2023)Ruiz, Li, Jampani, Pritch, Rubinstein, and Aberman]{ruiz2023dreambooth}
Nataniel Ruiz, Yuanzhen Li, Varun Jampani, Yael Pritch, Michael Rubinstein, and Kfir Aberman.
\newblock Dreambooth: Fine tuning text-to-image diffusion models for subject-driven generation.
\newblock In \emph{{IEEE/CVF} Conference on Computer Vision and Pattern Recognition, {CVPR} 2023, Vancouver, BC, Canada, June 17-24, 2023}, pp.\  22500--22510. {IEEE}, 2023.
\newblock \doi{10.1109/CVPR52729.2023.02155}.
\newblock URL \url{https://doi.org/10.1109/CVPR52729.2023.02155}.

\bibitem[Schuhmann et~al.(2021)Schuhmann, Kaczmarczyk, Komatsuzaki, Katta, Vencu, Beaumont, Jitsev, Coombes, and Mullis]{schuhmann2021laion}
Christoph Schuhmann, Robert Kaczmarczyk, Aran Komatsuzaki, Aarush Katta, Richard Vencu, Romain Beaumont, Jenia Jitsev, Theo Coombes, and Clayton Mullis.
\newblock Laion-400m: Open dataset of clip-filtered 400 million image-text pairs.
\newblock In \emph{NeurIPS Workshop Datacentric AI}, number FZJ-2022-00923. J{\"u}lich Supercomputing Center, 2021.

\bibitem[Schuhmann et~al.(2022{\natexlab{a}})Schuhmann, Beaumont, Vencu, Gordon, Wightman, Cherti, Coombes, Katta, Mullis, Wortsman, Schramowski, Kundurthy, Crowson, Schmidt, Kaczmarczyk, and Jitsev]{schuhmann2022laion}
Christoph Schuhmann, Romain Beaumont, Richard Vencu, Cade Gordon, Ross Wightman, Mehdi Cherti, Theo Coombes, Aarush Katta, Clayton Mullis, Mitchell Wortsman, Patrick Schramowski, Srivatsa Kundurthy, Katherine Crowson, Ludwig Schmidt, Robert Kaczmarczyk, and Jenia Jitsev.
\newblock {LAION-5B:} an open large-scale dataset for training next generation image-text models.
\newblock In Sanmi Koyejo, S.~Mohamed, A.~Agarwal, Danielle Belgrave, K.~Cho, and A.~Oh (eds.), \emph{Advances in Neural Information Processing Systems 35: Annual Conference on Neural Information Processing Systems 2022, NeurIPS 2022, New Orleans, LA, USA, November 28 - December 9, 2022}, 2022{\natexlab{a}}.
\newblock URL \url{http://papers.nips.cc/paper\_files/paper/2022/hash/a1859debfb3b59d094f3504d5ebb6c25-Abstract-Datasets\_and\_Benchmarks.html}.

\bibitem[Schuhmann et~al.(2022{\natexlab{b}})Schuhmann, Köpf, Vencu, Coombes, and Beaumont]{laioncoco}
Christoph Schuhmann, Andreas Köpf, Richard Vencu, Theo Coombes, and Romain Beaumont.
\newblock Laion-coco.
\newblock \url{https://huggingface.co/datasets/laion/laion-coco}, 2022{\natexlab{b}}.

\bibitem[Schuhmann et~al.(2023)Schuhmann, Köpf, Vencu, Coombes, and Beaumont]{laion2023laioncoco}
Christoph Schuhmann, Andreas Köpf, Richard Vencu, Theo Coombes, and Romain Beaumont.
\newblock Laion coco: 600m synthetic captions from laion2b-en, 2023.
\newblock URL \url{https://laion.ai/blog/laion-coco/}.

\bibitem[Shao et~al.(2019)Shao, Li, Zhang, Peng, Yu, Zhang, Li, and Sun]{shao2019objects365}
Shuai Shao, Zeming Li, Tianyuan Zhang, Chao Peng, Gang Yu, Xiangyu Zhang, Jing Li, and Jian Sun.
\newblock Objects365: {A} large-scale, high-quality dataset for object detection.
\newblock In \emph{2019 {IEEE/CVF} International Conference on Computer Vision, {ICCV} 2019, Seoul, Korea (South), October 27 - November 2, 2019}, pp.\  8429--8438. {IEEE}, 2019.
\newblock \doi{10.1109/ICCV.2019.00852}.
\newblock URL \url{https://doi.org/10.1109/ICCV.2019.00852}.

\bibitem[Sharma et~al.(2018)Sharma, Ding, Goodman, and Soricut]{sharma-etal-2018-conceptual}
Piyush Sharma, Nan Ding, Sebastian Goodman, and Radu Soricut.
\newblock Conceptual captions: A cleaned, hypernymed, image alt-text dataset for automatic image captioning.
\newblock In Iryna Gurevych and Yusuke Miyao (eds.), \emph{Proceedings of the 56th Annual Meeting of the Association for Computational Linguistics (Volume 1: Long Papers)}, pp.\  2556--2565, Melbourne, Australia, 2018. Association for Computational Linguistics.
\newblock \doi{10.18653/v1/P18-1238}.
\newblock URL \url{https://aclanthology.org/P18-1238}.

\bibitem[Shen et~al.(2022)Shen, Li, Hu, Xie, Yang, Zhang, Gan, Wang, Yuan, Liu, Keutzer, Darrell, Rohrbach, and Gao]{shen2022k}
Sheng Shen, Chunyuan Li, Xiaowei Hu, Yujia Xie, Jianwei Yang, Pengchuan Zhang, Zhe Gan, Lijuan Wang, Lu~Yuan, Ce~Liu, Kurt Keutzer, Trevor Darrell, Anna Rohrbach, and Jianfeng Gao.
\newblock {K-LITE:} learning transferable visual models with external knowledge.
\newblock In Sanmi Koyejo, S.~Mohamed, A.~Agarwal, Danielle Belgrave, K.~Cho, and A.~Oh (eds.), \emph{Advances in Neural Information Processing Systems 35: Annual Conference on Neural Information Processing Systems 2022, NeurIPS 2022, New Orleans, LA, USA, November 28 - December 9, 2022}, 2022.
\newblock URL \url{http://papers.nips.cc/paper\_files/paper/2022/hash/63fef0802863f47775c3563e18cbba17-Abstract-Conference.html}.

\bibitem[Shevade et~al.(2005)Shevade, Sundaram, and Yen-Kan]{shevade2005collaborative}
Bageshree Shevade, Hari Sundaram, and Min Yen-Kan.
\newblock A collaborative annotation framework.
\newblock In \emph{2005 IEEE International Conference on Multimedia and Expo}, pp.\  1346--1349. IEEE, 2005.

\bibitem[Shtedritski et~al.(2023)Shtedritski, Rupprecht, and Vedaldi]{shtedritski2023does}
Aleksandar Shtedritski, Christian Rupprecht, and Andrea Vedaldi.
\newblock What does {CLIP} know about a red circle? visual prompt engineering for vlms.
\newblock In \emph{{IEEE/CVF} International Conference on Computer Vision, {ICCV} 2023, Paris, France, October 1-6, 2023}, pp.\  11953--11963. {IEEE}, 2023.
\newblock \doi{10.1109/ICCV51070.2023.01101}.
\newblock URL \url{https://doi.org/10.1109/ICCV51070.2023.01101}.

\bibitem[Sohl{-}Dickstein et~al.(2015)Sohl{-}Dickstein, Weiss, Maheswaranathan, and Ganguli]{sohl2015deep}
Jascha Sohl{-}Dickstein, Eric~A. Weiss, Niru Maheswaranathan, and Surya Ganguli.
\newblock Deep unsupervised learning using nonequilibrium thermodynamics.
\newblock In Francis~R. Bach and David~M. Blei (eds.), \emph{Proceedings of the 32nd International Conference on Machine Learning, {ICML} 2015, Lille, France, 6-11 July 2015}, volume~37 of \emph{{JMLR} Workshop and Conference Proceedings}, pp.\  2256--2265. JMLR.org, 2015.
\newblock URL \url{http://proceedings.mlr.press/v37/sohl-dickstein15.html}.

\bibitem[Soviany et~al.(2022)Soviany, Ionescu, Rota, and Sebe]{soviany2022curriculum}
Petru Soviany, Radu~Tudor Ionescu, Paolo Rota, and Nicu Sebe.
\newblock Curriculum learning: A survey.
\newblock \emph{International Journal of Computer Vision}, 130\penalty0 (6):\penalty0 1526--1565, 2022.

\bibitem[Steven~Tey(2023)]{sharegpt}
ChatGPT Steven~Tey.
\newblock Sharegpt.
\newblock \url{https://sharegpt.com/}, 2023.

\bibitem[Sun et~al.(2023{\natexlab{a}})Sun, Pan, Ge, Li, Duan, Wu, Zhang, Zhou, Qin, Wang, Dai, Qiao, Wang, and Li]{sun2024journeydb}
Keqiang Sun, Junting Pan, Yuying Ge, Hao Li, Haodong Duan, Xiaoshi Wu, Renrui Zhang, Aojun Zhou, Zipeng Qin, Yi~Wang, Jifeng Dai, Yu~Qiao, Limin Wang, and Hongsheng Li.
\newblock Journeydb: {A} benchmark for generative image understanding.
\newblock In Alice Oh, Tristan Naumann, Amir Globerson, Kate Saenko, Moritz Hardt, and Sergey Levine (eds.), \emph{Advances in Neural Information Processing Systems 36: Annual Conference on Neural Information Processing Systems 2023, NeurIPS 2023, New Orleans, LA, USA, December 10 - 16, 2023}, 2023{\natexlab{a}}.
\newblock URL \url{http://papers.nips.cc/paper\_files/paper/2023/hash/9bc59aff4685e39e1a8175d5303248a1-Abstract-Datasets\_and\_Benchmarks.html}.

\bibitem[Sun et~al.(2023{\natexlab{b}})Sun, Cui, Zhang, Zhang, Yu, Luo, Wang, Rao, Liu, Huang, et~al.]{sun2023generative2}
Quan Sun, Yufeng Cui, Xiaosong Zhang, Fan Zhang, Qiying Yu, Zhengxiong Luo, Yueze Wang, Yongming Rao, Jingjing Liu, Tiejun Huang, et~al.
\newblock Generative multimodal models are in-context learners.
\newblock \emph{ArXiv preprint}, abs/2312.13286, 2023{\natexlab{b}}.
\newblock URL \url{https://arxiv.org/abs/2312.13286}.

\bibitem[Sun et~al.(2023{\natexlab{c}})Sun, Fang, Wu, Wang, and Cao]{sun2023eva}
Quan Sun, Yuxin Fang, Ledell Wu, Xinlong Wang, and Yue Cao.
\newblock Eva-clip: Improved training techniques for clip at scale.
\newblock \emph{ArXiv preprint}, abs/2303.15389, 2023{\natexlab{c}}.
\newblock URL \url{https://arxiv.org/abs/2303.15389}.

\bibitem[Sun et~al.(2023{\natexlab{d}})Sun, Yu, Cui, Zhang, Zhang, Wang, Gao, Liu, Huang, and Wang]{sun2023generative}
Quan Sun, Qiying Yu, Yufeng Cui, Fan Zhang, Xiaosong Zhang, Yueze Wang, Hongcheng Gao, Jingjing Liu, Tiejun Huang, and Xinlong Wang.
\newblock Generative pretraining in multimodality.
\newblock \emph{ArXiv preprint}, abs/2307.05222, 2023{\natexlab{d}}.
\newblock URL \url{https://arxiv.org/abs/2307.05222}.

\bibitem[Taori et~al.(2023)Taori, Gulrajani, Zhang, Dubois, Li, Guestrin, Liang, and Hashimoto]{alpaca}
Rohan Taori, Ishaan Gulrajani, Tianyi Zhang, Yann Dubois, Xuechen Li, Carlos Guestrin, Percy Liang, and Tatsunori~B. Hashimoto.
\newblock Stanford alpaca: An instruction-following llama model.
\newblock \url{https://github.com/tatsu-lab/stanford_alpaca}, 2023.

\bibitem[Team(2024)]{team2024chameleon}
Chameleon Team.
\newblock Chameleon: Mixed-modal early-fusion foundation models.
\newblock \emph{ArXiv preprint}, abs/2405.09818, 2024.
\newblock URL \url{https://arxiv.org/abs/2405.09818}.

\bibitem[Tian et~al.(2024)Tian, Zhu, Xiong, Wang, Chen, Wang, Chen, Lu, Lu, Zhou, Li, Qiao, and Dai]{tian2024mminterleaved}
Changyao Tian, Xizhou Zhu, Yuwen Xiong, Weiyun Wang, Zhe Chen, Wenhai Wang, Yuntao Chen, Lewei Lu, Tong Lu, Jie Zhou, Hongsheng Li, Yu~Qiao, and Jifeng Dai.
\newblock Mm-interleaved: Interleaved image-text generative modeling via multi-modal feature synchronizer, 2024.

\bibitem[Tong et~al.(2024)Tong, II, Wu, Woo, IYER, Akula, Yang, Yang, Middepogu, Wang, Pan, Fergus, LeCun, and Xie]{tong2024cambrian}
Shengbang Tong, Ellis L~Brown II, Penghao Wu, Sanghyun Woo, ADITHYA~JAIRAM IYER, Sai~Charitha Akula, Shusheng Yang, Jihan Yang, Manoj Middepogu, Ziteng Wang, Xichen Pan, Rob Fergus, Yann LeCun, and Saining Xie.
\newblock Cambrian-1: A fully open, vision-centric exploration of multimodal {LLM}s.
\newblock In \emph{The Thirty-eighth Annual Conference on Neural Information Processing Systems}, 2024.
\newblock URL \url{https://openreview.net/forum?id=Vi8AepAXGy}.

\bibitem[Touvron et~al.(2023{\natexlab{a}})Touvron, Lavril, Izacard, Martinet, Lachaux, Lacroix, Rozi{\`e}re, Goyal, Hambro, Azhar, et~al.]{touvron2023llama}
Hugo Touvron, Thibaut Lavril, Gautier Izacard, Xavier Martinet, Marie-Anne Lachaux, Timoth{\'e}e Lacroix, Baptiste Rozi{\`e}re, Naman Goyal, Eric Hambro, Faisal Azhar, et~al.
\newblock Llama: Open and efficient foundation language models.
\newblock \emph{ArXiv preprint}, abs/2302.13971, 2023{\natexlab{a}}.
\newblock URL \url{https://arxiv.org/abs/2302.13971}.

\bibitem[Touvron et~al.(2023{\natexlab{b}})Touvron, Martin, Stone, Albert, Almahairi, Babaei, Bashlykov, Batra, Bhargava, Bhosale, et~al.]{touvron2023llama2}
Hugo Touvron, Louis Martin, Kevin Stone, Peter Albert, Amjad Almahairi, Yasmine Babaei, Nikolay Bashlykov, Soumya Batra, Prajjwal Bhargava, Shruti Bhosale, et~al.
\newblock Llama 2: Open foundation and fine-tuned chat models.
\newblock \emph{ArXiv preprint}, abs/2307.09288, 2023{\natexlab{b}}.
\newblock URL \url{https://arxiv.org/abs/2307.09288}.

\bibitem[van~den Oord et~al.(2017)van~den Oord, Vinyals, and Kavukcuoglu]{van2017neural}
A{\"{a}}ron van~den Oord, Oriol Vinyals, and Koray Kavukcuoglu.
\newblock Neural discrete representation learning.
\newblock In Isabelle Guyon, Ulrike von Luxburg, Samy Bengio, Hanna~M. Wallach, Rob Fergus, S.~V.~N. Vishwanathan, and Roman Garnett (eds.), \emph{Advances in Neural Information Processing Systems 30: Annual Conference on Neural Information Processing Systems 2017, December 4-9, 2017, Long Beach, CA, {USA}}, pp.\  6306--6315, 2017.
\newblock URL \url{https://proceedings.neurips.cc/paper/2017/hash/7a98af17e63a0ac09ce2e96d03992fbc-Abstract.html}.

\bibitem[Wang et~al.(2023{\natexlab{a}})Wang, Zhang, Chu, Cao, Zhou, Wu, Wang, He, and Lin]{wang2023v3det}
Jiaqi Wang, Pan Zhang, Tao Chu, Yuhang Cao, Yujie Zhou, Tong Wu, Bin Wang, Conghui He, and Dahua Lin.
\newblock V3det: Vast vocabulary visual detection dataset.
\newblock In \emph{{IEEE/CVF} International Conference on Computer Vision, {ICCV} 2023, Paris, France, October 1-6, 2023}, pp.\  19787--19797. {IEEE}, 2023{\natexlab{a}}.
\newblock \doi{10.1109/ICCV51070.2023.01817}.
\newblock URL \url{https://doi.org/10.1109/ICCV51070.2023.01817}.

\bibitem[Wang et~al.(2022)Wang, Yang, Men, Lin, Bai, Li, Ma, Zhou, Zhou, and Yang]{wang2022OFA}
Peng Wang, An~Yang, Rui Men, Junyang Lin, Shuai Bai, Zhikang Li, Jianxin Ma, Chang Zhou, Jingren Zhou, and Hongxia Yang.
\newblock Unifying architectures, tasks, and modalities through a simple sequence-to-sequence learning framework.
\newblock \emph{ArXiv preprint}, abs/2202.03052, 2022.
\newblock URL \url{https://arxiv.org/abs/2202.03052}.

\bibitem[Wang et~al.(2023{\natexlab{b}})Wang, Shi, Li, Wang, Huang, Xing, Chen, Li, Zhu, Cao, et~al.]{wang2023all}
Weiyun Wang, Min Shi, Qingyun Li, Wenhai Wang, Zhenhang Huang, Linjie Xing, Zhe Chen, Hao Li, Xizhou Zhu, Zhiguo Cao, et~al.
\newblock The all-seeing project: Towards panoptic visual recognition and understanding of the open world.
\newblock In \emph{The Twelfth International Conference on Learning Representations}, 2023{\natexlab{b}}.

\bibitem[Wu et~al.(2023{\natexlab{a}})Wu, Zhang, Zhang, Chen, Liao, Wang, Li, Sun, Yan, Zhai, et~al.]{wu2023q}
Haoning Wu, Zicheng Zhang, Erli Zhang, Chaofeng Chen, Liang Liao, Annan Wang, Chunyi Li, Wenxiu Sun, Qiong Yan, Guangtao Zhai, et~al.
\newblock Q-bench: A benchmark for general-purpose foundation models on low-level vision.
\newblock \emph{ArXiv preprint}, abs/2309.14181, 2023{\natexlab{a}}.
\newblock URL \url{https://arxiv.org/abs/2309.14181}.

\bibitem[Wu et~al.(2023{\natexlab{b}})Wu, Zhang, Xiong, Oguz, Gee, and Nie]{wu2023role}
Yifan Wu, Pengchuan Zhang, Wenhan Xiong, Barlas Oguz, James~C Gee, and Yixin Nie.
\newblock The role of chain-of-thought in complex vision--language reasoning task.
\newblock \emph{ArXiv preprint}, abs/2311.09193, 2023{\natexlab{b}}.
\newblock URL \url{https://arxiv.org/abs/2311.09193}.

\bibitem[Xie et~al.(2020)Xie, Deng, Li, Liu, and Tao]{xie2020multi}
De~Xie, Cheng Deng, Chao Li, Xianglong Liu, and Dacheng Tao.
\newblock Multi-task consistency-preserving adversarial hashing for cross-modal retrieval.
\newblock \emph{IEEE Transactions on Image Processing}, 29:\penalty0 3626--3637, 2020.

\bibitem[Xu et~al.(2023{\natexlab{a}})Xu, Li, Wu, Tseng, Bhiwandiwalla, Rosenman, Lal, Che, and Duan]{xu-etal-2023-managertower}
Xiao Xu, Bei Li, Chenfei Wu, Shao-Yen Tseng, Anahita Bhiwandiwalla, Shachar Rosenman, Vasudev Lal, Wanxiang Che, and Nan Duan.
\newblock {M}anager{T}ower: Aggregating the insights of uni-modal experts for vision-language representation learning.
\newblock In Anna Rogers, Jordan Boyd-Graber, and Naoaki Okazaki (eds.), \emph{Proceedings of the 61st Annual Meeting of the Association for Computational Linguistics (Volume 1: Long Papers)}, pp.\  14507--14525, Toronto, Canada, 2023{\natexlab{a}}. Association for Computational Linguistics.
\newblock \doi{10.18653/v1/2023.acl-long.811}.
\newblock URL \url{https://aclanthology.org/2023.acl-long.811}.

\bibitem[Xu et~al.(2023{\natexlab{b}})Xu, Wu, Rosenman, Lal, Che, and Duan]{xu2023bridgetower}
Xiao Xu, Chenfei Wu, Shachar Rosenman, Vasudev Lal, Wanxiang Che, and Nan Duan.
\newblock Bridgetower: Building bridges between encoders in vision-language representation learning.
\newblock In Brian Williams, Yiling Chen, and Jennifer Neville (eds.), \emph{Thirty-Seventh {AAAI} Conference on Artificial Intelligence, {AAAI} 2023, Thirty-Fifth Conference on Innovative Applications of Artificial Intelligence, {IAAI} 2023, Thirteenth Symposium on Educational Advances in Artificial Intelligence, {EAAI} 2023, Washington, DC, USA, February 7-14, 2023}, pp.\  10637--10647. {AAAI} Press, 2023{\natexlab{b}}.
\newblock \doi{10.1609/AAAI.V37I9.26263}.
\newblock URL \url{https://doi.org/10.1609/aaai.v37i9.26263}.

\bibitem[Xu et~al.(2020)Xu, Wang, Yang, Zuo, Shen, and Shen]{xu2020cross}
Xing Xu, Tan Wang, Yang Yang, Lin Zuo, Fumin Shen, and Heng~Tao Shen.
\newblock Cross-modal attention with semantic consistence for image--text matching.
\newblock \emph{IEEE transactions on neural networks and learning systems}, 31\penalty0 (12):\penalty0 5412--5425, 2020.

\bibitem[Yang et~al.(2023)Yang, Zhang, Li, Zou, Li, and Gao]{yang2023set}
Jianwei Yang, Hao Zhang, Feng Li, Xueyan Zou, Chunyuan Li, and Jianfeng Gao.
\newblock Set-of-mark prompting unleashes extraordinary visual grounding in gpt-4v.
\newblock \emph{ArXiv preprint}, abs/2310.11441, 2023.
\newblock URL \url{https://arxiv.org/abs/2310.11441}.

\bibitem[Yao et~al.(2022)Yao, Han, Wen, Liang, Xu, Zhang, Li, Xu, and Xu]{yao2022detclip}
Lewei Yao, Jianhua Han, Youpeng Wen, Xiaodan Liang, Dan Xu, Wei Zhang, Zhenguo Li, Chunjing Xu, and Hang Xu.
\newblock Detclip: Dictionary-enriched visual-concept paralleled pre-training for open-world detection.
\newblock In Sanmi Koyejo, S.~Mohamed, A.~Agarwal, Danielle Belgrave, K.~Cho, and A.~Oh (eds.), \emph{Advances in Neural Information Processing Systems 35: Annual Conference on Neural Information Processing Systems 2022, NeurIPS 2022, New Orleans, LA, USA, November 28 - December 9, 2022}, 2022.
\newblock URL \url{http://papers.nips.cc/paper\_files/paper/2022/hash/3ba960559212691be13fa81d9e5e0047-Abstract-Conference.html}.

\bibitem[Yao et~al.(2024)Yao, Zhang, Zhang, Liu, Chua, and Sun]{yao2024cpt}
Yuan Yao, Ao~Zhang, Zhengyan Zhang, Zhiyuan Liu, Tat-Seng Chua, and Maosong Sun.
\newblock Cpt: Colorful prompt tuning for pre-trained vision--language models.
\newblock \emph{AI Open}, 5:\penalty0 30--38, 2024.

\bibitem[Yin et~al.(2023)Yin, Fu, Zhao, Li, Sun, Xu, and Chen]{yin2023survey}
Shukang Yin, Chaoyou Fu, Sirui Zhao, Ke~Li, Xing Sun, Tong Xu, and Enhong Chen.
\newblock A survey on multimodal large language models.
\newblock \emph{ArXiv preprint}, abs/2306.13549, 2023.
\newblock URL \url{https://arxiv.org/abs/2306.13549}.

\bibitem[You et~al.(2023)You, Zhang, Gan, Du, Zhang, Wang, Cao, Chang, and Yang]{you2023ferret}
Haoxuan You, Haotian Zhang, Zhe Gan, Xianzhi Du, Bowen Zhang, Zirui Wang, Liangliang Cao, Shih-Fu Chang, and Yinfei Yang.
\newblock Ferret: Refer and ground anything anywhere at any granularity.
\newblock In \emph{The Twelfth International Conference on Learning Representations}, 2023.

\bibitem[Yu et~al.(2023{\natexlab{a}})Yu, Shi, Pasunuru, Muller, Golovneva, Wang, Babu, Tang, Karrer, Sheynin, et~al.]{yu2023scaling}
Lili Yu, Bowen Shi, Ramakanth Pasunuru, Benjamin Muller, Olga Golovneva, Tianlu Wang, Arun Babu, Binh Tang, Brian Karrer, Shelly Sheynin, et~al.
\newblock Scaling autoregressive multi-modal models: Pretraining and instruction tuning.
\newblock \emph{ArXiv preprint}, abs/2309.02591, 2023{\natexlab{a}}.
\newblock URL \url{https://arxiv.org/abs/2309.02591}.

\bibitem[Yu et~al.(2023{\natexlab{b}})Yu, Sun, Zhang, Cui, Zhang, Wang, and Liu]{yu2023capsfusion}
Qiying Yu, Quan Sun, Xiaosong Zhang, Yufeng Cui, Fan Zhang, Xinlong Wang, and Jingjing Liu.
\newblock Capsfusion: Rethinking image--text data at scale.
\newblock \emph{ArXiv preprint}, abs/2310.20550, 2023{\natexlab{b}}.
\newblock URL \url{https://arxiv.org/abs/2310.20550}.

\bibitem[Yu et~al.(2023{\natexlab{c}})Yu, Hu, Yao, Zhang, Zhao, Wang, Wang, Pan, Xue, Li, et~al.]{yu2023reformulating}
Tianyu Yu, Jinyi Hu, Yuan Yao, Haoye Zhang, Yue Zhao, Chongyi Wang, Shan Wang, Yinxv Pan, Jiao Xue, Dahai Li, et~al.
\newblock Reformulating vision--language foundation models and datasets towards universal multimodal assistants.
\newblock \emph{ArXiv preprint}, abs/2310.00653, 2023{\natexlab{c}}.
\newblock URL \url{https://arxiv.org/abs/2310.00653}.

\bibitem[Yue et~al.(2023)Yue, Ni, Zhang, Zheng, Liu, Zhang, Stevens, Jiang, Ren, Sun, et~al.]{yue2023mmmu}
Xiang Yue, Yuansheng Ni, Kai Zhang, Tianyu Zheng, Ruoqi Liu, Ge~Zhang, Samuel Stevens, Dongfu Jiang, Weiming Ren, Yuxuan Sun, et~al.
\newblock Mmmu: A massive multi-discipline multimodal understanding and reasoning benchmark for expert agi.
\newblock \emph{ArXiv preprint}, abs/2311.16502, 2023.
\newblock URL \url{https://arxiv.org/abs/2311.16502}.

\bibitem[Zeng et~al.(2022)Zeng, Zhang, and Li]{zeng2021multi}
Yan Zeng, Xinsong Zhang, and Hang Li.
\newblock Multi-grained vision language pre-training: Aligning texts with visual concepts.
\newblock In Kamalika Chaudhuri, Stefanie Jegelka, Le~Song, Csaba Szepesv{\'{a}}ri, Gang Niu, and Sivan Sabato (eds.), \emph{International Conference on Machine Learning, {ICML} 2022, 17-23 July 2022, Baltimore, Maryland, {USA}}, volume 162 of \emph{Proceedings of Machine Learning Research}, pp.\  25994--26009. {PMLR}, 2022.
\newblock URL \url{https://proceedings.mlr.press/v162/zeng22c.html}.

\bibitem[Zeng et~al.(2023)Zeng, Zhang, Li, Wang, Zhang, and Zhou]{zeng2023x}
Yan Zeng, Xinsong Zhang, Hang Li, Jiawei Wang, Jipeng Zhang, and Wangchunshu Zhou.
\newblock X 2-vlm: All-in-one pre-trained model for vision-language tasks.
\newblock \emph{IEEE Transactions on Pattern Analysis and Machine Intelligence}, 2023.

\bibitem[Zhang et~al.(2024{\natexlab{a}})Zhang, Gao, Gan, Dufter, Wenzel, Huang, Shah, Du, Zhang, Li, et~al.]{zhang2024mm1}
Haotian Zhang, Mingfei Gao, Zhe Gan, Philipp Dufter, Nina Wenzel, Forrest Huang, Dhruti Shah, Xianzhi Du, Bowen Zhang, Yanghao Li, et~al.
\newblock Mm1. 5: Methods, analysis \& insights from multimodal llm fine-tuning.
\newblock \emph{ArXiv preprint}, abs/2409.20566, 2024{\natexlab{a}}.
\newblock URL \url{https://arxiv.org/abs/2409.20566}.

\bibitem[Zhang et~al.(2023{\natexlab{a}})Zhang, Mo, Chen, Sun, and Su]{zhang2024magicbrush}
Kai Zhang, Lingbo Mo, Wenhu Chen, Huan Sun, and Yu~Su.
\newblock Magicbrush: {A} manually annotated dataset for instruction-guided image editing.
\newblock In Alice Oh, Tristan Naumann, Amir Globerson, Kate Saenko, Moritz Hardt, and Sergey Levine (eds.), \emph{Advances in Neural Information Processing Systems 36: Annual Conference on Neural Information Processing Systems 2023, NeurIPS 2023, New Orleans, LA, USA, December 10 - 16, 2023}, 2023{\natexlab{a}}.
\newblock URL \url{http://papers.nips.cc/paper\_files/paper/2023/hash/64008fa30cba9b4d1ab1bd3bd3d57d61-Abstract-Datasets\_and\_Benchmarks.html}.

\bibitem[Zhang et~al.(2021)Zhang, Li, Hu, Yang, Zhang, Wang, Choi, and Gao]{zhang2021vinvl}
Pengchuan Zhang, Xiujun Li, Xiaowei Hu, Jianwei Yang, Lei Zhang, Lijuan Wang, Yejin Choi, and Jianfeng Gao.
\newblock Vinvl: Revisiting visual representations in vision-language models.
\newblock In \emph{{IEEE} Conference on Computer Vision and Pattern Recognition, {CVPR} 2021, virtual, June 19-25, 2021}, pp.\  5579--5588. Computer Vision Foundation / {IEEE}, 2021.
\newblock \doi{10.1109/CVPR46437.2021.00553}.
\newblock URL \url{https://openaccess.thecvf.com/content/CVPR2021/html/Zhang\_VinVL\_Revisiting\_Visual\_Representations\_in\_Vision-Language\_Models\_CVPR\_2021\_paper.html}.

\bibitem[Zhang et~al.(2023{\natexlab{b}})Zhang, Zhang, Gu, Zhou, Lipka, Yang, and Sun]{zhang2023llavar}
Yanzhe Zhang, Ruiyi Zhang, Jiuxiang Gu, Yufan Zhou, Nedim Lipka, Diyi Yang, and Tong Sun.
\newblock Llavar: Enhanced visual instruction tuning for text-rich image understanding.
\newblock \emph{ArXiv preprint}, abs/2306.17107, 2023{\natexlab{b}}.
\newblock URL \url{https://arxiv.org/abs/2306.17107}.

\bibitem[Zhang et~al.(2024{\natexlab{b}})Zhang, Zhang, Li, hai zhao, Karypis, and Smola]{zhang2024multimodal}
Zhuosheng Zhang, Aston Zhang, Mu~Li, hai zhao, George Karypis, and Alex Smola.
\newblock Multimodal chain-of-thought reasoning in language models.
\newblock \emph{Transactions on Machine Learning Research}, 2024{\natexlab{b}}.
\newblock ISSN 2835-8856.
\newblock URL \url{https://openreview.net/forum?id=y1pPWFVfvR}.

\bibitem[Zhao et~al.(2024)Zhao, Cai, Si, Ma, An, Chen, Liu, Wang, Han, and Chang]{zhao2024mmicl}
Haozhe Zhao, Zefan Cai, Shuzheng Si, Xiaojian Ma, Kaikai An, Liang Chen, Zixuan Liu, Sheng Wang, Wenjuan Han, and Baobao Chang.
\newblock {MMICL}: Empowering vision-language model with multi-modal in-context learning.
\newblock In \emph{The Twelfth International Conference on Learning Representations}, 2024.
\newblock URL \url{https://openreview.net/forum?id=5KojubHBr8}.

\bibitem[Zhu et~al.(2023{\natexlab{a}})Zhu, Ding, Ge, Ge, Zhao, Zhao, Wang, and Shan]{zhu2023vl}
Jinguo Zhu, Xiaohan Ding, Yixiao Ge, Yuying Ge, Sijie Zhao, Hengshuang Zhao, Xiaohua Wang, and Ying Shan.
\newblock Vl-gpt: A generative pre-trained transformer for vision and language understanding and generation.
\newblock \emph{ArXiv preprint}, abs/2312.09251, 2023{\natexlab{a}}.
\newblock URL \url{https://arxiv.org/abs/2312.09251}.

\bibitem[Zhu et~al.(2023{\natexlab{b}})Zhu, Hessel, Awadalla, Gadre, Dodge, Fang, Yu, Schmidt, Wang, and Choi]{zhu2023multimodal}
Wanrong Zhu, Jack Hessel, Anas Awadalla, Samir~Yitzhak Gadre, Jesse Dodge, Alex Fang, Youngjae Yu, Ludwig Schmidt, William~Yang Wang, and Yejin Choi.
\newblock Multimodal {C4:} an open, billion-scale corpus of images interleaved with text.
\newblock In Alice Oh, Tristan Naumann, Amir Globerson, Kate Saenko, Moritz Hardt, and Sergey Levine (eds.), \emph{Advances in Neural Information Processing Systems 36: Annual Conference on Neural Information Processing Systems 2023, NeurIPS 2023, New Orleans, LA, USA, December 10 - 16, 2023}, 2023{\natexlab{b}}.
\newblock URL \url{http://papers.nips.cc/paper\_files/paper/2023/hash/1c6bed78d3813886d3d72595dbecb80b-Abstract-Datasets\_and\_Benchmarks.html}.

\end{thebibliography}
\bibliographystyle{iclr2025_conference}

\appendix

\clearpage

\hypersetup{linkcolor=darkblue}

\tableofcontents

\clearpage

\hypersetup{linkcolor=linkcolor}

\noindent\makebox[\linewidth]{\rule{\linewidth}{1pt}}
\ \\
This appendix is organized as follows. 
\begin{itemize}[leftmargin=3em]
    \item In \Sref{sec:appendix:broader_impact}, we discuss the broader impact of our work.
    \item In \Sref{sec:appendix:limitation}, we discuss the limitations of our work and potential future research directions.
    \item In \Sref{sec:appendix:experiment}, we provide more experimental results and analyses, including concept overlap, image editing, self-generated annotations in evaluation, more cases on SEED-Bench, \datasetname{} directly as SFT data, the impact of the nature of image--text interleaved, etc.
    \item In \Sref{sec:appendix:discussion}, we provide discussions on comparison with existing MLLM frameworks, applicability of \datasetname{}, annotation synthesis, comparison with existing interleaved datasets, etc.\looseness=-1
    \item In \Sref{sec:appendix:experiment_details} (referred by \Sref{sec:method:fine-tuning} \& \ref{sec:experiment}), we provide more training and evaluation details of our experiments.
    \item In \Sref{sec:appendix:dataset} (referred by \Sref{sec:dataset}), we provide complete details of \datasetname{} dataset.
    \item In \Sref{sec:appendix:ic_dataset} (referred by \Sref{sec:experiment:scalable}), we provide more details of \icdatasetname{} dataset.
    \item In \Sref{sec:appendix:sft_data} (referred by \Sref{sec:method:fine-tuning} \& \ref{sec:experiment:sft}), we provide more details of SFT data we used in our experiments.
\end{itemize}

\noindent\makebox[\linewidth]{\rule{\linewidth}{1pt}}

\section{Broader Impact}
\label{sec:appendix:broader_impact}
In this paper, we introduce \datasetname{}, a new dataset with multimodal multi-grained concept annotations for MLLMs.
We first collect, pre-process and complement four public large-scale human-annotated object detection datasets.
With a well-designed structured template, we transform these datasets into our \datasetname{} dataset to address the lack of such datasets in the MLLM field and support our exploration.
Under an autoregressive discrete framework, we explore the potential of \datasetname{} for MLLMs by evaluations and analyses on various downstream vision--language benchmarks.
We demonstrate that \datasetname{} can provide multi-grained concept annotations to help MLLMs better locate and learn concepts, thus facilitating vision--language alignment across multiple granularities simultaneously.
We will open-source the models and the code of the complete pipeline from data processing, model training to evaluation for facilitating more reproducible research.
We hope that our work can inspire more research on multi-grained concept annotations for MLLMs.

We do not anticipate a specific negative impact, but, as with any multimodal large language model (MLLM), we recommend to exercise caution. 
\datasetname{} dataset is constructed from four public widely-used object detection datasets and will not bring any potential ethical or societal issues.
We will follow the usage rules and copyright of original datasets. 
Newly generated data parts will follow the usage rules and copyrights of the corresponding open-source or closed-source models.
We will require researchers who use our code and models to follow the principles of positive AI research to avoid model abuse and negative societal impact.

\section{Limitation and Future Work}
\label{sec:appendix:limitation}

\subsection{Automatic Annotation Synthesis}
\label{sec:appendix:limitation:automatic}
Nowadays, both LLMs and MLLMs can continuously improve the performance and generalization ability by scaling up training data.
Although our \datasetname{} dataset, based on several human-annotated object detection datasets and our well-designed data pipeline, can provide high-quality multi-grained concept annotations, it is still limited by the scale of the original datasets.
A natural idea is to automatically synthesize a variety of different types of annotations for any image.
This is a promising direction for future research, which can scale up to more concepts and continuously improve MLLMs.

Recent works~\citep{peng2023kosmos,pan2023kosmosg,wang2023all,rasheed2023glamm,li2024monkey,yang2023set,li2024densefusion} have proposed to automatically synthesize various annotations for large-scale web images with external open/close source models and complicated pipelines.
\textsc{Kosmos-2}~\citep{peng2023kosmos} extracted noun phrases from image captions and detected bounding boxes from images to synthesize Grounded image--text pairs (\textsc{GrIT}) to improve the grounding capability of MLLMs.
\textsc{Kosmos-G}~\citep{pan2023kosmosg} synthesized image captions and segment object regions for images to construct compositional image generation instruction tuning data to improve the zero-shot subject-driven image generation capability of MLLMs.
All-Seeing~\citep{wang2023all} and GLaMM~\citep{rasheed2023glamm} designed complicated pipelines to combine multiple external models to synthesize various annotations for large-scale web images from SA-1B~\citep{kirillov2023segment}, including detailed image captions, object labels, object regions, question-answer pairs about objects in the image, scene graphs, and so on, to improve the image/region-level captioning and grounding capability of MLLMs.
SoM~\citep{yang2023set} synthesized semantic segmentation masks for images to improve the image segmentation and grounding capability of MLLMs.\looseness=-1

Although these works have shown that automatically synthesized annotations can improve the performance of MLLMs, cascading multiple external models and complicated pipelines may introduce noise and bias to the synthesized annotations. 
Existing works introduced complex data pipelines and costly manual reviews (especially for pre-training data exceeding 1M) to reduce noise, but their released data still contains quite a bit of hallucinations, noise, and redundancy.
Moreover, these works mainly focus on exploring the potential of automatically synthesized annotations in the multimodal comprehension or generation capability of MLLMs, and have not explored both under a unified framework at the same time.
Among them, All-Seeing and GLaMM are similar to \datasetname{} proposed in this paper.
Both of them synthesize various annotations for images, including image captions, category labels and object regions provided in \datasetname{}. 
They don't provide label descriptions but provide visual question--answering pairs or scene graphs, or other annotations.
However, the final data forms of them are similar to the related works in traditional VLMs, where each image is paired with isolated different types of annotations.
They need to introduce additional components and loss functions to utilize different annotations, such as object region annotations, to optimize the model's comprehension capability at different granularities separately.
Besides, their experiments mainly explore the object recognition and segmentation, image/region-level captioning and grounding capability of MLLMs, which are actually downstream tasks that obviously benefit from their synthesized/annotated different types of annotations.

In contrast, in this paper, from the perspective of concept annotations, we explore the potential of multi-grained concept annotations for MLLMs for the first time, in both multimodal comprehension and multimodal generation.
We mainly focus on general multimodal benchmarks instead of specific benchmarks that fine-grained annotations are good at (\eg{} object recognition and grounding benchmarks).
We want to evaluate and analyse the potential of multi-grained concept annotations for MLLMs in a more comprehensive, general and fair way, by comparing and collaborating with widely-used coarse-grained image--caption data.

We believe that the automatic synthesis of multi-grained concept annotations for any image is a promising direction for future research.
In addition, incorporating more different types of annotations, including not only various annotations found in existing works, but also annotations for text-rich images and table/chart images, would be valuable for future research.
We have explored the potential of multi-grained concept annotations for MLLMs in both multimodal comprehension and generation, including image captioning, text-to-image generation, visual question answering, multi-choice benchmarks, and image editing.
It is interesting to explore more benchmarks on object recognition and segmentation, grounding capability and other vision--language tasks in the future.
Finally, in this work, we focus on concrete concepts, especially objects, attributes of objects, and relationships between objects.
It is very interesting to explore more concrete concepts and also abstract concepts, such as emotions, events, and so on, in the future.\looseness=-1

\subsection{Collaboration Strategy of \datasetname{} and Image--Caption Data}
\label{sec:appendix:limitation:collaboration}

In \Sref{sec:experiment:scalable}, we investigate comparison and collaboration between \datasetname{} and coarse-grained image--caption dataset \icdatasetname{} on $4$ benchmarks for multimodal comprehension and generation.
Simply joint training with \datasetname{} and \icdatasetname{} cannot bring performance improvements compared to training with \datasetname{} only, and even leads to obvious performance degradation on COCO, NoCaps and VIST.
Considering that \icdatasetname{} is $\sim 15$ times larger than \datasetname{}, we also try to balance the two datasets by using $2 \sim 4$ times of \datasetname{}\footnote{We follow the advice from \citet{muennighoff2024scaling} to repeat data no more than $4$ times.}.
However, repeating \datasetname{}, \eg{} $4$\datasetname{}$+$\icdatasetname{}, introduces only minor performance fluctuations and results in lower average performance.

Therefore, we follow the curriculum learning strategy~\citep{mccann2019the} to train them in different orders and achieves significantly better average performance by training on \icdatasetname{} first and then on \datasetname{}.
This is consistent with recent findings~\citep{hu2024minicpm,li2024llavanext-ablations} that training with high-quality data late in the pre-training phase leads to better performance. 
Moreover, considering that the noise in \icdatasetname{} still cause a slight performance drop on VIST (\textcolor{indexcolor}{\{$2,5$\}}-th rows in \Tref{tab:scalable}), and jointly training on \datasetname{} and \icdatasetname{} outperforms \icdatasetname{} on both tasks (\textcolor{indexcolor}{\{$1,3$\}}-th rows in \Tref{tab:scalable}). 
Hence, we first jointly train on \datasetname{} and \icdatasetname{} to alleviate the effect of noise in \icdatasetname{}, and then on \datasetname{}, eventually achieving the best average performance (\textcolor{indexcolor}{$6$}-th row in \Tref{tab:scalable}).
Similarly, further increasing \datasetname{} repetitions, \eg{} {$3$\datasetname{}$+$\icdatasetname{} $\rightarrow$ \datasetname{}}, only yields minor performance fluctuations and leads to noticeable average performance drops.
We speculate that repetition of \datasetname{} leads to overfitting of \datasetname{} and insufficient learning of IC, resulting in performance declines and fluctuation. 
We believe that the collaboration strategy of \datasetname{} and image--caption data is an interesting research question about data mixing~\citep{liu2024regmix}, data repetition~\citep{muennighoff2024scaling} and curriculum learning~\citep{soviany2022curriculum}.
Automatic annotation synthesis for large-scale image--caption data we discussed in \Apref{sec:appendix:limitation:automatic} is also a promising solution to scale up to more concepts and continuously improve MLLMs without suffering from the problem of data duplication.

\subsection{A MLLM Framework for Multimodal Comprehension and Generation}
In this work, we standardize a generative vision--language framework based on existing MLLMs, which consists of a LLM and several visual modules. We follow previous works~\citep{ge2023planting,zhu2023vl} to freeze all visual modules and most of the LLM parameters during training to greatly improve efficiency.
Although the performance of the framework is competitive on various vision--language benchmarks,  it is limited by the frozen visual modules and parameter-efficient fine-tuning of the LLM.
In the future, we will explore full-parameter fine-tuning of the LLM and visual modules to further improve the performance of the framework.
Besides, optimizing the MLLM framework by improving image resolution~\citep{liu2024improved,li2024monkey,liu2024llavanext}, introducing more visual experts or different levels of unimodal representations into MLLMs~\citep{xu2023bridgetower,tong2024cambrian,xu-etal-2023-managertower}, and so on, are also promising directions to further improve the performance and extending the capability of the framework to more multimodal tasks, \eg{} multimodal in-context learning~\citep{zhao2024mmicl,qin2024factors} and multimodal chain-of-thought reasoning~\citep{zhang2024multimodal,chen-etal-2024-m3cot}.

\begin{figure}[t]
	\centering
	\includegraphics[width=0.95\textwidth]{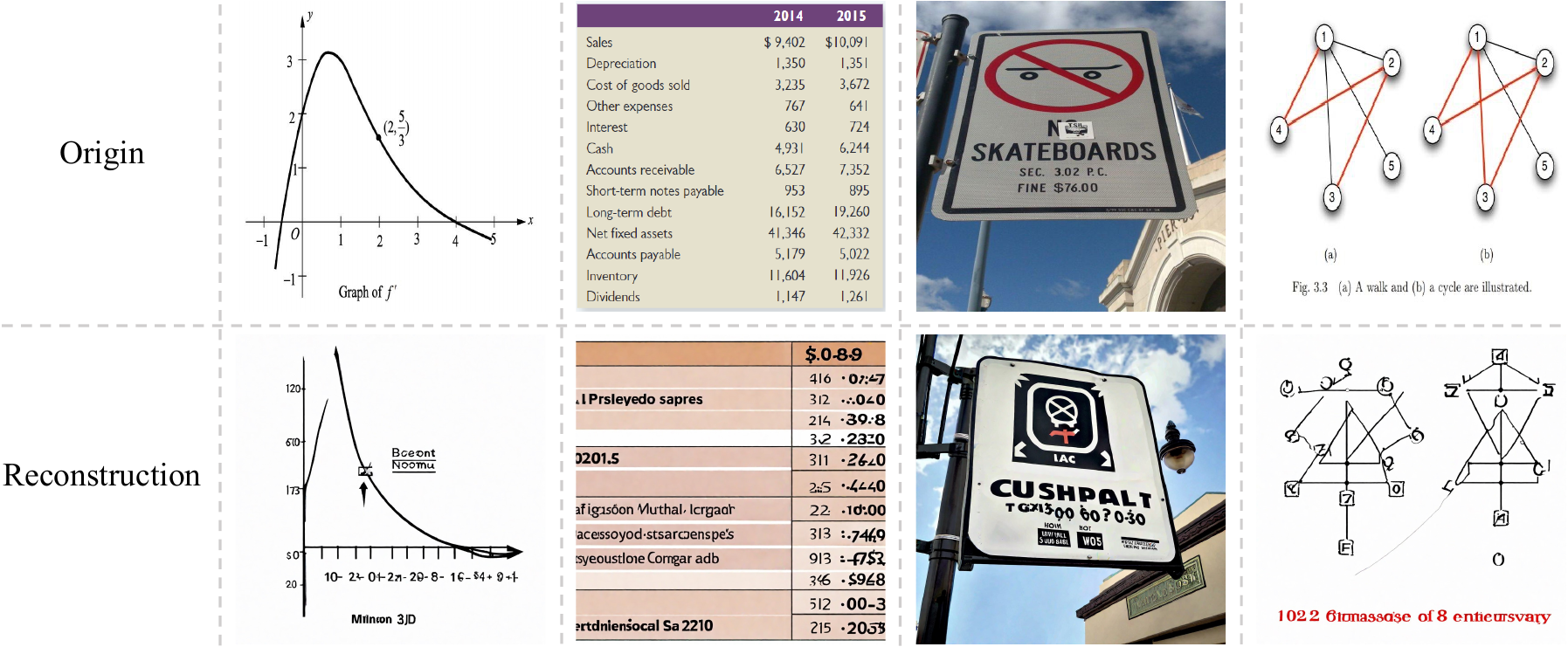}
	\caption{
        Visualization of the original and reconstructed images by the visual tokenizer inherited from LaVIT~\citep{jin2023unified} used in this paper.
        Text-rich and chart images cannot be well reconstructed.
    }
	\label{fig:app:tokenizer}
\end{figure}

\paragraph{Visual Tokenizer.}
Unlike traditional visual tokenizers like VQ-VAE~\citep{van2017neural} that reconstruct from original image pixels, SEED~\citep{ge2023planting} followed \textsc{BEiT} v2~\citep{peng2022beitv2} to train the visual tokenizer by reconstructing from visual embeddings with high-level semantics, and found that the latter strategy can better capture high-level semantics instead of low-level image details in the former strategy, which is more effective for multimodal comprehension. 
As the concurrent work of SEED, LaVIT also found that high-level semantics matters for downstream tasks, and further supported dynamic sequence length varying from the image. 
Compare SEED and LaVIT, we take the visual tokenizer from LaVIT, which can better preserve image details and semantic information.
However, no matter which visual tokenizer is used, the vector quantization process is a trade-off between the cost/difficulty of learning and the capacity of representing visual information. For example, as shown in \Fref{fig:app:tokenizer}, the visual tokenizer of LaVIT we used in this work cannot well reconstruct the original image by the visual decoder and diffusion model, especially text-rich images and chart images.
Chameleon~\citep{team2024chameleon} also found such limitations of traditional visual tokenizers (reconstructed from original image pixels) on heavy OCR-related tasks.

We also conduct experiments on downstream VL benchmarks containing text-rich images and table/chart images to further validate the limitation of the visual tokenizer. 
Our \methodnamewithic{} achieves $32.0$ accuracy on MMMU~\citep{yue2023mmmu} validation set, where the human performance, the best model performance and random performance are $88.6,59.4,22.1$, respectively; and achieves $26.0$ accuracy on MathVista~\citep{lu2024mathvista} testmini set, where the human performance, the best model performance and random performance are $60.3,63.9,17.9$, respectively.
In the future, we will explore more advanced visual tokenizers to better capture high-level semantics and low-level image details for multimodal comprehension on general, text-rich and table/chart images.

\paragraph{LLM Part.}
We conduct experiments with LLaMA-2-7B as the LLM part in our framework.
LLaMA-2-7B is widely used in the LLM and MLLM fields due to its good performance and generalizability, ensuring the reliability, reproducibility and scalability of our experimental results.
Especially, most of the existing SOTA MLLMs (e.g., EMU~\citep{sun2023generative}, LaVIT~\citep{jin2023unified}, SEED-LLaMA~\citep{ge2023making}, DreamLLM~\citep{dong2024dreamllm}, VL-GPT~\citep{zhu2023vl}) all use the LLaMA series or its variants to initialize the LLM part of their models.
Since our research aims to explore the potential of \datasetname{} as a new data paradigm for MLLMs, we follow this common practice to use LLaMA-2-7B in our framework.
It will be interesting to explore more advanced LLMs in the future to further improve the performance of our framework, considering recent findings that the model size scaling of LLM is more effective than image encoder in yielding improved performance~\citep{li2024llavanext-ablations}.

\section{More Experimental Results and Analysis}
\label{sec:appendix:experiment}

\subsection{Concept Overlap}
\label{sec:appendix:concept_overlap}

\begin{figure}[t]
	\centering
	\includegraphics[width=0.95\textwidth]{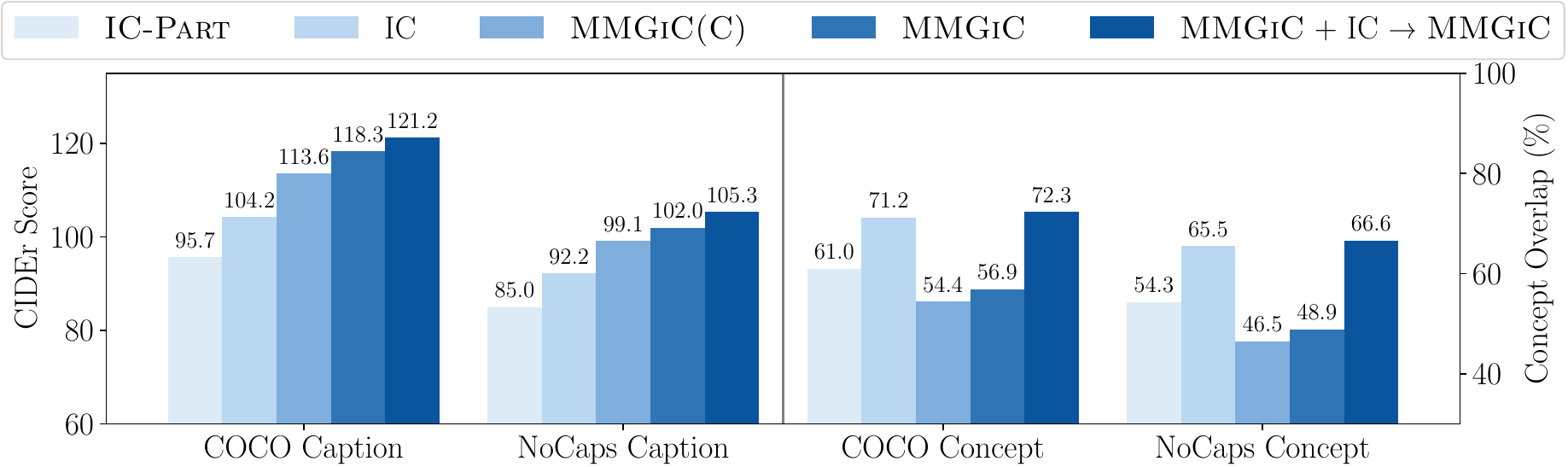}
	\caption{
        Comparison of performance (CIDEr Score) and concept overlap (\%) for different training data settings on image captioning datasets COCO and NoCaps.
    }
	\label{fig:app:concept_overlap}
\end{figure}

In \Sref{sec:experiment:sft}, we find that \datasetname{} and \icdatasetname{} have their own strengths in depth and breadth of concept representation.
To further verify our findings, inspired by \textsc{K-Lite}~\citep{shen2022k}, we investigate the concept overlap between different training data settings and downstream datasets in terms of dataset statistics.
The concept overlap is computed as the percentage of concepts in a downstream dataset that are covered by the training data.
We select two image captioning datasets, COCO and NoCaps, and directly take noun chunks extracted from their ground-truth captions as concepts.
We select the \textcolor{indexcolor}{\{$0,1,2,6$\}}-th data recipes in \Tref{tab:scalable} and also include \datasetnamecaptiononly{} from \textcolor{indexcolor}{\{$0$}-th data recipe in \Tref{tab:data_recipes} for better comparison.
As shown in \Fref{fig:app:concept_overlap}, we can see that: 
\begin{itemize}[leftmargin=2em, topsep=0em, itemsep=0em]
    \item Comparing \icdatasetnamepart{} and \icdatasetname{}: with the number of image--caption pairs increasing from less than $4$M to almost $52$M, the concept overlap increases significantly ($10.2$\% and $11.2$\%) and the performance also improves obviously ($8.5$ and $7.2$ points) on both datasets.
    \item Comparing \datasetnamecaptiononly{} and \datasetname{}: the performance improvement ($4.7$ and $2.9$ points) may not come from improved concept overlap (only $2.5$\% and $2.4$\%). 
	This provides side evidence that fine-grained concept annotations in \datasetname{} can integrate and complement coarse-grained concept annotations in \datasetnamecaptiononly{}, to help MLLMs learn concepts deeply and thus improve performance.
    \item Comparing \datasetname{} and {\datasetname{}$+$\icdatasetname{} $\rightarrow$ \datasetname{}}: \datasetname{} can collaborate with \icdatasetname{} to achieve higher concept overlap ($15.4$\% and $17.7$\%).
    Appropriate curriculum learning strategies can effectively integrate the strengths of both in terms of depth and breadth of concept representation, thus improving performance on both datasets ($2.9$ and $3.3$ points).
    \item Comparing \icdatasetnamepart{}, \icdatasetname{} and \datasetnamecaptiononly{}: although they are all image--caption pairs and captions are all synthesized by the same pipeline (\Apref{sec:appendix_data:caption_synthesis}), \datasetnamecaptiononly{} can achieve better performance on both datasets with lower concept overlap and the same number of image--caption pairs as \icdatasetnamepart{}. 
	We attribute this to the fact that the images in \datasetnamecaptiononly{} have higher quality, \ie{} contain more concepts and are more complex than \icdatasetnamepart{} and \icdatasetname{} collected from the web.
    Specially, the number of unique noun chunks in the three training data are: $1.7$M, $10.9$M, and $0.4$M, respectively, while the average number of unique noun chunks per image are: $3.0$, $3.0$, and $8.8$, respectively.
	This is consistent with recent findings~\citep{dai2024nvlm,li2024llavanext-ablations} that the impact of data quality is greater than data scale.
\end{itemize}

\subsection{Analysis on Image Editing}
\label{sec:appendix:image_editing} 
To further investigate how \datasetname{} can better help MLLMs utilize concepts to align vision and language and improve the capability of generating concepts we conduct both qualitative and quantitative analyses on image editing task.

\begin{table}[t]
    \centering
    \tablestyle{6pt}{1.1}
    \caption{
        Zero-shot evaluation on the image editing benchmark DreamBench~\citep{ruiz2023dreambooth} after SFT.
        DINO: cosine similarity between generated and real images via DINO~\citep{caron2021emerging}.
        The best results are \textbf{bold} and the second-best are \underline{underlined}.
    }
    \vspace{1em}
    \label{tab:sft_generation_2}
    \scalebox{0.9}{
        \begin{tabular}{c|l|ccc}
            \toprule
            & \multirow{2}{*}{Model} & \multicolumn{3}{c}{DreamBench} \\
            &  & CLIP-T & CLIP-I & DINO \\
            \midrule
            \multicolumn{5}{l}{\demph{\it{SOTA MLLMs as upper bound references, not comparable}}}\\
            
            \midrule
            \multirow{2}{*}{\textcolor{indexcolor}{a}} & \demph{\textsc{Kosmos-G}}~\citep{pan2023kosmosg} & \demph{28.70} & \demph{84.70} & \demph{69.40} \\
            & \demph{Emu2-Gen-37B}~\citep{sun2023generative2} & \demph{28.70} & \demph{85.00} & \demph{76.60} \\
            \hdashline
            {\textcolor{indexcolor}{b}} & \demph{SEED-LLaMA-I-8B}~\citep{ge2023making} & \demph{27.06} & \demph{79.27} & \demph{54.42} \\
            \midrule
            \multicolumn{5}{l}{\it{Our comparable baselines}}\\
            \midrule
            \textcolor{indexcolor}{0} & \methodnameic{} & 27.50 & \underline{82.92} & \underline{67.41} \\
            \textcolor{indexcolor}{1} & \methodname{} & \textbf{27.69} & 82.62 & 67.36 \\
            \textcolor{indexcolor}{2} & \methodnamewithic{} & \underline{27.62} & \textbf{83.06} & \textbf{68.30} \\
            \bottomrule
        \end{tabular}
    }
\end{table}

Although \datasetname{} does not contain any samples about image editing, with our structured template, each sample of \datasetname{} can be seen as an image--text interleaved document with an image, multiple cropped regions from the image, textual annotations, and also template instruction text.
Such design can help MLLMs better understand each region in the image, as well as the relationships within and between regions, thus better learning and generating concepts, and aligning images and instructions.
We hypothesize that this may help MLLMs learn more diverse image generation abilities.

\paragraph{Quantitative Analysis.}
To verify this, we include about $0.32$M image editing instruction samples from Instructpix2pix~\citep{brooks2023instructpix2pix} and MagicBrush~\citep{zhang2024magicbrush} during the SFT stage, and evaluate the performance of our baselines on the image editing benchmark DreamBench~\citep{ruiz2023dreambooth}, as shown in \Tref{tab:sft_generation_2}.
Compared \datasetname{} with \icdatasetname{}, \icdatasetname{} performs better on image (subject) fidelity (CLIP-I and DINO), and \datasetname{} performs better on text fidelity (CLIP-T, i.e., adherence to editing instructions). 
We attribute this to the fact that the structured template and rich multimodal annotations in \datasetname{} can help MLLMs better understand and follow editing instructions, while the massive image--caption pairs and higher concept breadth in \icdatasetname{} can help MLLMs better generate and edit images.
Furthermore, we also find that the collaboration of \datasetname{} and \icdatasetname{} combines the advantages of both, and achieves better average performance on both image and text fidelity.
Compared with existing SOTA MLLMs, \datasetnamewithic{} not only significantly outperforms SEED-LLaMA~\citep{ge2023making} (using more pre-training data, SFT data, and full-param fine-tuning), but also achieves comparable performance to generation-only MLLMs (\textsc{Kosmos-G}~\citep{pan2023kosmosg} and Emu2-Gen-37B~\citep{sun2023generative2}) trained with far more image generation and editing data.

\paragraph{Qualitative Analysis.}
We also show some qualitative examples of image editing from MLLMs pre-trained with \datasetname{} in \Fref{fig:case_study_2} \figright{}.
The top two examples show that \datasetname{} can precisely understand editing instructions and perform appropriate editing, while the bottom example shows that \datasetname{} can synthesize image precisely based on the image--text interleaved sequences.
Considering that \datasetname{} dataset does not contain any samples about above two emergent abilities of image editing and multimodal in-context image synthesis, 
we hypothesize that integrating multi-grained concept annotations into image--text interleaved documents through our structured template can help MLLMs better understand and locate objects in the image, to more accurately edit the objects in the image, and synthesize images based on the image--text interleaved sequences.

\subsection{Self-Generated Multi-Grained Concept Annotations in Evaluation}
\label{sec:appendix:self_generated_seed}

\begin{figure}[t]
	\centering
    \includegraphics[width=0.5\textwidth]{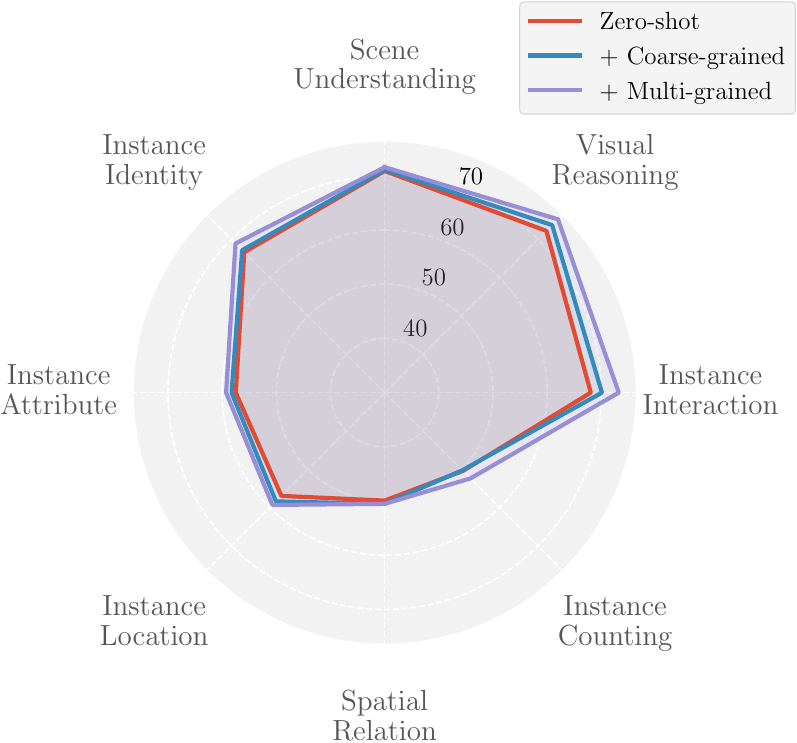}
	\caption{
        Detailed evaluation of \methodnamewithic{} with different evaluation strategies on $8$ dimensions of SEED-Bench-IMG.
	}
	\label{fig:app:self_generated_seed}
\end{figure}

Recent works~\citep{sun2023generative,wu2023role,mitra2023compositional} have proposed to ask MLLMs to answer questions with the help of image annotations (\eg{} captions, or scene graphs) generated by themselves or external models during evaluation.
Inspired by these works, we naturally ask MLLMs trained with \datasetname{} to self-generate coarse- and fine-grained concept annotations for images, like annotations provided in \datasetname{}.
Considering the inference efficiency and avoiding noise, we only generate category labels as fine-grained concept annotations.
Specifically, we explore different evaluation strategies on SEED-Bench-IMG in \fref{fig:app:self_generated_seed}.
After training with \datasetname{}, fine-grained concept annotations can help MLLMs generate better coarse-grained image captions to improve performance, especially on the ``\textit{Instance Location, Instance Interaction, Visual Reasoning}'' dimensions.
Furthermore, self-generate multi-grained concept annotations can integrate the advantages of both granularities, achieving $1.68$ points improvement in overall accuracy compared with the zero-shot evaluation, especially on the ``\textit{Instance Identity, Instance Counting, Instance Interaction, Visual Reasoning}'' dimensions.
This demonstrates that multi-grained concept annotations can help MLLMs better understand and reason about concepts in the image during evaluation.

\subsection{Qualitative Analysis of Different-Grained Concept Annotations}
\label{sec:appendix:seed_case_study}

\begin{figure}[!hp]
	\centering
	\includegraphics[width=\textwidth]{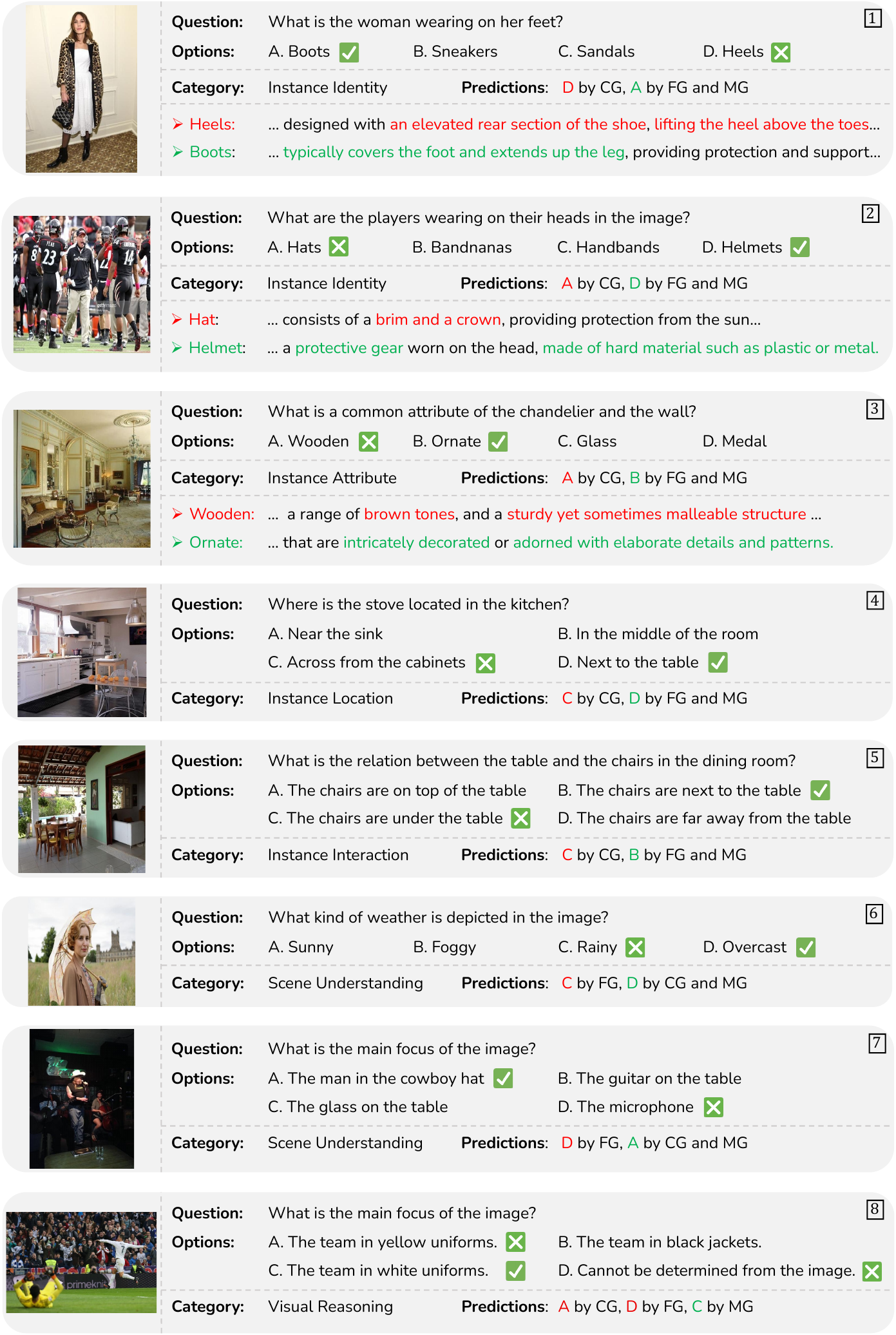}
	\caption{
		Case study of \methodname{} trained with different-grained concept annotations from \datasetname{} on SEED-Bench-IMG.
		CG, FG, and MG denote MLLMs trained with coarse-, fine-, and multi-grained concept annotations from \datasetname{}, respectively.
		✅ denote the ground truth; ❎ denote incorrect prediction(s).
        The bottom right of the first three examples show the associated label--description pairs from \datasetname{}.
	}
	\label{fig:app:seed_case_study}
\end{figure}

We have provided the quantitative results of \methodname{} with different-grained concept annotations from \datasetname{} on each evaluation dimension of SEED-Bench-IMG in \Sref{sec:experiment:meso}.
Here, we further provide corresponding qualitative analysis to better analyse the advantages of coarse-, fine-, and multi-grained concept annotations in \datasetname{} for MLLMs.
As shown in \Fref{fig:app:seed_case_study}, we can see that:
\begin{itemize}[leftmargin=2em, topsep=0em, itemsep=0em]
	\item Examples {\scriptsize $\boxed{1}$} and {\scriptsize $\boxed{2}$} from the ``Instance Identity'' dimension: 
	while CG can provide a holistic description of the whole image by the image caption, MLLMs trained with CG cannot distinguish between ``Boots'' and ``Heels'' in {\scriptsize $\boxed{1}$}, and ``Helmets'' and ``Hats'' in {\scriptsize $\boxed{2}$}.
	However, FG can provide visual details ``covers the foot and extends up the leg'' and relevant knowledge ``protective gear'' by label--description pairs, which can help MLLMs better understand these concepts. 
	Object regions can further help MLLMs better locate and learn these concepts.
	Therefore, MLLMs trained with FG or MG can capture these local details and identify concepts correctly.
	\item Example {\scriptsize $\boxed{3}$} from the ``Instance Attribute'' dimension:
	after locating the ``chandelier'' and ``wall'' in the image, MLLMs trained with FG or MG can recognize the shared attribute ``Ornate'' correctly by the visual details ``decorated with elaborate details and patterns''.
	\item Example {\scriptsize $\boxed{4}$} from the ``Instance Location'' dimension:
	MLLMs trained with FG or MG can identify and locate the ``stove'' in the image, and then correctly analyse its surrounding objects and the spatial relationships between them.
	\item Example {\scriptsize $\boxed{5}$} from the ``Instance Interaction'' dimension:
	the interaction between the table and the chairs requires the model to correctly understand the local details of their positions and spatial relationships.
	The image shows that the table is placed around the chairs, rather than completely under the table.
	Such details can be correctly captured by MLLMs trained with FG or MG.
	\item Example {\scriptsize $\boxed{6}$} from the ``Scene Understanding'' dimension:
	while the lady in the image is holding an ``umbrella'', her surroundings show that the sky is covered with clouds, but it is not raining, which implies that the weather is ``overcast''.
	MLLMs trained with FG seem to be too focused on the ``umbrella'' to the overall scene of the image, while MLLMs trained with CG or MG can predict the correct answer by better understanding the global context of the image.
	\item Example {\scriptsize $\boxed{7}$} from the ``Scene Understanding'' dimension:
	similarly, MLLMs trained with FG pay too much attention to the ``microphone'' in the image.
	In fact, ``the man in the cowboy hat'' is standing prominently in the scene. 
	The lighting, his position, and his stance all draw attention to him as the main subject.
	\item Example {\scriptsize $\boxed{8}$} from the ``Visual Reasoning'' dimension:
	the image shows a football match, where a player in the ``white uniform'' is celebrating, and a player in the ``yellow uniform'' is lying on the ground.
	The audience in the background is cheering for the player in the ``white uniform'' and taking pictures of him.
	Since the image contains rich objects and details, MLLMs trained with FG have difficulty confirming the main focus of the image, while MLLMs trained with CG believe that the visually more prominent ``yellow'' player is the main focus.
	However, \datasetname{} with our structured template can help MLLMs combine the advantages of CG and FG, make different-grained concept annotations complement each other, and better understand and reason about the image from both global scene and local details.
	Hence, MLLMs trained with MG can correctly identify the ``white'' player as the main focus of the image.
\end{itemize}

\subsection{\datasetname{} as Instruction Fine-Tuning Data}

\begin{table}[t]
    \centering
    \tablestyle{2pt}{1.1}
    \caption{
        Zero-shot evaluation on multimodal comprehension and generation benchmarks after SFT.
        MMB: MMBench; SQA$^\mathrm{I}$: ScienceQA-IMG; SEED$^\mathrm{I}$: SEED-Bench-IMG.
        The best results are \textbf{bold}.
    }
    \vspace{1em}
    \label{tab:sft_mgc}
    \scalebox{0.71}{
        \begin{tabular}{c|l|cc|ccc|ccccc|ccc|cc}
            \toprule
            & \multirow{2}{*}{Model} & \multicolumn{2}{c|}{Image Captioning} & \multicolumn{3}{c|}{VQA} & \multicolumn{5}{c|}{Multi-Choice Benchmark} & \multicolumn{3}{c|}{MS-COCO-30K} & \multicolumn{2}{c}{VIST} \\
            &  & COCO & NoCaps & VQAv2 & GQA & VizWiz & POPE & MME & MMB & SQA$^\mathrm{I}$ & SEED$^\mathrm{I}$ & {FID} ($\downarrow$) & CLIP-T & CLIP-I & {FID} ($\downarrow$) & CLIP-I \\
            \midrule
            \textcolor{indexcolor}{0} & \methodnameic{} & 108.13 & { }{ }92.71 & 70.28 & 56.02 & 52.62 & 81.14 & 1646.71 & 59.54 & 65.94 & 58.41 & 8.11 & 30.90 & 70.72 & 38.19 & 65.37 \\
            \textcolor{indexcolor}{1} & {\ }+ \datasetname{}  & \textbf{119.03} & \textbf{105.49} & \textbf{70.33} & \textbf{56.67} & \textbf{52.63} & \textbf{82.31} & \textbf{1656.29} & \textbf{59.64} & \textbf{66.04} & \textbf{59.60} & \textbf{7.65} & \textbf{31.26} & \textbf{71.66} & \textbf{34.51} & \textbf{66.61} \\
            \bottomrule
        \end{tabular}
    }
\end{table}

If we regard the template text of our structured template as multiple instructions and multi-grained concept annotations as responses to the instructions, then \datasetname{} can be regarded as image--text interleaved instruction fine-tuning data.
It can be directly used in the SFT stage of existing MLLMs pre-trained with image--caption data only.
As shown in \Tref{tab:sft_mgc}, the \textcolor{indexcolor}{$0$}-th row corresponds to the performance of \methodnameic{}, an MLLM that pre-trained with only image--caption data, \icdatasetname{}, and fine-tuned with our default SFT data in \Sref{sec:method:fine-tuning} ($1.21$M samples from public instruction datasets and $1$M samples from an aesthetic image--caption dataset).
To explore the potential of \datasetname{} directly as instruction fine-tuning data, we add $1$M samples from \datasetname{} to the SFT stage of \methodnameic{}.
\methodnameic{}+\datasetname{} in the \textcolor{indexcolor}{$1$}-th row achieves better performance on all benchmarks, especially on the benchmarks that deeply inspect common concrete concepts, \eg{} COCO, NoCaps, POPE, SEED-Bench, MS-COCO-30K, and VIST.
This demonstrates \datasetname{} can be used not only as pre-training data, but also as instruction fine-tuning data to help MLLMs learn concepts better, which further ensure the generality of \datasetname{} across different MLLM frameworks.

\subsection{Impact of the Nature of Image--Text Interleaved}

\begin{table}[t]
    \centering
    \tablestyle{6pt}{1.2}
    \caption{
        Zero-shot evaluation on multimodal generation benchmarks.
        For each block, the best result are \textbf{bold}.
    }
    \vspace{1em}
    \label{tab:MMC4}
    \scalebox{0.9}{
        \begin{tabular}{c|l|ccc|cc}
            \toprule
            & \multirow{2}{*}{{Training Data}} &  \multicolumn{3}{c}{MS-COCO-30K} & 
            \multicolumn{2}{c}{VIST} \\
            \cmidrule{3-7}
            & {} & FID ($\downarrow$) & CLIP-T & CLIP-I & FID ($\downarrow$)& CLIP-I \\
            \midrule
            \textcolor{indexcolor}{0} & \datasetnamecaptiononly{} & { }{ }\textbf{7.20} & 30.81 & 71.62 & { }{ }67.61 & 62.22 \\
            \textcolor{indexcolor}{1} & \datasetname{} & { }{ }7.36 & \textbf{31.57} & \textbf{72.24} & { }{ }\textbf{35.33} & \textbf{66.10} \\
            \hdashline
            \textcolor{indexcolor}{2} & MMC4-Pairs~\citep{zhu2023multimodal} & \textbf{16.55} & \textbf{29.32} & \textbf{69.13} & 101.68 & 60.43 \\
			\textcolor{indexcolor}{3} & MMC4~\citep{zhu2023multimodal} & 17.81 & 28.68 & 68.48 &  { }{ }\textbf{40.31} & \textbf{66.00} \\
            \bottomrule
        \end{tabular}
    }
\end{table}

Unlike MS-COCO-30K~\citep{chen2015microsoft} that generates a new image based on a single caption, VIST~\citep{huang-etal-2016-visual} (Visual Storytelling) is a multimodal generation benchmark that requires MLLMs to generate a new image based on interleaved image--caption context within the same story.
Compared with image--caption data, \datasetname{} has the nature of image--text interleaved, which may help MLLMs better understand the image--text context in VIST.
Hence, in \Tref{tab:MMC4}, we compare the performance of \datasetnamecaptiononly{} and \datasetname{} (\textcolor{indexcolor}{\{$0,3$\}}-th data recipes in \Tref{tab:data_recipes}) on both MS-COCO-30K for simple text-to-image generation and VIST for in-context image synthesis.
In addition, we also pre-train MLLMs with MMC4\footnote{We use a subset of MMC4, \ie{} MMC4-core-fewer-faces, which has a moderate sample size of $22.4$M images and $5.5$M image--text interleaved training samples, which is similar to the sample size of our \datasetname{}. Following the processing pipeline of MM-Interleaved~\citep{tian2024mminterleaved}, we process MMC4, and then convert each data sample into a discrete token sequence similar to our \datasetname{}.}~\citep{zhu2023multimodal} and MMC4-Pairs (split each MMC4 document into multiple individual image--text pairs) to ablate the impact of multi-grained concept annotations.

Comparing the top two rows in \Tref{tab:MMC4}, after introducing the nature of image--text interleaved into our \datasetname{}, we find that most metrics increase, especially on VIST.
However, for MMC4-Pairs and MMC4 in the bottom two rows, in-context image synthesis performance remains increases but text-to-image generation performance decreases.
The results demonstrate that the nature of image--text interleaved can greatly benefit in-context image synthesis but may not be beneficial for simple text-to-image generation.
We hypothesize that the multi-grained concept annotations in our \datasetname{} can work with the nature of image--text interleaved to help MLLMs better learn concepts and utilize concepts to align vision and language, thus improving performance on both benchmarks.

\subsection{Evaluation on Q-Bench and HallusionBench}

\begin{table}[t]
    \centering
    \tablestyle{6pt}{1.1}
    \caption{
        Zero-shot evaluation on Q-Bench and HallusionBench after SFT.
    }
    \label{tab:qbench_and_hallusion}
    \vspace{1em}
    \scalebox{0.85}{
        \begin{tabular}{c|l|c|c}
            \toprule
            & Model & Q-Bench & HallusionBench \\
            \midrule
            \textcolor{indexcolor}{a} & \demph{LLaVA-1.5-7B}~\citep{liu2023improvedllava} & \demph{58.70} & \demph{27.60} \\
            \textcolor{indexcolor}{b} & \demph{SEED-LLaMA-I-8B}~\citep{ge2023making} & \demph{47.22} & \demph{32.26}  \\
            \midrule
            \textcolor{indexcolor}{0} & \methodnameic{} & 56.66 & 32.58 \\
            \textcolor{indexcolor}{1} & \methodname{} & 57.59 & 30.37 \\
            \textcolor{indexcolor}{2} & \methodnamewithic{} & \textbf{58.26} & \textbf{33.39}\\

            \bottomrule
        \end{tabular}
    }
\end{table}

In this paper, we mainly focus on benchmarks that inspect the comprehension and generation capabilities of MLLMs on concrete concepts.
Here, we further evaluate on two new benchmarks, Q-Bench~\citep{wu2023q} and HallusionBench~\citep{liu2023hallusionbench}.
They focus on low-level vision abilities (clearness, brightness, etc.), visual illusion and language hallucination (especially abstract concepts), respectively.
As shown in \Tref{tab:qbench_and_hallusion}, we can see that for the low level vision abilities in Q-Bench, although \datasetname{} mainly focus on objects, attributes and relations in images, it still helps MLLM achieve better performance than \icdatasetname{}.
We attribute this to the fact that concepts like clearness and brightness are concrete concepts that can also benefit from multi-grained concept annotations (\eg{} captions, label descriptions
) in \datasetname{}.
As for HallusionBench, \icdatasetname{} achieves significantly better performance than \datasetname{} since it contains more concepts, especially abstract concepts.
Most importantly, the combination of \datasetname{} and \icdatasetname{} can achieve the best performance on both benchmarks, which demonstrates that their combination can effectively integrate the strengths of both in terms of depth and breadth of concept representation, thus improving performance on both benchmarks.\looseness=-1

\subsection{Text-Only Performance}

\begin{table}[t]
    \centering
    \label{tab:text_only}
    \tablestyle{6pt}{1.2}
    \caption{
        Evaluation on the MMLU~\citep{hendrycks2021measuring} benchmark with standard 5-shot setting and Accuracy as the metric.
    }
    \vspace{1em}
    \scalebox{0.9}{
        \begin{tabular}{c|l|c}

            \toprule
            & \multicolumn{1}{c|}{Training Setting} &  
            MMLU \\
            \midrule
            \textcolor{indexcolor}{0} & LLaMA-2-7B-base & 46.22 \\
            \textcolor{indexcolor}{1} & \quad + \datasetname{} & 29.98 \\
            \textcolor{indexcolor}{2} & \quad + \datasetname{} + SFT & \textbf{46.50} \\
            \bottomrule
        \end{tabular}
    }
\end{table}

We follow common practice~\citep{sun2023generative,ge2023making,jin2023unified} to evaluate the zero-shot performance of MLLMs on multimodal comprehension and generation benchmarks.
Notable, some recent works also explored the text-only performance of MLLMs.
For example, LaVIT~\citep{jin2023unified} and MM1~\citep{mckinzie2024mm1} found that training MLLMs only with multimodal data could result in a significant degradation of the text-only performance. 
To address this issue, they mixed a substantial amount of text-only data, \eg{} $66$\% text-only and $33$\% multimodal data, thus preserving the original text capability of the model.
Besides, VILA~\citep{lin2023vila} observed that while multimodal pre-training will hurt text-only performance, by simply incorporating some high-quality text-only instruction data during the SFT stage, the model can achieve similar text-only performance to the initial LLM.

To improve efficiency, we follow the approach of VILA to incorporate text-only instruction data, \ie{} ShareGPT\citep{sharegpt} and Alpaca~\citep{alpaca}, into our default SFT data in \Sref{sec:method:fine-tuning}.
We also conduct experiments on the de-facto text-only comprehensive multi-choice benchmark, MMLU~\citep{hendrycks2021measuring}, to quickly evaluate the text-only performance of our models.
As shown in \Tref{tab:text_only}, we can see that there is indeed a significant performance drop after pre-training with \datasetname{}.
However, after SFT, the MMLU performance of the model recovers well and is similar to the original LLM.
\section{Discussion}
\label{sec:appendix:discussion}

\subsection{Difference between Our Framework with Existing MLLM Frameworks}
\label{sec:appendix:discussion:framework}
This paper mainly focuses on MLLMs that are multimodal generalists capable of both multimodal comprehension and generation.
Existing MLLMs~\citep{sun2023generative,jin2023unified,ge2023making,zhu2023vl,sun2023generative2,yu2023scaling,lu2022unified,lu2024unified} are typically trained with an autoregressive objective, \ie{} generate the prediction of the next token in a multimodal sequence containing discrete textual tokens, and discrete visual tokens or continuous visual embeddings.
MLLMs process images in different ways.
For example, EMU~\citep{sun2023generative,sun2023generative2} and VL-GPT~\citep{zhu2023vl} transform images into continuous visual embeddings by visual encoders, and then adopt a separate regression head to predict visual embeddings.
In contrast, CM3Leon~\citep{yu2023scaling}, Unified-IO~\citep{lu2022unified,lu2024unified} and SEED-LLaMA~\citep{ge2023making} tokenize images into discrete visual tokens by visual tokenizers, and then predict visual tokens by the extended vocabulary with the unified language model head.
Technically, different training datasets, training settings (\eg{} full- or partial-param), framework, evaluation settings (\eg{} image resolution), etc., lead to non-comparable and unfair comparisons of our baselines with existing MLLMs.
Their computation (more training data and fully fine-tune the LLM params) and data resources are extremely expensive and large (well over $10$ times that of our work).
Our autoregressive discrete framework for MLLMs is based on SEED-LLaMA and LaVIT, which consists of several visual modules and a LLM with an extended vision--language vocabulary, as shown in \Fref{fig:method}.
Next, we describe the differences between our framework and these two MLLMs.
\begin{itemize}[leftmargin=2em, topsep=0em, itemsep=0em]
	\item \textbf{Visual Modules:} our visual modules are inherited from LaVIT~\citep{jin2023unified}, which consists of a visual encoder, a visual tokenizer, a visual decoder and a diffusion model.
	The differences between the two MLLMs mainly lie in the design of the latter three modules.
	The visual tokenizer of SEED-LLaMA uses the Casual Q-Former from BLIP-2~\citep{li2023blip} to compress visual representations into a fixed-length sequence of visual embeddings, while LaVIT is more flexible and designs a dynamic visual tokenizer with token selector and token merger, which can obtain a more flexible and effective sequence of visual embeddings.
	For the discrete visual token sequence predicted by the LLM, SEED-LLaMA uses an MLP as the visual decoder to compress and reconstruct it into the image embedding (1 token) and directly generate images through the frozen \href{https://huggingface.co/stabilityai/stable-diffusion-2-1-unclip}{unCLIP-SD}, while LaVIT uses multiple transformer blocks as the visual decoder to reconstruct the discrete visual token sequence into a continuous feature map by learned queries (256 tokens), and then uses the \href{https://huggingface.co/runwayml/stable-diffusion-v1-5}{conditional denoising U-Net} to progressively recover image pixels from a Gaussian noise with the feature map as the condition, which can retain more image information generated by the LLM, thus achieving better image generation results.
	We freeze all visual modules and pre-tokenize images into discrete visual token sequences, and only load vision modules to generate actual images when evaluate on downstream image generation tasks.
	This can greatly reduce the training cost and improve efficiency.
	Notably, the vision modules in our framework, though inherits from LaVIT, only performs "Stage 1 Tokenizer Training," while LaVIT's vision module also performs "Stage 2: Unified Vision--Language Pre-training."
	
	\item \textbf{Extended VL Vocabulary:} as we stated in \Sref{sec:method:framework}, SEED-LLaMA and LaVIT directly learn new visual token embeddings and initialize them with the distribution of original textual token embeddings.
	Visual tokens in the VL vocabulary correspond one-to-one with visual latent codes in the visual codebook.
	Considering the large difference in embedding dimensions between visual token embeddings in the VL vocabulary ($4096$) and visual latent codes in the visual codebook ($32$), we follow the factorized embedding parameterization in ALBERT~\citep{lan2020albert} to replace visual token embeddings $|\mathrm{V}| \times 4096$ with two smaller embedding matrices $|\mathrm{V}| \times 32, 32 \times 4096$, where $|\mathrm{V}|$ is the number of visual tokens in the VL vocabulary.
	The former matrix is directly initialized with visual latent codes in the visual codebook.
	In our preliminary experiments, we found that our initialization method of the VL vocabulary can significantly improve the performance of multimodal comprehension tasks, and has similar performance on multimodal generation tasks.
	
	\item \textbf{Training Objective:} SEED-LLaMA and our framework both treat the multimodal sequence as a discrete sequence of image--text interleaved tokens, and train the LLM with an autoregressive objective to generate predictions of the next token.
	To mitigate the loss of detailed information caused by vector quantization, LaVIT designs two different training objectives for image-to-text and text-to-image cases. For image-to-text, LaVIT is similar to EMU and uses the continuous visual features of images as the condition to predict all discrete textual tokens and calculate loss for textual tokens only; for text-to-image, LaVIT is similar to SEED-LLaMA and our framework, which tokenizes images into discrete visual token sequences and calculates loss for both textual and visual tokens.
	To pursue simplicity, efficiency, and scalability, our framework is similar to SEED-LLaMA, which tokenizes images into discrete visual token sequences, so that different types of multimodal training data can be converted into discrete sequences of image--text interleaved tokens, and trained with a unified autoregressive objective to fully learn the extended vision--language vocabulary of the LLM.

	\item \textbf{Training Data and Settings:} For our framework, we use \datasetname{} with $4$M samples and \icdatasetname{} with $52$M image--caption pairs for pre-training; we use $1.21$M multimodal instruction samples, $1$M aesthetic image--caption pairs and $1$M samples from \datasetname{} for supervised fine-tuning; during pre-training and SFT stages, our framework freezes all parameters of vision modules and most parameters of the LLM, and only tunes partial parameters of the LLM (<8\%): the extended VL vocabulary, additional LoRA modules, norm layers and a language model head layer.
	Existing MLLMs typically use more pre-training and fine-tuning data, and fully fine-tune all LLM parameters and partial vision module parameters.
	Take SEED-LLaMA as an example, uses more pre-training data ($\sim$770M vs <60M), SFT data ($\sim$145M vs <4M), and full-param fine-tuning.
	Its computation and data resources are extremely expensive and large (well over $10$ times that of our work).
	As shown in \Tref{tab:sft_understanding} and \ref{tab:sft_generation_1}, compared with SEED-LLaMA, our baseline \methodnamewithic{} requires significantly less data ($\sim$1/15) and partial-param fine-tuning (<8\%), achieving significant advantages on most benchmarks and comparable performance on COCO and VizWiz.
	Comparing with LaVIT, even though our framework does not convert images into continuous visual features to trade off lower loss of image representation and better multimodal comprehension performance, we still achieve significant advantages on three VQA benchmarks, and comparable performance on other benchmarks.
\end{itemize}

It should be emphasized that it is difficult to compare our baselines with existing MLLMs fairly, and the purpose of this paper is not to develop new frameworks, training objectives or benchmark SOTAs, but to explore the potential of multi-grained concept annotations for MLLMs.
To this end, we explore different data recipes for multi-grained concept annotations and investigate the comparison and collaboration between \datasetname{} with widely-used coarse-grained image--caption data.
Evaluations and in-depth analyses on $12$ benchmarks for multimodal comprehension and generation are conducted in both pre-training and supervised fine-tuning stages.

\paragraph{Comparison with Grounding MLLMs.}
Fine-grained object region annotations are crucial for grounding VLMs/MLLMs.
They usually convert bounding box coordinates into discrete tokens in text or visual markers in images.
Take \textsc{Kosmos-2}~\citep{peng2023kosmos} as an example, it falls into the former category, utilizing object bounding box coordinates through special discrete position tokens placed after corresponding texts in the caption.
Unlike \textsc{Kosmos-2}, \datasetname{} converts object bounding box coordinates into corresponding object regions (cropped from images), transforming pure textual object-level annotations into multimodal annotations. Structured templates integrate different granularities into unified image--text interleaved documents, enhancing MLLM's ability to locate concepts in textual annotations to corresponding regions in images.
With the unique complex context processing capabilities of MLLMs and the LM autoregressive loss, our framework can explicitly utilize multi-grained annotations in \datasetname{} to optimize multimodal alignment across multiple granularities simultaneously, improving concept understanding and generation.

\subsection{Applicability of \datasetname{} to Existing MLLMs}
\label{sec:appendix:discussion:applicability}

Our framework combines the advantages of LaVIT and SEED-LLaMA to allow efficient exploration on various multimodal tasks, using only LM autoregressive loss without extra modules and loss functions for \datasetname{}, maintaining the framework's generality and efficiency.
Compared to the current image--text pairs or image--text interleaved documents used by MLLMs, \datasetname{} can be seen as image--text interleaved documents with textual (caption, label, label description) and visual (object region) multi-grained concept annotations. They are well-integrated through our carefully designed structured templates, forming image--text interleaved documents that can be directly used by various MLLMs.
Therefore, both our \datasetname{} dataset and our framework exhibit good generality. \datasetname{} do not impose any requirements or limitations on the model architecture or loss function of MLLMs. 
It can be directly applied to various MLLMs, whether they process images as continuous vectors (\eg{} EMU, LaVIT, VL-GPT) or discrete tokens (\eg{} CM3Leon, SEED-LLaMA, Chameleon).
Our experiments on $12$ multimodal comprehension and generation benchmarks also ensure the reliability and credibility of our conclusions, making sure that \datasetname{} can be quickly applied to other MLLMs' training.
Overall, our model framework and training loss follow CM3Leon and SEED-LLaMA's spirit of unified discrete sequence modeling, offering simplicity, efficiency, and scalability. We did not make any special architectural or loss design modifications for MLLMs, ensuring good generalization and reliability of our framework, training loss, dataset, and experimental explorations.

Moreover, our focus is to explore the potential of multi-grained concept annotations for MLLMs. 
Directly fine-tuning existing SOTA MLLMs cannot fully demonstrate this point. 
Thus, we explore comparison and collaboration between \datasetname{} and \icdatasetname{} starting from the multimodal pre-training phase.
Unfortunately, all SOTA MLLMs use the LAION-5B dataset (and its subsets) for pre-training. 
However, the official data download channels were closed by the LAION team due to \href{https://laion.ai/notes/laion-maintenance/}{safety reviews} (from December 19, 2023). 
This prevents us from retraining existing SOTA MLLMs from the pre-training stage in a controlled environment to directly prove \datasetname{}'s superiority over \icdatasetname{}.
Therefore, we collected $\sim$250M image--caption pairs from other widely used and recognized datasets and selected $\sim$52M as the \icdatasetname{} dataset for our experiments.
Through the evaluation of three baselines across $12$ multimodal understanding and generation benchmarks, we demonstrate the potential of \datasetname{} for MLLMs.
Combining \datasetname{} and \icdatasetname{} can leverage their respective advantages in terms of depth and breadth of concept representation, further improving MLLM performance.

\subsection{Annotation Synthesis in \datasetname{}}
\label{sec:appendix:discussion:annotation_synthesis}

\datasetname{} is constructed from several public human-annotated object detection datasets, which covers many common concepts but currently cannot scale up to cover more concepts in the real world.
We only automatically synthesize image captions for all images with the de-facto captioning model BLIP-2~\citep{li2023blip} and ranking model CLIP~\citep{radford2021learning}, and synthesize label descriptions for all labels with the strong GPT-4~\citep{achiam2023gpt}.
For synthesized captions, we randomly sample $1000$ captions to manually check the quality, and find that the quality is acceptable.
For synthesized label descriptions, we carefully design prompt templates and human-annotated in-context examples, and manually check all synthesized label descriptions to ensure the quality.
Besides, with the help of WordNet, we check category labels, and update the polysemous labels based on the specific data samples for better differentiation, \eg{} ``batter'' $\rightarrow$ ``batter (ballplayer)'' or ``batter (cooking)''.\looseness=-1

In this paper, we want to use and check existing human-annotated annotations as much as possible, and minimize the automatic generation of annotations, thus reducing noise that interferes with our exploration of multi-grained concept annotations for MLLMs.
Hence, we take the above strategies to ensure the quality of synthesized annotations (captions and label descriptions).
Moreover, in \datasetname{}, fine-grained attribute and relationship labels are only available in partial images from the original datasets.
Although we try to automatically generate fine-grained attributes and relationship annotations with external open-source models based on image captions (and images), after careful manual checks, the synthesized fine-grained annotations are found to be of low quality, with non-negligible hallucinations and noise, which are very likely to negatively impact MLLMs.
Thus, only high-quality manually annotated fine-grained annotations were used in \datasetname{} to avoid unnecessary hallucinations and noise.
We focus our limited resources and effort on carefully handling existing data with meticulous manual checks and in-depth experimental exploration to ensure the quality of \datasetname{} and the reliability of our exploration.

To further ensure that such partially missing fine-grained annotations do not introduce additional bias, we compare \datasetname{} with only object labels and label descriptions, as well as \datasetname{} with all category labels and label descriptions, \ie{} \datasetnamenoregion{}.
We find that the performance is still improved by the partially available fine-grained attribute and relationship labels.
More importantly, as shown in \Sref{sec:experiment:meso} and \Fref{fig:meso_seed_all} \figleft{}, compared to \datasetname{} with only image captions, multi-grained concept annotations in \datasetname{} can still help MLLMs achieve significant improvements of $2.15$ and $10.31$ points in the corresponding "Instance Attribute" and "Instance Interaction" dimensions, respectively.
Similar results are also observed in \Apref{sec:appendix:self_generated_seed} and \Fref{fig:app:self_generated_seed}, where MLLMs trained with \datasetname{} can self-generate fine-grained category labels during the inference stage to further improve the performance by $1.79$ and $5.16$ points in the corresponding "Instance Attribute" and "Instance Interaction" dimensions, respectively.
Above experimental results demonstrate that partially missing fine-grained attribute and relationship labels do not affect the effectiveness of \datasetname{} and the reliability of our exploration.

\subsection{Difference between \datasetname{} and Existing Image--Text Interleaved Datasets}
\label{sec:appendix:discussion:interleaved}

MMC4~\citep{zhu2023multimodal} and OBELICS~\citep{laurenccon2024obelics} are two widely-used open-source large-scale image--text interleaved datasets.
They are constructed from massive HTML documents.
MINT-1T~\citep{awadalla2024mint} further enhances the scale and diversity, uniquely including data from PDFs and ArXiv documents.
These datasets are large-scale, diverse, and also noisy.
They are important for MLLMs to reason across image and text modalities, and have been used to further improve the capabilities and performance of MLLMs, especially in image--text interleaved scenarios.
Our proposed \datasetname{} can also be seen as image--text interleaved documents with higher-quality and better image--text alignment, which has multi-grained concept annotations to help MLLMs better locate and learn concepts, thereby promoting vision--language alignment across multiple granularities simultaneously.
It is a promising direction to automatically synthesize multi-grained concept annotations for these web-scale datasets.

\subsection{Image-Independent Label Description Generation}
In~\Sref{sec:dataset:annotation} and ~\Apref{sec:appendix_data:label_des_gen}, we describe the process of generating label descriptions for category labels in \datasetname{}.
Considering that our description generator, GPT-4, do not see the images both during training and annotation synthesis, such general image-independent generation of label descriptions may introduce hallucinations and noise.
In this paper, to ensure the quality of synthesized label descriptions and avoid hallucinations, we particularly consider the following aspects:
\begin{itemize}[leftmargin=2em]
	\item \textbf{Effectiveness of Synthesized Label Descriptions.} As mentioned in~\Sref{sec:dataset:annotation}, many previous works use WordNet and LLM to obtain general image-independent label descriptions, and successfully improve the model's understanding of concepts corresponding to these labels. The experimental results in~\Tref{tab:data_recipes} of ~\Sref{sec:experiment:data_recipes} also show that label descriptions can bring significant performance improvements.
	\item \textbf{Feasible Synthesis for Common Concepts.} Whether in previous works~\citep{menon2023visual} or in the four open-source datasets used in this paper, since the target labels are \textbf{common} concepts, LLM can generate label descriptions with good generality and rich visual information even without seeing the image.
	\item \textbf{Disambiguate Polysemous Labels.} We also use WordNet to disambiguate the polysemous labels in \datasetname{}, such as ``batter'' $\rightarrow$ ``batter (ballplayer)'' or ``batter (cooking)'', to further reduce the ambiguity in the label descriptions, improve the generality and quality.
	\item \textbf{Careful Quality Control.} We not only carefully manual-check all generated label descriptions to ensure their quality and avoid introducing additional noise or hallucinations, but also carefully design prompt templates and manual annotation examples (in~\Apref{sec:appendix_data:label_des_gen}) to guide and improve the quality of the annotations generated by LLM. Interestingly, we find that the powerful GPT-4 can even automatically identify invalid noisy labels in the labels with the customized prompt templates, helping us to clean up these invalid annotations.
\end{itemize}

\subsection{A General MLLM Framework with an Autoregressive Discrete Objective}
In~\Sref{sec:method:framework}, we describe a general MLLM framework with an autoregressive discrete objective, which is mainly based on SEED-LLaMA and LaVIT.
In this paper, our core research goal is not to explore what loss function can better leverage the potential of \datasetname{}, hence we follow the general loss function of next-token prediction, which is common and effective for both MLLMs and the use of fine-grained annotations.
\begin{itemize}[leftmargin=2em]
	\item \textbf{Feasibility of Next-Token Prediction.} Using next-token prediction as a learning target to promote multimodal comprehension and generation is a common practice in VLM~\citep{wang2022OFA,lu2022unified} and MLLM~\citep{sun2023generative,ge2023making,team2024chameleon} work, and this paper follows this general loss function. The experiments of previous work and this paper have shown the effectiveness of this loss function in promoting VLMs and MLLMs.
	\item \textbf{Leveraging Fine-Grained Annotations.} No matter in VLM~\citep{wang2022OFA,lu2022unified,chen2021pix2seq} or MLLM~\citep{peng2023kosmos,you2023ferret,zhang2024mm1}, there have been many works that leverage fine-grained annotations by directly using next-token prediction as a learning target, and have achieved good results in various downstream tasks.
	\item \textbf{Fair Comparison and Applicability.} This paper avoids introducing additional components and loss functions (such as bounding box prediction), to ensure the fair comparison of baselines trained with \datasetname{} (image captions and fine-grained annotations) and \icdatasetname{} (only image captions).
	This paper integrates multi-grained concept annotations into image-text interleaved documents through the structured template, so that \datasetname{} and \icdatasetname{} can be explored in a fair and comparable way under a general MLLM framework and an autoregressive loss function, to ensure the applicability and comparability of the experimental results in different MLLM frameworks.
\end{itemize}

\subsection{Performance of Different Data Recipes}

In~\Sref{sec:experiment:data_recipes} and~\Tref{tab:data_recipes}, we compare the performance of different data recipes for multi-grained concept annotations.
It is worth noting that:
\begin{itemize}[leftmargin=2em]
	\item \textbf{Evaluation Format Preference.} The image captioning and image generation tasks in~\Tref{tab:data_recipes} have natural advantages for MLLMs trained with coarse-grained annotations, since the evaluation data format of these two tasks is completely consistent with the training data format of coarse-grained annotations, \ie{} directly generating target captions or images based on given images or captions.
	As for the training data format based on multi-grained annotations, such as the \textcolor{indexcolor}{$3$}-rd row, contains multi-grained concept annotations, which are significantly different from the evaluation data format. For example, in the image generation task, the training data format of the \textcolor{indexcolor}{$3$}-rd row is to generate target images based on given image captions and multi-grained concept annotations, which is more complex and different.
	\item \textbf{Significant Improvements Under Adverse Conditions.} Even under such adverse evaluation conditions, compared to the \textcolor{indexcolor}{$0$}-th row, the \textcolor{indexcolor}{$3$}-rd row with multi-grained annotations has achieved significant improvements in the CIDEr metric of the image captioning task ($+4.66$ and $+2.9$), the FID and CLIP-I metrics of the VIST dataset in the image generation task ($+32.28$ and $+3.88$), and also achieved certain improvements in the CLIP-T and CLIP-I metrics of the MS-COCO-30K dataset ($+0.76$ and $+0.62$), and is close to the \textcolor{indexcolor}{$0$}-th row in FID ($7.36$ vs. $7.20$).
	This does not mean that the rich information introduced by multi-grained annotations has not brought significant improvements. On the one hand, the evaluation data format is more favorable for the \textcolor{indexcolor}{$0$}-th row, and on the other hand, the MS-COCO-30K dataset is relatively easy and saturated, so the magnitude of improvement is not as large as in other datasets, but still has certain improvements.
	Specifically, its evaluation data is similar in style to the images in our training data, which are COCO-style images of common concrete concepts.
	Moreover, for example, the best reference model LaViT-v2-7B uses more than $200$M pre-training data to get $7.10$ FID, $31.93$ CLIP-T and $71.06$ CLIP-V; our \textcolor{indexcolor}{$3$}-rd row uses only $4$M \datasetname{} as pre-training data to get $7.36$ FID, $31.57$ CLIP-T and $72.24$ CLIP-V.
\end{itemize}
Overall, the \textcolor{indexcolor}{$3$}-rd row in~\Tref{tab:data_recipes} has achieved significant performance improvements in most metrics compared to the \textcolor{indexcolor}{$0$}-th row, which clearly demonstrates the potential of multi-grained concept annotations for MLLMs.

\subsection{Collaboration between \datasetname{} and \icdatasetname{}}
In~\Sref{sec:experiment:scalable} and~\ref{sec:experiment:sft}, we explore the collaboration between \datasetname{} and \icdatasetname{}.
We can observe that:
\begin{itemize}[leftmargin=2em]
	\item \textbf{Respective Advantages.} As stated in~\Sref{sec:experiment:scalable}, \datasetname{} and \icdatasetname{} have respective advantages in the depth and breadth of concept representation. Therefore, take~\Tref{tab:sft_understanding} as an example, when comparing \textcolor{indexcolor}{$2$}-nd and \textcolor{indexcolor}{$1$}-st rows, even though \icdatasetname{} contains $52$M images, the performance improvement is not very significant on tasks that inspect in-depth understanding of common concrete concepts, but is significant on tasks that require a broader understanding of concepts. 
	\item \textbf{Further Analysis.} In~\Apref{sec:appendix:concept_overlap}, we further analyze this phenomenon from the perspective of dataset statistics and concept overlap (K-LITE). 
	We found that although the concept overlap between the training data and the downstream evaluation dataset increases significantly with the increase of \icdatasetname{} ($4$M $\rightarrow$ $52$M), the performance is still lower than that of the MLLM trained with \datasetname{}, which has a significantly lower concept overlap. 
	We attribute this to multi-grained concept annotations and higher image quality of \datasetname{}.
	\item \textbf{Importance of Collaboration.} Most importantly, the collaboration between \datasetname{} and \icdatasetname{} can achieve the best performance on all tasks, which further shows the importance of multi-grained concept annotations in MLLM.
	Therefore, we believe that the limitations of the performance improvement of \icdatasetname{} on some tasks, and the further performance improvement brought by the combination with \datasetname{}, further highlight the importance of the multi-grained concept annotation dataset \datasetname{} for future MLLM research.
	\item \textbf{Feasibility of Automated Large-Scale Multi-Grained Data Construction.} 
	While the collaboration between \datasetname{} and \icdatasetname{} dataset can bring obvious benefits, the scale of the \datasetname{} dataset ($4$M) remains relatively small compared to the vast scale of image--caption datasets.
	It is indeed a huge challenge to further expand the data scale by manual annotation or sophisticated automated systems.
	However, as discussed in~\Sref{sec:appendix:limitation:automatic}, 
	existing works such as All-Seeing~\citep{wang2023all} and GLaMM~\citep{rasheed2023glamm} have automatically constructed large-scale (even billion-level) open-source datasets with detailed object-level annotations to enhance the performance of MLLM on various object-level and grounding tasks. 
	Although we observe some hallucinations and noise in these datasets, the performance improvements achieved by these two datasets as preliminary valuable attempts to synthesize large-scale fine-grained annotations, demonstrate the feasibility of automated construction of such large-scale datasets.
	Therefore, \datasetname{} and existing works in automated construction of large-scale fine-grained annotations are complementary, and it is feasible to further expand the scale of \datasetname{} dataset by building automated systems in the future. 
\end{itemize}
\section{Experimental Details}
\label{sec:appendix:experiment_details}

\subsection{Training Details}
\label{sec:appendix:train_details}
The training hyperparameters and cost are shown in \Tref{tab:train_hyperparameters}.

\begin{table}[t]
    \centering
	\tablestyle{6pt}{1.2}
	\caption{
		Training hyperparameters and cost for both pre-training and supervised fine-tuning (SFT) stages.
		\datasetname{} and \icdatasetname{} refer to two pre-training datasets.
	}
	\label{tab:train_hyperparameters}
	\vspace{1em}
	\scalebox{0.9}{
		\begin{tabular}{l|ccc}
			\toprule
			\multicolumn{4}{c}{\textit{Training hyperparameters}} \\

			\midrule
			Training Data & \datasetname{} & \icdatasetname{} & SFT Data\\

			\midrule
			Optimizer & AdamW & AdamW & AdamW\\
			Learning Rate & 2e-4 & 2e-4 & 4e-4\\
			Weight Decay & 0.05 & 0.05 & 0.05\\
			Training Epochs & 1 & 1 & 1\\
			Warmup Ratio & 0.1 & 0.1 & 0.05 \\
			Min learning rate ratio & 0.1 & 0.1 & 0.1 \\
			Learning Rate Scheduler & Cosine & Cosine & Cosine\\
			Batch Size & 512 & 512 & 256\\
			Maximum Token Length & 2048 & 2048 & 2048\\
			\midrule
			\multicolumn{4}{c}{\textit{Training Cost}} \\
			\midrule
			GPU Device & \multicolumn{3}{c}{32$\times$NVIDIA A100-80G} \\
			Training Time & $\sim$11h & $\sim$20h & $\sim$5h\\
			\bottomrule
		\end{tabular}
	}
\end{table}

\paragraph{Training Loss.}
Similar to LLMs~\citep{brown2020language,touvron2023llama2} and MLLMs~\citep{ge2023making,jin2023unified} under the autoregressive discrete framework, as we discussed in \Sref{sec:method:pretraining}, we train our MLLMs with an autoregressive objective to maximize the likelihood of predicting the next visual or textual token in a multimodal discrete token sequence as follows:
\begin{equation}
	L=\sum_{i=1}^{|u|} \log P\left(u_i \mid u_1, \ldots, u_{i-1}\right) \nonumber
\end{equation}
where $u$ denotes the multimodal token sequence, and $u_i$ denotes the $i$-th token in the sequence.
We can divide the multimodal token sequence into two parts: visual tokens and textual tokens.
In the MLLM training, we find that the loss of the visual tokens $L_V$ is larger than the loss of the textual tokens $L_T$. 
To balance the loss of the visual and textual tokens, we introduce a loss scale ratio $\alpha=0.1$ to scale the visual token loss.
The final loss is defined as $L = \alpha \cdot L_V + L_T$.

\paragraph{Image-First and Text-First.}
Our pre-training data consists of image--caption pairs (\icdatasetname{}) and image--text interleaved documents (\datasetname{}).
We follow common practice~\citep{alayrac2022flamingo} to randomly select the image-first or text-first data template for each data sample as shown in \Sref{sec:appendix_data:template}.
The difference between the two templates is the order of appearance of the image and text in the sample, which determines the condition of each token during autoregressive training.
For the multimodal instruction data we collect from public datasets, we follow their original data formats and do not change the order of appearance of the image and text in the sample.

\paragraph{Training Parameters.}
In all training stages, we only tune partial parameters of the LLM: the VL vocabulary, additional LoRA modules~\citep{hu2021lora}, norm layers and a language model head layer, to greatly improve efficiency. All data are pre-tokenized, and we don't load all visual modules during training. 

\begin{figure}[t]
	\centering
	\includegraphics[width=0.95\textwidth]{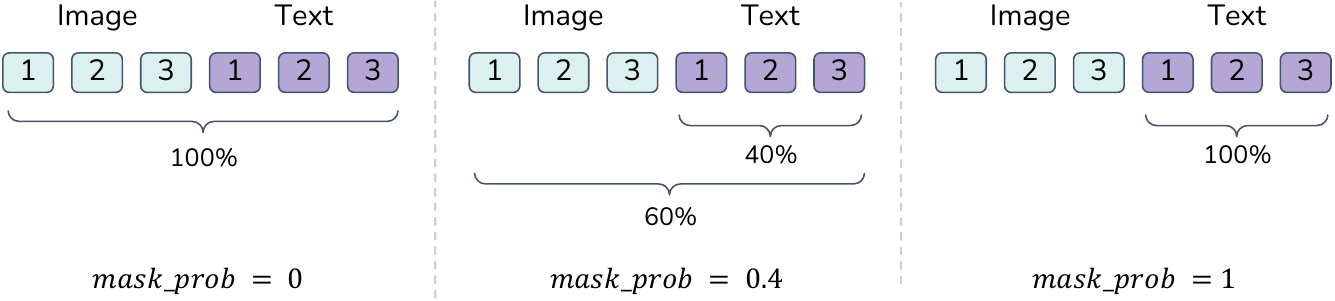}
	\caption{
		A brief illustration of the hyperparameter $mask\_prob$ in the pre-training stage when the template is image-first.
	}
	\label{fig:mask_prob}
\end{figure}

\paragraph{Loss Calculation Strategy.}
Moreover, in SFT stage, we follow LLaVA~\citep{liu2023visual} to only calculate loss for the answer tokens.
In the pre-training stage, we find that existing MLLMs typically calculate loss for all tokens or only for the textual tokens when visual tokens as the condition, and vise versa.
We conduct experiments to investigate the impact of different loss calculation strategies on the performance of MLLMs.
Take the image-first template as an example, we introduce a hyperparameter $mask\_prob$ to control the probability of only computing the loss for the textual tokens in each sample.
As shown in \Fref{fig:mask_prob}, when $mask\_prob = 0$, we calculate the loss for all tokens in each sample; when $mask\_prob = 0.4$, we calculate the loss for the textual tokens only in $40\%$ of the samples, and calculate the loss for all tokens in the rest $60\%$ of the samples; when $mask\_prob = 1$, we calculate the loss for the textual tokens only in all samples.
$mask\_prob$ can be seen as a sample-level mask ratio of modality tokens, which controls the difficulty of the pre-training task.
Higher $mask\_prob$ means easier pre-training task since more tokens are masked out.
We conduct experiments with $0.1$ as the step size to investigate the impact of $mask\_prob = [0, 1]$ on the performance of MLLMs on image captioning and image generation benchmarks.
Interestingly, the situation is more complicated than we thought.
For image captioning tasks, the performance increases with the increase of $mask\_prob$, but the performance suddenly drops at $mask\_prob = 1.0$, even lower than the performance at $mask\_prob = 0.0$.
For image generation tasks, the performance decreases first and then increases with the increase of $mask\_prob$ (the inflection point is $mask\_prob = 0.5$, the performance is better than that at $mask\_prob = 0.0$), and then decreases again (the inflection point is $mask\_prob = 0.8$).
To balance the performance of comprehension and generation tasks, we choose $mask\_prob = 0.9$ as the final hyperparameter.
Since this is not the focus of this paper, we do not conduct in-depth research and analysis on this, but we believe that this is an interesting phenomenon worthy of further research, and we encourage future researchers to conduct in-depth research on this.

\subsection{Evaluation Details}
\label{sec:appendix:eval_details}
\paragraph{Evaluation Benchmarks and Metrics.} 
We show the evaluation benchmarks and their corresponding data splits and metrics in \Tref{tab:eval_bench_pretrain} and \Tref{tab:eval_bench_sft} for the pre-training and supervised fine-tuning stages, respectively.
\begin{table}[t]
    \centering
	\tablestyle{5pt}{1.2}
	\caption{
		Evaluation benchmarks, eval splits and eval metrics for the pre-training stage.
	}
	\vspace{1em}
	\label{tab:eval_bench_pretrain}
	\scalebox{0.9}{
		\begin{tabular}{llll}
			\toprule
			Dataset & Task Description & Eval Split & Metric \\
			\midrule

			COCO  & Scene description   & \texttt{test} &  CIDEr \\
			NoCaps  & Scene description   & \texttt{test} &  CIDEr \\
			\midrule
			MS-COCO & Text-Conditional Image Synthesis  &  \texttt{val} ($30$K)   & FID, CLIP-I, CLIP-T \\
			VIST & Contextual image synthesis  & \texttt{val} & FID, CLIP-I\\
			\bottomrule[0.95pt]
		\end{tabular}
	}
\end{table}

\begin{table}[t]
	\centering
	\tablestyle{5pt}{1.2}
	\caption{
		Evaluation benchmarks, eval splits and eval metrics after SFT.
	}
	\label{tab:eval_bench_sft}
	\vspace{1em}
	\scalebox{0.8}{
		\begin{tabular}{llll}
			\toprule
			Dataset & Task Description & Eval Split & Metric \\
			\midrule
			COCO  & Scene description   & \texttt{test} &  CIDEr \\
			NoCaps  & Scene description   & \texttt{test} &  CIDEr \\
			VQAv2  & Scene understanding QA   & \texttt{test} & VQA Acc \\
			GQA & Scene understanding QA & \texttt{test} &  VQA Acc \\
			VizWiz & Scene understanding QA & \texttt{test} &  VQA Acc \\
			POPE & Visual Hallucination & - & F1-score \\
			MME & Multimodal Comprehension & \texttt{test} & MME score \\
			MMBench & Multimodal Comprehension & \texttt{dev} & Acc \\
			ScienceQA-IMG & Multimodal Comprehension & \texttt{test} & Acc \\
			SEED-Bench-IMG & Multimodal Comprehension & \texttt{test} & Acc \\
			\midrule
			MS-COCO & Text-Conditional Image Synthesis  &  \texttt{val} ($30$K)   & FID, CLIP-I, CLIP-T \\
			VIST & Contextual Image Synthesis  & \texttt{val} & FID, CLIP-I\\
			DreamBench & Subject-driven Image Editing  & 
			From \textsc{Kosmos-G} & DINO, CLIP-I, CLIP-T \\
			\bottomrule
		\end{tabular}
	}
\end{table}

\paragraph{Evaluation Strategy.}
In this paper, we evaluate the performance of MLLMs on both multimodal comprehension and generation tasks, including image captioning, text-to-image generation, visual question answering, multi-choice benchmarks, and image editing.
We follow standard evaluation strategies and dataset splits to ensure the comparability of our experimental results.
We implement the generation of visual and textual tokens in the evaluation stage based on the \href{https://github.com/vllm-project/vllm}{vLLM} library, and generate the final image through the \href{https://github.com/huggingface/diffusers}{Diffusers} library.
Notably, multi-choice benchmarks have different designs and implementations in task instruction templates and evaluation strategies.
The impact of different task instruction templates on the model's performance is significant and cannot be ignored.
Therefore, we manually design at least $8$ task instruction templates for each dataset and select the best performance as the final result to ensure the stability of the experimental results.
The \href{https://huggingface.co/blog/open-llm-leaderboard-mmlu}{evaluation strategy} is more complicated, and some benchmarks, such as MMBench~\citep{liu2023mmbench}, also introduce an external LLM (GPT-4~\citep{achiam2023gpt}) as the choice extractor. 
To pursue simplicity and efficiency, we directly select the highest probability of the candidate option tokens, \ie{} ``Yes'' and ``No'' for POPE~\citep{li-etal-2023-evaluating}, ``A'', ``B'', ``C'', ``D'', \etc{} for other benchmarks, as the final answer.
We don't introduce any external LLMs in our experiments since it will increase the cost and leads to unstable results~\citep{liu2023hallusionbench} compared with the above strategy.
Similarly, we don't evaluate on benchmarks that use GPT-4 as the default evaluator, since it will increase the cost and has strong bias on models that tuned with GPT-4 synthetic data~\citep{lin2023vila}.
\section{Details of \datasetname{} Dataset}
\label{sec:appendix:dataset}
In this section, we provide the details for constructing \datasetname{} dataset. The statistics of \datasetname{} dataset are shown in \Tref{tab:pretrain_data_statistics}.

\subsection{Dataset Collection}
\label{sec:appendix:collection}

We collect four large-scale object detection dataset, including Open Images~\citep{OpenImages}, Objects365~\citep{shao2019objects365}, V3Det~\citep{wang2023v3det} and Visual Genome~\citep{krishna2017visual} for \datasetname{}.

BigDetection~\citep{cai2022bigdetection} has found that Open Images and Objects365 are biased towards certain scale (small or large) of objects, while the improved object annotations for these two datasets in BigDetection are balanced across object scales. 
Therefore, we use the improved object annotations of Open Images and Objects365 from BigDetection. 
For Open Images, we merge the BigDetection object annotations with the attribute and relationship annotations of Open Images. Specifically, for each object region in BigDetection, we calculate the $IoU$ with each Open Images object region in the same image, and select the Open Images object region with the largest $IoU$ to pair with it.  
We only keep the relationship and attribute annotations in Open Images where subject object and the object object are both successfully paired. 
We do not use the LVIS~\citep{gupta2019lvis} subset in BigDetection since all images in LVIS come from COCO~\citep{lin2014microsoft}, which is our downstream evaluation dataset. 
For V3Det and Visual Genome, we use all their original annotations for further pre-processing.

Since BigDetection dataset does not involve any changes to the positions of the objects in the Open Images and the objects in the same image are a subset of the objects in the Open Images, the matching process is correct and appropriate. 
We use $IoU$ instead of exact matching because BigDetection uses coordinates to provide object positions, whereas Open Images use ratios, where using exact match may introduce errors.

For V3Det, there are $60$ images that were not successfully downloaded, and we simply discarded these images.

\begin{table}[t]
    \centering
    \tablestyle{6pt}{1.2}
    \caption{
		Statistics of \datasetname{} dataset and four object detection datasets underlying it.
		The last $4$ columns are based on the final tokenized dataset and the other columns are before tokenization.
		The {\# Concepts} column denotes the number of unique category label--description pairs in each dataset.
		Note that Visual Genome and V3Det are repeated $3$ times to keep data balance in the final dataset.
    }
    \label{tab:pretrain_data_statistics}
	\vspace{1em}
    \scalebox{0.9}{
		\begin{tabular}{c|l|ccr|cccc}
			\toprule
			\multicolumn{1}{l|}{\multirow{2}{*}{}} & \multirow{2}{*}{Dataset} & \multirow{2}{*}{\# Regions} & \multirow{2}{*}{\# Images} & \multirow{2}{*}{\# Concepts} & {\# Visual} & {\# Textual} & {{ }{ }\# Used} & \multirow{2}{*}{\# Samples} \\
			\multicolumn{1}{l|}{} & & & & & { }{ }Tokens & { }{ }{ }Tokens & Regions &\\ 
			\midrule
			\textcolor{indexcolor}{0} & Open Images & { }{ }7.16M & 1.44M & 642{ }{ } & 0.70B & 0.47B & { }{ }6.21M & 1.44M\\
			\textcolor{indexcolor}{1} & Objects365 & 15.71M & 1.78M & 600{ }{ } & 1.31B & 0.97B & 12.58M & 1.78M\\
			\textcolor{indexcolor}{2} & V3Det & { }{ }1.18M & 0.18M & 12,976{ }{ } & 0.26B & 0.21B & { }{ }2.30M & 0.53M \\
			\textcolor{indexcolor}{3} & Visual Genome & { }{ }1.50M & 0.10M & 20,830{ }{ } & 0.28B & 0.28B & { }{ }2.77M & 0.31M \\
			\textcolor{indexcolor}{4} & \textbf{\datasetname{}}& \textbf{25.55M} & \textbf{3.48M} & \textbf{35,048}{ }{ } & \textbf{2.54B} & \textbf{1.92B} & \textbf{23.86M}& \textbf{4.06M} \\ 
			\bottomrule
		\end{tabular}
    }
\end{table}

\subsection{Dataset Pre-processing}
\paragraph{Open Images, Objects365 and V3Det.}
There are $3$ steps to pre-process these three datasets. 
Notice that we directly use the train and dev splits from the original datasets and don't use the test set.\looseness=-1
\begin{enumerate}[leftmargin=2em, topsep=0em, itemsep=0em]
	\item \textbf{Clean labels}: to remove the noise that may be introduced by long-tail distribution, we count the number of occurrences of each label in each dataset, and then select a threshold to remove low-frequency labels (typically noise). 
	For V3Det, we remove object labels with frequency less than 3. 
	For Open Images and Objects365, we don't remove any label since each label appears many times and has high quality.
	\item \textbf{Clean objects and object-related information}: we remove objects that have illegal coordinates, exceed the image range or the corresponding label is already removed. 
	For Open Images, we also remove the attribute and relationship annotations corresponding to these illegal objects. 
	\item \textbf{Clean images}: 
	since the de-facto visual encoder (\ie{} Vision Transformer, ViT) requires a square input image, we resize all images to a fixed resolution of $224 \times 224$ pixels.
	Following standard practices~\citep{zeng2023x}, to improve the data quality, we first remove images with short edges less than $224$ pixels or aspect ratios greater than $3.0$ or less than $0.33$, to prevent the image from changing too much after resizing.
	We also remove the images without any objects.
\end{enumerate}

\paragraph{Visual Genome.}
Visual Genome is a dataset trying to connect structured image concepts to language. 
Each image is annotated with:
1) image-level info, \ie{} image captions; 
2) region-level info, including objects in the region and the region description;
3) object-level info, including object label and location, the attributes of objects, the relationships between objects, with all the labels mapped to WordNet synsets.
We use the image-level info and object-level info for further pre-processing which require the following steps. \footnote{
	We found that region-level annotations may not be suitable for constructing \datasetname{} since: $1.$many regions only contains a single object; $2.$the annotations of the objects in region may not be exactly in the region; $3.$the region description may not be exactly matched with the region itself. 
	Hence, We use $v1.4$ version of the dataset which includes image-level and object-level annotations except the attribute of objects. Since we found that the objects in $v1.2$ is the subset of that in $v1.4$, we use the attribute annotations in $v1.2$ 
} 

\begin{enumerate}[leftmargin=2em, topsep=0em, itemsep=0em]
	\item \textbf{Clean images}: since all images need to be resized to $224 \times 224$ and input to the visual modules, we remove the images with the shorter side length less than $224$, images with an aspect ratio greater than $3$ or less than $0.33$ to prevent the image from changing too much after resize.
	\item \textbf{Clean object annotations}: we remove all objects that have more than one synset as we found that they are mostly noisy. We also remove objects that have illegal coordinates or exceed the image range.
	\item \textbf{Construct, clean object label, clean object annotations again}: in Visual Genome, the same object label may have different semantics distinguished by the synsets they are mapped to. 
	To mitigate the noise brought by long-tail distribution, firstly, we count the frequency of each (object label, synset) pair if the mapped synset is not empty. Secondly, for those object labels that aren't mapped to any synset, we regard that it is mapped to the synset that has the highest frequency of the same object label in the first step (if there isn't any same object label, we regard that it is mapped to the empty synset) and add its frequency to the first step. Finally, we remove all the (object label, synset) pairs with frequency less than $2$ and also remove all the corresponding objects. 
	To get the final object label name, for the object labels which have more than two types of synsets, we manually check them with help of WordNet and if so, we add some keywords corresponding to the synset behind the object label name as the final name for better differentiation (e.g. (batter, batter.n.01) $\rightarrow$ ``batter (baseball player)'', (batter, batter.n.02) $\rightarrow$ ``batter (semiliquid mixture)''). For other object labels, we directly use the object label name as the final name. 
	\item \textbf{Clean images again}: we remove all the images that don't have any object annotation.
	\item \textbf{Clean relationship and attribute annotations}: firstly, we remove all the relationships where the corresponding object is removed by step $2$ and step $3$. Secondly, we count the frequency of each label  and remove the relationship/attribute whose frequency is less than $5$ or is empty. We don't add additional information in label name as in step $3$ as we have checked manually that each label name has exactly one semantic.
	\item \textbf{Remove unreasonable labels and annotations}: for each label, we ask ChatGPT if it is an reasonable label in \Sref{sec:appendix_data:label_des_gen}. For unreasonable labels, we remove them and the corresponding object, attribute, relationship annotations.
	\item \textbf{Dataset split}: for images in Visual Genome with a coco id, we follow Karpathy's split: images in `val' and `restval' are used as the validation set, and images in `test' are discarded. For images in Visual Genome with a flickr id but no coco id, we match them with images from flickr30k and discard the matched images. All other images are used as the training set.
\end{enumerate}

\subsection{Object Annotation Pre-processing}
\label{sec:appendix_data:obj_annotation_preprocessing}
The fine-grained object annotations provided in these datasets include bounding box coordinates and category labels for each object.
To accommodate varying aspect ratios of object bounding boxes and the requirement for a square input image, following \textsc{Kosmos-G}~\citep{pan2023kosmosg}, we crop a new larger square object region $S_i$ that contains the original object region $R_i, R_i \subseteq S_i$, with their centers aligned as closely as possible.
Since $S_i$ may include not only the object region $R_i$ but also the surrounding object regions, we design a simple but effective strategy to update the object label annotations of $S_i$ by merging the object label annotations of $R_i$ with the object label annotations of surrounding object regions, to improve the quality of object annotations. 
For each $R_i$ and its corresponding new region $S_i$, we calculate the intersection-over-area (IoA) between $S_i$ and all $R_{j(j \neq i)}$, \ie{} $IoA(S_i, R_j) = \frac{Area(S_i \cap R_j)}{Area(R_j)}$.
If $IoA(S_i, R_j) \geq 0.8$, we consider $R_j$ as a part of $S_i$ and update the annotations of $R_j$ to the annotations of $S_i$.
This strategy can help to include more relevant object regions in the region $S_i$ and improve the quality of object annotations.
After that, we remove $S_i$ with duplicate annotations and keep the unique ones.

Since each visual token in our framework corresponds to a $14 \times 14$ pixel region, we remove $S_i$ with edge length less than $28$ pixels or greater than $182$ pixels to ensure that the object region is not too small or too large.
We also remove all the images that don't have any bounding box and resize all images to $224 \times 224$ pixels.
Finally, for objects in the image, \datasetname{} provides visual tokens of cropped regions and textual tokens of corresponding fine-grained category labels and location descriptions.
The pre-processed object regions in the images are ready to be tokenized and fill in the template in \Sref{sec:appendix_data:template}.

Specifically, instead of transform bounding box coordinates into textual tokens in the text~\citep{chen2021pix2seq,liu2023visual,peng2023kosmos} or visual markers in the image~\citep{yao2024cpt,shtedritski2023does,yang2023set}, for each object in the image, we directly provide visual tokens for the cropped region and textual tokens for the corresponding category labels.
This not only upgrades our data from simple image--text pairs to more complex image--text interleaved data which has shown to be beneficial for vision--language learning~\citep{lin2023vila,mckinzie2024mm1}, but also helps to locate and align concepts in the image and in the text by providing both the whole image and object regions.

\subsection{Caption Synthesis}
\label{sec:appendix_data:caption_synthesis}
Since only partial images are annotated with image captions in these datasets, following BLIP~\citep{li2022blip} and LAION-COCO~\citep{laion2023laioncoco}, we automatically synthesize captions for all images in the following steps:
\begin{enumerate}[leftmargin=2em]
    \item Generate $10$ candidate captions for each image with BLIP-2~\citep{li2023blip};
    \item Rank candidate captions based on image--caption similarity scores calculated by CLIP~\citep{radford2021learning};
    \item Filter too short ($<5$ words) or too long ($>25$ words) captions or captions with low image--caption similarity scores ($< 0.25$);
    \item Select the Top-$1$ caption as the final caption if exists, otherwise repeat the above steps for $10$ times. If no caption is selected, we select the caption with the highest similarity score.
\end{enumerate}

\subsection{Label Description Generation}
\label{sec:appendix_data:label_des_gen}

Label descriptions are corresponding concept descriptions of concrete concepts in the image, which convey understanding about a concept by visually observed details and relevant knowledge.
Inspired by the success of previous works~\citep{shen2022k,yao2022detclip,menon2023visual} that introduces label descriptions from WordNet~\citep{miller1995wordnet}, Wiktionary~\citep{meyer2012wiktionary} and LLMs~\citep{brown2020language} to help understand concepts, with the help of GPT-4~\citep{achiam2023gpt}, we generate label descriptions for fine-grained category labels (object, object attribute, and relationship between objects) and manually check them.
We design prompt templates for each type of category labels and carefully provided human-annotated examples. For V3Det, we directly use the label descriptions it provided.
For Open Images, Objects365 and Visual Genome, we generate label descriptions for object labels, attribute labels and relationship labels respectively.

Specifically, in system prompt, we first describe the task and add some tips that LLM might focus when generating descriptions. Then we give some examples. 
In user prompt, we instruct the LLM to generate the description.
We also ask LLM to generate `Invalid' first if the given label is noisy in Visual Genome These labels along with their corresponding annotations will be further removed.  
The full prompts are shown in \Tref{tab:label_des_gen_OI_object}, \ref{tab:label_des_gen_OI_attribute}, \ref{tab:label_des_gen_OI_relationship}, \ref{tab:label_des_gen_VG_object}, \ref{tab:label_des_gen_VG_attribute}, \ref{tab:label_des_gen_VG_relationship}.

\subsection{Template}
\label{sec:appendix_data:template}

After completing all above steps, for each image sample, we have the following annotations:

\begin{itemize}[leftmargin=2em]
	\item \textbf{Coarse-grained concept annotations}: captions and the localized narrative captions.\footnote{Only part of images in Open Images have localized narrative captions.}
	\item \textbf{Fine-grained concept annotations:} object labels, attribute labels and relationship labels along with their label descriptions.
	\item \textbf{Region-level annotations}: the object regions of the image obtained in \Sref{sec:appendix_data:obj_annotation_preprocessing} along with their locations and object labels.
\end{itemize}

We design templates to formulate the above annotations into data samples used in the pre-training stage.
 There are $2$ types of templates: image-first template shown in \Tref{tab:pretrain_template_image_first} and text-first template shown in \Tref{tab:pretrain_template_text_first}. 
 Each template has $2$ parts: image-annotation part and object-annotation part. 

\paragraph{Image-first template.} 
For image-annotation part or object-annotation part, we place the corresponding text \textbf{after} the image or region, aiming to enable the model to learn visual understanding.
\begin{itemize}[leftmargin=2em]
	\item \textbf{Image part}: we first place the image. Secondly, we place the coarse-grained concept annotation including caption and the localized narrative caption. Thirdly, we place image-level fine-grained concept annotations including object labels, attribute labels and relationship labels along with their descriptions. \textbf{For object labels}, we first place the label name and then place the label description, each object label name in the image appears exactly once.  \textbf{For attribute labels}, we first place the attribute label and then place the associated objects, followed by the attribute label description. \textbf{For relationship labels}, we first place the relationship label and then place the associated subjects and objects, followed by the relationship label description. \footnote{We design the template in this way to reduce input length and duplication.}  
	\item \textbf{Object-annotation part}: we place the object regions of the image with their location descriptions and associated object labels after them.
\end{itemize}

\paragraph{Text-first template.} 
Different from the image-first template, we place the corresponding text \textbf{before} the image or region, aiming to enable MLLMs to learn visual generation. Note that we place the captions close to the images, to help MLLMs better learn to utilize caption to generate image.

\subsection{Pre-training sample generation}
\paragraph{Pre-tokenize images and regions.}
For images and regions, we directly use the visual tokenizer of LaVIT\citep{jin2023unified} to tokenize them into visual token sequences.
	
\paragraph{Fill in the template and tokenize data.} There are $4$ steps to generate a training sample. For V3Det and Visual Genome, we repeat the process $3$ times to include as many regions as possible.
 
\begin{enumerate}[leftmargin=2em]
	\item \textbf{Select a template}: we randomly select the image-first or text-first template with a probability of $0.5$.
	\item \textbf{Fill in the template with textual annotations}: 
	\textbf{For image-annotation part}, we first fill the corresponding text. Then we remove the descriptions with a probability of $0.5$ to prevent model overfitting which might be caused by description repetitions.
	\textbf{For each object-annotation}, we first fill in the region location. Specifically, the $224 \times 224$ square image is divided into a $3 \times 3$ grid, with each cell named sequentially from top to bottom and left to right as Top Left, Top Middle, Top Right, Middle Left, Center, Middle Right, Bottom Left, Bottom Middle, and Bottom Right. The location of a region is designated by the name of the grid cell where its center is located. Then we fill in the object label with the annotations obtained in \Sref{sec:appendix_data:obj_annotation_preprocessing}. \textbf{To get the final object-annotation part}, we shuffle the regions and place as many regions as possible without exceeding the maximum tokenized length.
	\item \textbf{Tokenize the data and fill the template with visual annotations}: firstly, we tokenize the data using the tokenizer of with \texttt{[IMG]}, \texttt{[/IMG]} added as special tokens. Secondly, we replace the positions corresponding to \texttt{[IMG]} using the corresponding visual token sequence with \texttt{[IMG]} and \texttt{[/IMG]} inserted before and after it. Thirdly, we add \texttt{<s>} token and \texttt{</s>} token to the whole token sequence.
	\item  \textbf{Dealing with samples with token length more than $2048$}: for those samples with object regions, we remove one  region a time and go back to step $2$ until the token sequence length is less than $2048$. For those samples without regions but with descriptions, we go back to step $2$ but fill in the template without descriptions. Other samples are discarded.
\end{enumerate}

\paragraph{Statistics.}  

\begin{figure}[t]
	\centering
	\includegraphics[width=0.95\textwidth]{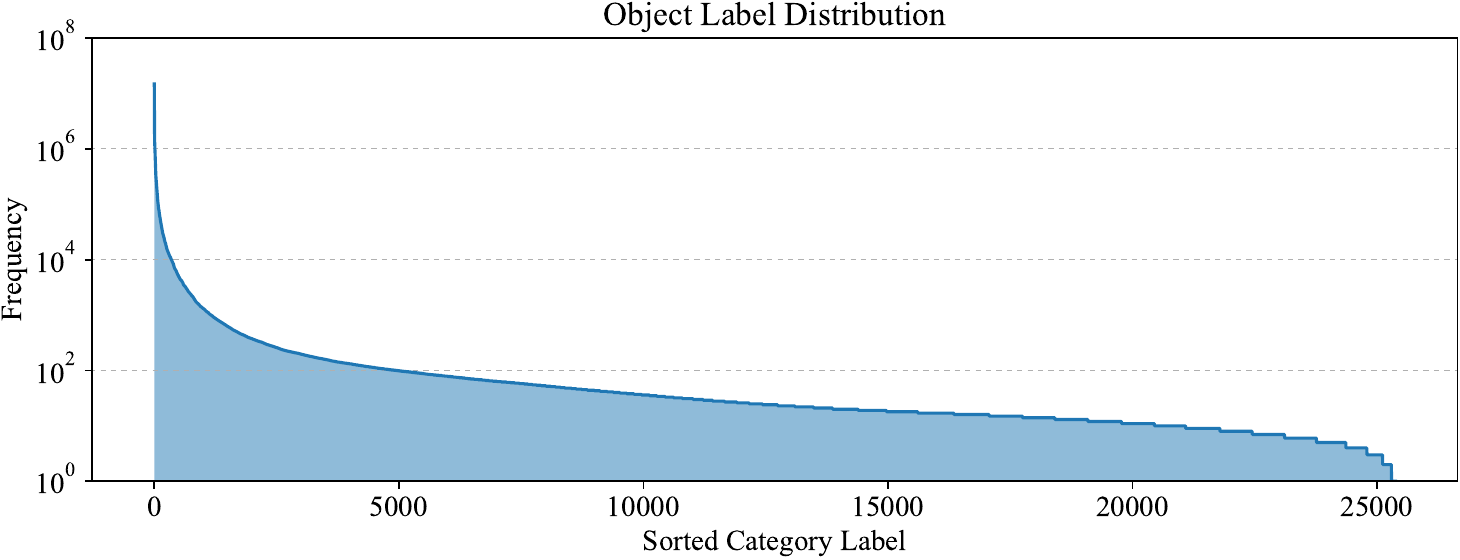}\vspace{1em}
	\includegraphics[width=0.95\textwidth]{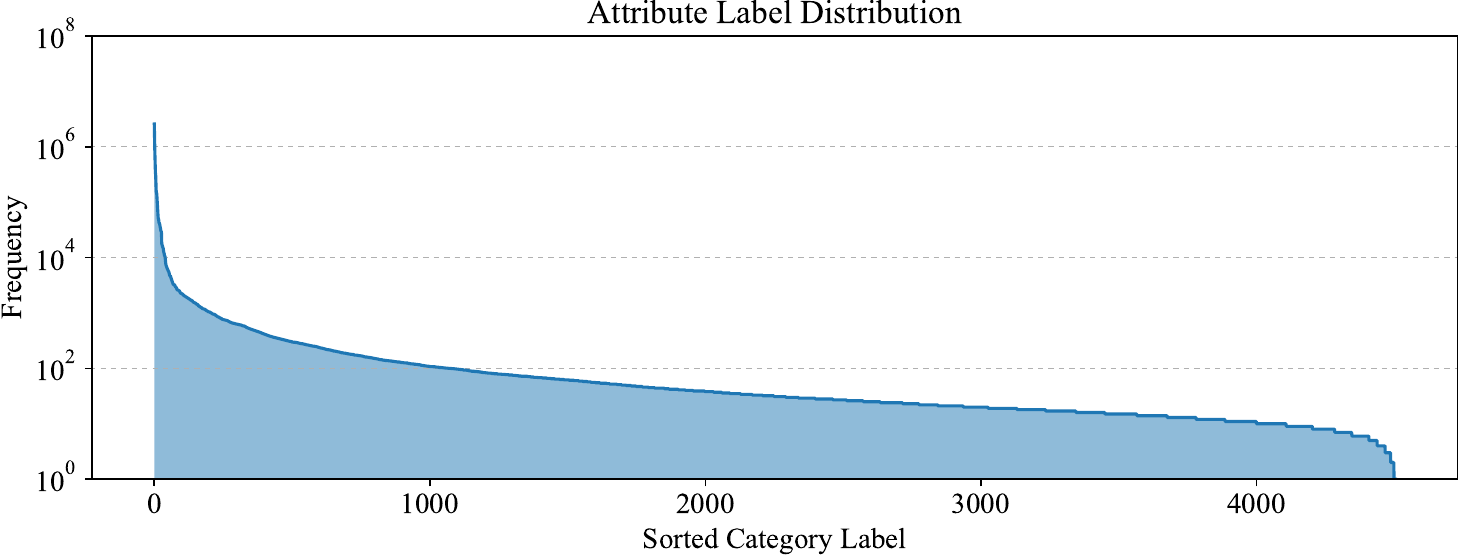}\vspace{1em}
	\includegraphics[width=0.95\textwidth]{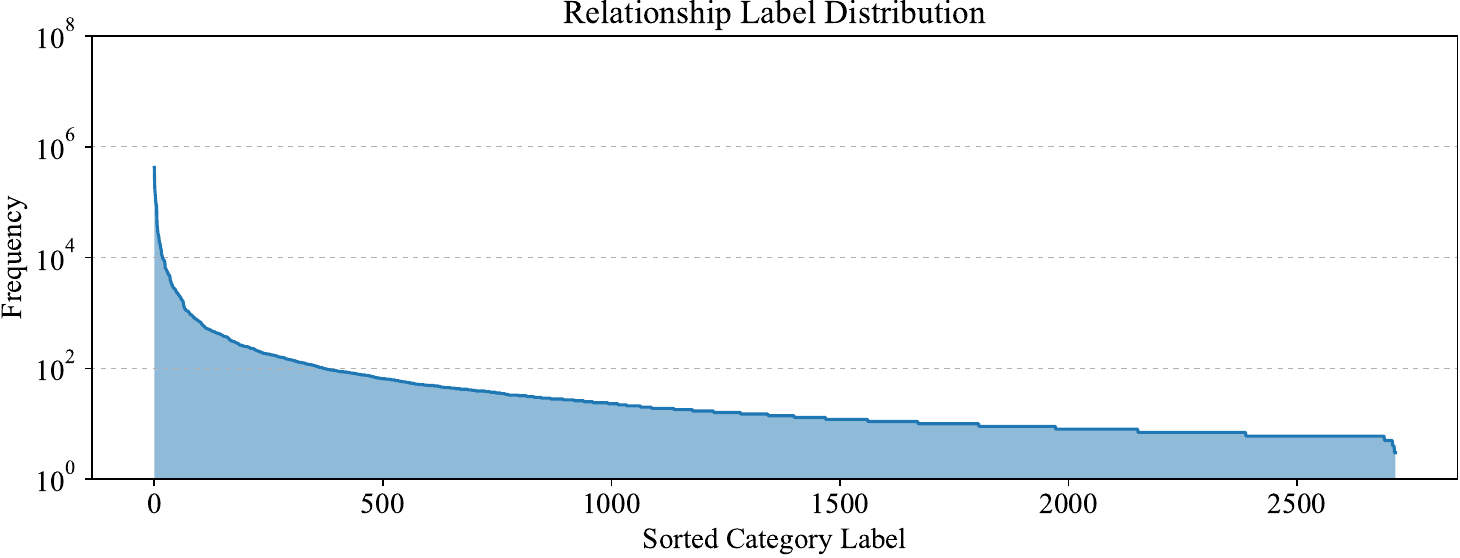}
	\caption{
		Label frequency distribution of objects, attributes and relationships in \datasetname{}.
	}
	\label{fig:concept_freq_destribution}
\end{figure}

\Tref{tab:pretrain_data_statistics} shows the statistics of the constructed \datasetname{}. 
The frequency distribution of object labels,  attribute labels and relationship labels are shown in \Fref{fig:concept_freq_destribution} respectively, which illustrates that our label types are broadly distributed. 
Moreover, only a few labels appear with particularly low frequency.
For object labels, attribute labels, and relationship labels, labels with frequency less than $5$ accounted for $5$\%, $1.5$\%, and $0.3$\% of the total number of their respective label types, respectively.\looseness=-1

\section{Details of \icdatasetname{} Dataset}
\label{sec:appendix:ic_dataset}
To strike a balance between increasing concept breadth, reducing data noise and improving efficiency, we collect several large-scale public image--caption datasets~\citep{ordonez2011sbu,sharma-etal-2018-conceptual,changpinyo2021conceptual,schuhmann2022laion,laioncoco,sun2024journeydb}, and follow LLaVA~\citep{liu2023visual} to first filter them by the frequency of noun-phrases extracted by spaCy from their given synthesized captions, and then automatically synthesize captions same as \datasetname{}.
We name this dataset as \icdatasetname{}, where $52$M unique images are collected and selected, almost $15$ times more than \datasetname{}. 
\icdatasetname{} contains $3.19$B visual tokens and $5.14$B textual tokens.

\subsection{Dataset Collection}
As we discussed in the last paragraph of \Sref{sec:appendix:discussion:applicability}, LAION-5B dataset (and its subsets) cannot be downloaded from their official website.
Therefore, we collect several large-scale public image--caption datasets from other widely-used sources:
\begin{itemize}[leftmargin=2em, topsep=0em, itemsep=0em]
	\item \href{https://github.com/salesforce/BLIP#pre-training-datasets-download}{BLIP-Captions}: BLIP~\citep{li2022blip} provides image urls, web captions and synthesized COCO-style captions for CC3M~\citep{sharma-etal-2018-conceptual}, CC12M~\citep{changpinyo2021conceptual}, SBU~\citep{ordonez2011sbu}, LAION-400M~\citep{schuhmann2021laion} (BLIP only uses and provides $115$M of them).
	Finally, we collect valid $\sim 116$M images with both web and synthesized captions and denote them as BLIP-Captions.
	\item \href{https://huggingface.co/datasets/BAAI/CapsFusion-120M}{CapsFusion-120M}: similarly, CapsFusion~\citep{yu2023capsfusion} provides image urls, web captions and synthesized COCO-style captions from LAION-2B~\citep{schuhmann2022laion}.
	Moreover, they also propose a LLM-based pipeline, denoted as CapsFusion, to merge two types of captions into a single caption, which is a well-structured sentence retaining the detailed real-world knowledge from web captions.
	We will discuss it later in the \Sref{sec:appendix:ic_data:caption_synthesis}.
	\item \href{https://huggingface.co/datasets/guangyil/laion-coco-aesthetic}{LAION-COCO-Aesthetic} and \href{https://huggingface.co/datasets/JourneyDB/JourneyDB}{JourneyDB}: we follow LaVIT~\citep{jin2023unified} to collect aesthetic image--caption data to further improve the diversity and aesthetic quality of image--caption data.
	LAION-COCO-Aesthetic is an unofficial dataset that contains 10\% samples of the LAION-COCO~\citep{laion2023laioncoco} dataset filtered by some text rules (remove url, special tokens, etc.), and image rules (image size > $384 \times 384$, aesthetic score $> 4.75$ and watermark probability $< 0.5$). It finally contains $\sim 8$M samples.
	JourneyDB is a large-scale generated aesthetic image--caption dataset that contains $\sim 4$M high-resolution Midjourney synthetic images and synthetic captions rewrote by GPT3.5 from real user prompts.
\end{itemize}

Finally, we collect $\sim 229$M general image--caption samples and $\sim 12$M aesthetic image--caption samples from the above open-source datasets.

\subsection{Data Filtering}

\begin{figure}[t]
	\centering
	\includegraphics[width=0.95\textwidth]{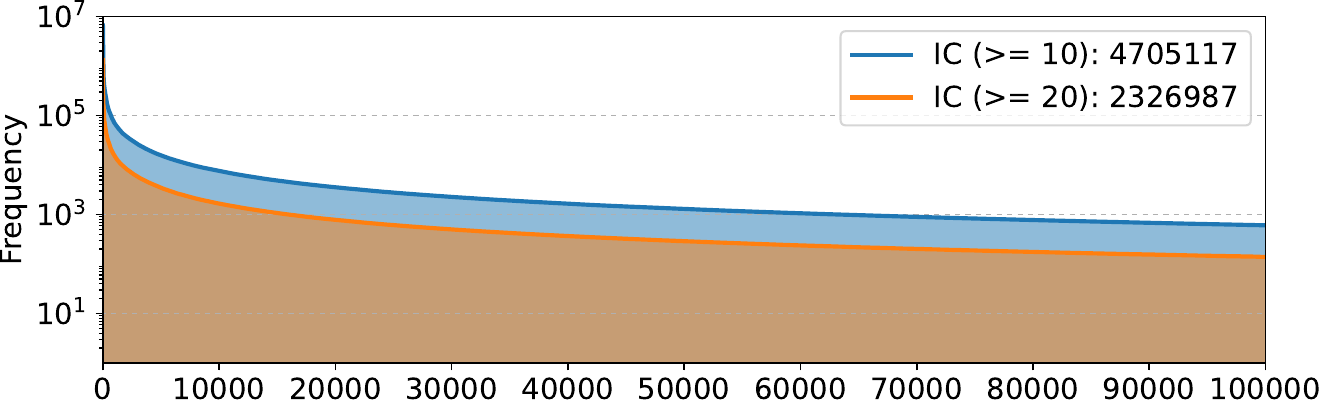}
	\caption{
		Comparison of noun-phrase statistics before and after filtering IC (not including aesthetic part).
		The x-axis represents the unique noun-phrases ordered by frequency in the descending order.
		The total number of unique noun-phrases are reported in the legend.
	}
	\label{fig:ic_noun_phrase}
\end{figure}

We follow LLaVA~\citep{liu2023visual} to downsample the above $\sim 229$M general image--caption samples based on the frequency of noun-phrases extracted by spaCy from their given synthesized captions, thereby reducing training costs while ensuring concept breadth (or concept coverage).
We denote the web captions, synthesized COCO-style captions and CapsFusion captions as \texttt{caption\_origin}, \texttt{caption\_coco} and \texttt{caption\_capsfusion} respectively.
We extract noun-phrases from \texttt{caption\_coco} of BLIP-Captions and \texttt{caption\_capsfusion} of CapsFusion-120M.
Following LLaVA, we skip noun-phrases whose frequency is smaller than $20$, as they are usually rare combinations of concepts and attributes that have already been covered by other captions.
We then start from the noun-phrases with the lowest remaining frequency, add the captions that contain this noun-phrase to the candidate pool. 
If the frequency of the noun-phrase is larger than $50$, we randomly choose a subset of size $50$ out of all its captions.
This results in $\sim 40$M image-text pairs that can be successfully downloaded.
The comparison of noun-phrase statistics before and after filtering \icdatasetname{} is shown in \Fref{fig:ic_noun_phrase}. 
The filtered dataset shows a good coverage of concepts whose frequency is higher from 20, but with a smaller number of image-text pairs.
Finally, we get $\sim 52$M image--caption pairs as \icdatasetname{}.

\subsection{Caption Synthesis}
\label{sec:appendix:ic_data:caption_synthesis}
As stated in \Sref{sec:appendix_data:caption_synthesis}, we follow BLIP and LAION-COCO to synthesize captions for \datasetname{}.
However, for \icdatasetname{}, their images are more diverse and the synthesized COCO-style captions may lack in-depth real-world details.
Following CapsFusion~\citep{yu2023capsfusion}, we synthesize \texttt{caption\_capsfusion} based on \texttt{caption\_origin} and \texttt{caption\_coco} for each image, except for JourneyDB.
Their GPT3.5-rewrote captions are already well-structured and contains detailed real-world knowledge, which can be directly seen as \texttt{caption\_capsfusion}.
In our preliminary experiments, we found that using both \texttt{caption\_coco} and \texttt{caption\_capsfusion} together is significantly better than using only one of them.
The final data template (image-first as an example) is:\looseness=-1
\begin{equation}
	\begin{aligned}
		& \text{Image: \texttt{[image]}} \\
		& \text{Caption: \texttt{[caption\_coco]}} \\
		& \text{Detailed caption: \texttt{[caption\_capsfusion]}} \\  \nonumber
	\end{aligned}
\end{equation}

\section{Details of SFT Data}
\label{sec:appendix:sft_data}
In this section, we provide the details of constructing our SFT data. The statistics of the collected dataset and final constructed  dataset are shown in  \Tref{tab:sft_statistics_1} and \Tref{tab:sft_statistics_2}, respectively.

\subsection{Dataset Collection}
The dataset we collected are shown in \Tref{tab:sft_statistics_1}. 
Note that we playback $1$M samples from \datasetname{} to avoid forgetting the knowledge in the pre-training stage. 
The preprocessing details are the same as \Sref{sec:appendix:dataset}. We also sample $1$M data from LAION-COCO-Aesthetic~\citep{laioncocoaesthetic} to keep the image captioning and text-to-image generation ability of the model, with half used as image captioning and half used as text-to-image generation respectively.

\subsection{Sample Generation}
We use the template shown in \Tref{tab:sft_template} to format our collected dataset to training samples following according to the dataset task. 
For text-to-image generation and image editing, we follow the template provided by VL-GPT~\citep{zhu2023vl}. 
For other task, we follow the template provided by LLaVA v1.5~\citep{liu2023improvedllava}.
We follow the system prompt used by VL-GPT, which is ``You are a helpful, respectful and honest assistant. Always answer as helpfully as possible, while being safe. Your answers should not include any harmful, unethical, racist, sexist, toxic, dangerous, or illegal content. Please ensure that your responses are socially unbiased and positive in nature''. 
The final statistics of the constructed dataset are shown in Table \Tref{tab:sft_statistics_2}.

\begin{table}[t]
    \centering
    \tablestyle{3pt}{1.1}
    \caption{
        Statistics of the collected SFT data.
		VG denotes Visual Genome~\citep{krishna2017visual}.
    }
    \label{tab:sft_statistics_1}
	\vspace{1em}
    \scalebox{0.8}{
		\begin{tabular}{lllc}
			\toprule
			Type   & Task & Datasets Involved & \# Samples \\
			\midrule
			\multirow{5}{*}{Mutlimodal Understanding} & Image conversation & UniMM-Chat, Llavar, LLaVA & 0.37M       \\
			& Open-ended VQA     & VQAv2, GQA, OKVQA, OCRVQA & 0.24M       \\
			& Multi-chocie VQA   & A-OKVQA & 0.07M        \\
			& Detailed Caption   & ShareGPT4V, LaionGPT4V & 0.11M      \\
			& Image Caption      & LAION-COCO-Aesthetic, TextCaps& 0.52M      \\
			\midrule
			\multirow{2}{*}{Mutlimodal Generation}    & Image Editing      & Instructpix2pix, MagicBrush   & 0.32M      \\
			& Text2Image & LAION-COCO-Aesthetic    & 0.50M       \\
			\midrule
			Text-only  & Text-only & Alpaca, ShareGPT & 0.09M      \\
			\midrule
			\multirow{2}{*}{\datasetname{} Playback}& Image-first        & Open Images, Objects365, V3Det, VG & 0.50M       \\
			& Text-first& Open Images, Objects365, V3Det, VG & 0.50M      \\
			\bottomrule
		\end{tabular}
    }
\end{table}

\begin{table}[t]
    \centering
    \tablestyle{6pt}{1.2}
    \caption{
        Statistics of the tokenized SFT data.
    }
    \label{tab:sft_statistics_2}
	\vspace{1em}
    \scalebox{0.9}{
		\begin{tabular}{l|ccc}
			\toprule
			Type  & \# Visual Tokens & \# Textual Tokens & \# Samples \\ \midrule
			Mutlimodal Understanding & 115M  & 355.0M & 1.31M        \\
			Mutlimodal Generation    & 104M & { }{ }90.7M& 0.82M       \\
			Text-only      & { }{ }{ }{ }0M    & { }{ }14.1M & 0.09M      \\
			\datasetname{} Playback   & 602M & 579.0M & 1.00M \\
			Total & \textbf{821M}  & \textbf{1038.8M}{ }{ } & \textbf{3.21M}     \\
			\bottomrule
		\end{tabular}
    }
\end{table}

\begin{table}[t]
    \centering
    \tablestyle{2pt}{1.1}
    \caption{
		SFT data template. 
		Each round starts with \texttt{<s>} and end ends with \texttt{</s>}. For each round, the instruction is the content between \texttt{[INST]} and \texttt{[/INST]}, the other content is the response.
		``Detail caption instruction'' means one of our hand-crafted detailed caption instructions, e.g., ``Explain the visual content of the image in great detail.'', and ``Analyse the image in a comprehensive and detailed manner.''.
		``Editing instruction'' is sample-specific, e.g., ``Change the table to a dog.'', and ``Remove one potted plant''.
    }
    \label{tab:sft_template}
	\vspace{1em}
    \scalebox{0.8}{
		\begin{tabular}{l|p{40em}}
			\toprule
			Task  & \multicolumn{1}{c}{Template} \\ \midrule
			Image Conversation & <s> [INST] {<}{<}SYS{>}{>}\textbackslash{}n\{SFT\_SYSTEM\_PROPMT\}\textbackslash{}n{<}{<}/SYS{>}{>}\textbackslash{}n\textbackslash{}n[IMG]\textbackslash{}n\{question\} [/INST] \{response\} </s><s>[INST] \{question\} [/INST] \{response\} </s> \\ \midrule
			Open-ended VQA     & <s> [INST] {<}{<}SYS{>}{>}\textbackslash{}n\{SFT\_SYSTEM\_PROPMT\}\textbackslash{}n{<}{<}/SYS{>}{>}\textbackslash{}n\textbackslash{}n[IMG]\textbackslash{}n\{question\}\textbackslash{}nAnswer the question using a single word or phrase. [/INST] \{response\} </s>\\ \midrule
			Multi-Choice   & <s> [INST] {<}{<}SYS{>}{>}\textbackslash{}n\{SFT\_SYSTEM\_PROPMT\}\textbackslash{}n{<}{<}/SYS{>}{>}\textbackslash{}n\textbackslash{}n[IMG]\textbackslash{}n\{question\}\textbackslash{}nAnswer with the option’s letter from the given choices directly. [/INST] \{response\} </s>          \\ \midrule
			Detailed Caption   & <s> [INST] {<}{<}SYS{>}{>}\textbackslash{}n\{SFT\_SYSTEM\_PROPMT\}\textbackslash{}n{<}{<}/SYS{>}{>}\textbackslash{}n\textbackslash{}n[IMG]\textbackslash{}n\{``Detail caption instruction''\} [/INST] \{response\} </s>        \\ \midrule
			Image Caption      & <s> [INST] {<}{<}SYS{>}{>}\textbackslash{}n\{SFT\_SYSTEM\_PROPMT\}\textbackslash{}n{<}{<}/SYS{>}{>}\textbackslash{}n\textbackslash{}n[IMG]\textbackslash{}nProvide a one-sentence caption for the provided image. [/INST] \{response\} </s>  \\ \midrule
			Image Editing      & <s> [INST] {<}{<}SYS{>}{>}\textbackslash{}n\{SFT\_SYSTEM\_PROMPT\}\textbackslash{}n{<}{<}/SYS{>}{>}\textbackslash{}n\textbackslash{}n[IMG] \{``Editing instruction''\} [/INST] Here is the edited image: [IMG] </s>     \\ \midrule
			Text2Image         & <s> [INST] {<}{<}SYS{>}{>}\textbackslash{}n\{SFT\_SYSTEM\_PROMPT\}\textbackslash{}n{<}{<}/SYS{>}{>}\textbackslash{}n\textbackslash{}nCreate an image that visually represents the description: \{Caption\}  [/INST] Here is the image: [IMG] </s>       \\ \midrule
			Text-only          & <s> [INST] {<}{<}SYS{>}{>}\textbackslash{}n\{SFT\_SYSTEM\_PROPMT\}\textbackslash{}n{<}{<}/SYS{>}{>}\textbackslash{}n\textbackslash{}n\{question\} [/INST] \{response\} </s><s>[INST] \{question\} [/INST] \{response\} </s>   \\ \midrule
			\datasetname{} (Text-first)   & <s> [INST] {<}{<}SYS{>}{>}\textbackslash{}n\{SFT\_SYSTEM\_PROPMT\}\textbackslash{}n{<}{<}/SYS{>}{>}\textbackslash{}n\textbackslash{}n\{Caption and fine-grained annotations in the image-level part\} [/INST] \{Image in the image-level part\} </s><s>[INST] \{Location and object label in the related region\} [/INST] \{Region of the related region\} </s>  \\ \midrule
			\datasetname{} (Image-first)  & <s> [INST] {<}{<}SYS{>}{>}\textbackslash{}n\{SFT\_SYSTEM\_PROPMT\}\textbackslash{}n{<}{<}/SYS{>}{>}\textbackslash{}n\textbackslash{}n\{Image in the image-level part\} [/INST] \{Caption and fine-grained annotations in the image-level part\} </s><s>[INST] \{Region of the related region\} [/INST] \{Location and object label in the related region\} </s> \\    
			\bottomrule         
		\end{tabular}
    }
\end{table}

\begin{table}[p]
    \centering
    \tablestyle{4pt}{1.1}
    \caption{
        Prompts for the object label description generation for Open Images and Objects365.
    }
    \label{tab:label_des_gen_OI_object}
	\vspace{1em}
    \scalebox{0.85}{
        \begin{tabular}{p{15cm}}
            \toprule
            \underline{\textbf{\textit{System Prompt}}} \\
			\# Instructions for Object Category Description Generation \\ 
			\\
			You are a helpful, respectful, and honest assistant. \\
			Now you are an expert in generating descriptions for category labels of objects in an image. \\
			You are given some examples, each with their object category label and description. \\
			Your goal is to generate a description of a given object category label to help people better understand it from both vision and language modalities. \\
			Informative, concise, accurate and clear descriptions are expected. \\ \\
			Here are some useful tips for generating object category descriptions: \\
			1. Universality: Focus on features common to most instances of the object category. \\
			2. Multiple Semantics: Select the meaning of the most relevant and likely object category label in an image. For instance, as an object category label, "bank" can be "a financial institution" or "a landform alongside a river", and the former is more likely to be the meaning in an image. \\
			3. Distinctive Features: Emphasize unique aspects differentiating the object from other similar objects. \\
			4. Relevant Knowledge: Incorporate important concepts, historical, or cultural information that enrich the understanding of the object but avoid excessive details. \\
			5. You can optionally focus on the following aspects when generating category descriptions: \\
			\quad 1. What are useful features for distinguishing the object of the given category label in an image? \\
			\quad 2. What does the object of the given category label in an image look like? \\
			\quad 3. What are the identifying characteristics of the object with the given category label in an image? \\
			\quad 4. What are the key visual indicators that help identify the object category label in an image? \\ \\
			
			Here are some examples: \\ \\
			
			``` \\
			Object Category Label: malayan tapir \\ \\
			
			Category Description: The Malayan tapir, a distinctive mammal found predominantly in Southeast Asia, is known for its unique coloration: It boasts a unique appearance, featuring a black body with a white front section. It resembles a large pig in shape, with a short, prehensile trunk, small eyes, and pointed ears. Primarily nocturnal, tapirs are herbivorous and thrive in dense forests near water sources. \\
			``` \\	
			\\
			``` \\
			Object Category Label: sheep \\ \\
			
			Category Description: Sheep are medium-sized, quadrupedal mammals, typically covered in curly woolen fleece that varies from white to brown. Recognizable by their stout, fluffy appearance, they have a distinctive head with a rounded snout and long, sometimes spiraled, ears. Often seen grazing in herds, sheep are known for their gentle demeanor and are primarily raised for wool, meat, and milk. \\
			``` \\	
			\\
			``` \\
			Object Category Label: compass \\ \\
			
			Category Description: A compass is a navigational instrument typically used for orientation and direction finding. It consists of a magnetized needle that aligns with Earth's magnetic field, pointing towards magnetic north. The device usually has a circular scale marked with directions (North, South, East, West) and degrees for precise navigation. \\
			``` \\	
			
			\underline{\textbf{\textit{User Prompt}}} \\
			\# Object Category Description Generation \\ \\
			
			Please directly generate an informative, concise, accurate and clear description for the given object category label in about 50 words. \\ \\
			
			``` \\
			Object Category Label: \{category label\} \\
			
			Category Description:\quad \\

            \bottomrule
        \end{tabular}
    }
\end{table}

\begin{table}[p]
    \centering
    \tablestyle{4pt}{1.1}
    \caption{
        Prompts for the attribute label description generation for Open Images. Note that there are 3 examples with 2 examples omitted.
    }
    \label{tab:label_des_gen_OI_attribute}
	\vspace{1em}
    \scalebox{0.85}{
        \begin{tabular}{p{15cm}}
            \toprule
            \underline{\textbf{\textit{System Prompt}}} \\

			\# Instructions for Attribute Category Description Generation                                                                                                                                                                                                                                                                                      \\
			\\
			You are a helpful, respectful, and honest assistant.                                                                                                                                                                                                                                                                                                        \\
			Now you are an expert in generating descriptions for attribute category labels of objects in an image.                                                                                                                                                                                                                                                      \\
			\\
			You are given some examples, each example is in the following format:                                                                                                                                                                                                                                                                                       \\
			\\
			```                                                                                                                                                                                                                                                                                                                                                         \\
			Attribute Category Label: the attribute category label required to generate a description.                                                                                                                                                                                                                                                                  \\
			\\
			Related Object Category Labels: the category labels of the objects which may be associated with the attribute category label, separated by commas.                                                                                                                                                                                                          \\
			\\
			Attribute Category Description: the generated description of the attribute category label.                                                                                                                                                                                                                                                                  \\
			```                                                                                                                                                                                                                                                                                                                                                         \\
			\\
			Your goal is to generate a description of a given attribute category label to help people better understand it from both vision and language modalities.                                                                                                                                                                                                    \\
			Informative, concise, accurate and clear descriptions are expected.                                                                                                                                                                                                                                                                                         \\
			\\
			Here are some useful tips for generating attribute category descriptions:                                                                                                                                                                                                                                                                                   \\
			\\
			1. Universality: Focus on features common to most instances of the attribute category.                                                                                                                                                                                                                                                                      \\
			2. Multiple Semantics: Select the meaning of the most relevant and likely attribute category label in an image. For instance, as an attribute category label, "awake" is an adjective meaning "not asleep" rather than a verb meaning "to stop sleeping or wake up from sleep".                                                                             \\
			3. Distinctive Features: Emphasize unique aspects differentiating the attribute from other similar attributes.                                                                                                                                                                                                                                              \\
			4. Appearance, Sensory and Effect: Describe how the attribute typically appears or is perceived on the objects. This could include color, texture, size, sense, function, or any other aspect that is significant for the attribute.                                                                                                                        \\
			5. You can optionally focus on the following aspects when generating attribute category descriptions:                                                                                                                                                                                                                                                       \\
			\quad 1. How is the attribute category label usually observed or sensed?                                                                                                                                                                                                                                                                                          \\
			\quad 2. What impact does this attribute have on the object, its perception or function?                                                                                                                                                                                                                                                                          \\
			\quad 3. What are the identifying characteristics of the given category label of an object in an image?                                                                                                                                                                                                                                                           \\
			\quad 4. What are the key visual indicators that help identify the attribute category label in an image?                                                                                                                                                                                                                                                          \\
			\\
			Here are some examples:                                                                                                                                                                                                                                                                                                                                     \\
			```                                                                                                                                                                                                                                                                                                                                                         \\
			Attribute Category Label: smiling                                                                                                                                                                                                                                                                                                                           \\
			\\
			Related Object Category Labels: Man, Woman, Girl, Boy, Baby, Child, Animal, ...                                                                                                                                                                                                                                                                             \\
			\\
			Attribute Category Description: Smiling is a facial gesture where the mouth corners lift upwards, often revealing the front teeth, and is usually associated with emotions like happiness, kindness, or amusement. It generally signifies a person feeling joyful or content, making it a universally understood symbol of positive emotions.               \\
			```                                                                                                                               
			\\

			......      
			
			\underline{\textbf{\textit{User Prompt}}} \\

			\# Attribute Category Description Generation                                                                                               \\ \\
			Please directly generate an informative, concise, accurate and clear description for the given attribute category label in about 50 words. \\ \\
			```                                                                                                                                        \\
			Attribute Category Label: \{attribute category\_label\}                                                                                    \\ \\
			Related Object Category Labels: \{related object category labels\}, ...                                                                    \\ \\
			Attribute Category Description: \quad \\

            \bottomrule
        \end{tabular}
    }
\end{table}

\begin{table}[p]
    \centering
    \tablestyle{4pt}{1.1}
    \caption{
        Prompts for the relationship label description generation for Open Images. Note that there are 3 examples with 2 examples omitted.
    }
    \label{tab:label_des_gen_OI_relationship}
	\vspace{1em}
    \scalebox{0.85}{
        \begin{tabular}{p{15cm}}
            \toprule
            \underline{\textbf{\textit{System Prompt}}} \\
			\# Instructions for Relationship Category Description Generation                                                                                                                                                                                                                                                                                                                          \\
			\\
			You are a helpful, respectful, and honest assistant.                                                                                                                                                                                                                                                                                                                                      \\
			Now you are an expert in generating descriptions for relationship category labels of objects in an image.                                                                                                                                                                                                                                                                                 \\
			\\
			You are given some examples, each example is in the following format:                                                                                                                                                                                                                                                                                                                     \\
			\\
			```                                                                                                                                                                                                                                                                                                                                                                                       \\
			Relationship Category Label: the relationship category label required to generate a description.                                                                                                                                                                                                                                                                                          \\
			\\
			Related Subject-Object Pairs: the subject-object category label pairs associated with the relationship category label. Each pair in the form of {[}Subject Category Label, Object Category Label{]}, different pairs are separated by commas.                                                                                                                                             \\
			\\
			Relationship Category Description: the generated description of the relationship category label.                                                                                                                                                                                                                                                                                          \\
			```                                                                                                                                                                                                                                                                                                                                                                                       \\
			\\
			Your goal is to generate a description of a given relationship label to help people better understand it from both vision and language modalities.                                                                                                                                                                                                                                        \\
			Informative, concise, accurate and clear descriptions are expected.                                                                                                                                                                                                                                                                                                                       \\
			\\
			Here are some useful tips for generating attribute category descriptions:                                                                                                                                                                                                                                                                                                                 \\
			\\
			1. Universality:Focus on features common to most instances of the relationship category.                                                                                                                                                                                                                                                                                                  \\
			2. Multiple Semantics: Select the meaning of the most relevant and likely relationship category label in an image. For instance, as a relationship category label, "truck" is a verb meaning "convey by truck" rather than a noun meaning "a large, heavy road vehicle used for carrying goods and materials".                                                                            \\
			3. Distinctive Features: Emphasize unique aspects differentiating the relationship from other similar relationships.                                                                                                                                                                                                                                                                      \\
			4. Nature of Relationship: Describe the nature of the relationship (e.g., spatial, action, functional, hierarchical, temporal, social) and how the subject and object interact or relate to each other.                                                                                                                                                                                   \\
			5. You can optionally focus on the following aspects when generating attribute category descriptions:                                                                                                                                                                                                                                                                                     \\
			\quad 1. How is the relationship expressed or manifested in the image?                                                                                                                                                                                                                                                                                                                          \\
			\quad 2. What are the key characteristics or significance of the relationship?                                                                                                                                                                                                                                                                                                                  \\
			\quad 3. What are common or typical scenarios in which this relationship is observed?                                                                                                                                                                                                                                                                                                           \\
			\quad 4. What are the key visual indicators that help identify the relationship category in an image?                                                                                                                                                                                                                                                                                           \\
			\\
			Here are some examples:                                                                                                                                                                                                                                                                                                                                                                   \\
			```                                                                                                                                                                                                                                                                                                                                                                                       \\
			Relationship Category Label: on                                                                                                                                                                                                                                                                                                                                                           \\
			\\
			Related Subject-Object Pairs: {[}Bell pepper, Countertop{]}, {[}Woman, Bicycle{]}, {[}Tomato, Cutting board{]}, ...                                                                                                                                                                                                                                                                       \\
			\\
			Relationship Category Description: The relationship label "on" indicates that the subject is positioned above and in contact with the surface of the object, without being suspended or supported by anything else.  This implies a direct interaction where an object rests upon or integrates into the surface, thereby becoming a prominent or integral part of its overall structure. \\
			```  \\

			......      \\

			\underline{\textbf{\textit{User Prompt}}} \\

			\# Relationship Category Description Generation                                                                                               \\
			\\
			Please directly generate an informative, concise, accurate and clear description for the given relationship category label in about 50 words. \\
			\\
			```                                                                                                                                           \\
			Relationship Category Label: \{relationship category\_label\}                                                                                 \\
			\\
			Related Subject-Object Pairs: \{related subject-object pairs\}, ...                                                                           \\
			\\
			Relationship Category Description:   \quad \\

            \bottomrule
        \end{tabular}
    }
\end{table}

\begin{table}[p]
    \centering
    \tablestyle{4pt}{1.1}
    \caption{
        Prompts for the relationship object label description generation for Visual Genome. Note that there are 3 examples with 2 examples omitted.
    }
    \label{tab:label_des_gen_VG_object}
	\vspace{1em}
    \scalebox{0.85}{
        \begin{tabular}{p{15cm}}
            \toprule
            \underline{\textbf{\textit{System Prompt}}} \\
			
			\# Instructions for Object Category Description Generation                                                                                                                                                                                                                                                                                                                                                                                                                                                             \\
			\\
			You are a helpful, respectful, and honest assistant.                                                                                                                                                                                                                                                                                                                                                                                                                                                                   \\
			Now you are an expert in generating descriptions for category labels of objects in an image.                                                                                                                                                                                                                                                                                                                                                                                                                           \\
			You are given some examples, each with their object category label and description.                                                                                                                                                                                                                                                                                                                                                                                                                                    \\
			Your goal is to generate a description of a given object category label to help people better understand it from both vision and language modalities.                                                                                                                                                                                                                                                                                                                                                                  \\
			Informative, concise, accurate and clear descriptions are expected.                                                                                                                                                                                                                                                                                                                                                                                                                                                    \\
			\\
			Attention, the given object category label may be invalid. In such cases, you should generate "Invalid." as its description and then explain why it is invalid after "Invalid.".                                                                                                                                                                                                                                                                                                                                       \\
			\\
			Here are some useful tips for generating object category descriptions when the given object category label is valid:                                                                                                                                                                                                                                                                                                                                                                                                   \\
			\\
			1. Universality: Focus on features common to most instances of the object category.                                                                                                                                                                                                                                                                                                                                                                                                                                    \\
			2. Multiple Semantics: Select the meaning of the most relevant and likely object category label in an image. For instance, as an object category label, "bank" can be "a financial institution" or "a landform alongside a river", and the former is more likely to be the meaning in an image.                                                                                                                                                                                                                        \\
			3. Distinctive Features: Emphasize unique aspects differentiating the object from other similar objects.                                                                                                                                                                                                                                                                                                                                                                                                               \\
			4. Relevant Knowledge: Incorporate important concepts, historical, or cultural information that enrich the understanding of the object but avoid excessive details.                                                                                                                                                                                                                                                                                                                                                    \\
			5. You can optionally focus on the following aspects when generating category descriptions:                                                                                                                                                                                                                                                                                                                                                                                                                            \\
			\quad 1. What are useful features for distinguishing the object of the given category label in an image?                                                                                                                                                                                                                                                                                                                                                                                                                     \\
			\quad 2. What does the object of the given category label in an image look like?                                                                                                                                                                                                                                                                                                                                                                                                                                             \\
			\quad 3. What are the identifying characteristics of the object with the given category label in an image?                                                                                                                                                                                                                                                                                                                                                                                                                   \\
			\quad4. What are the key visual indicators that help identify the object category label in an image?                                                                                                                                                                                                                                                                                                                                                                                                                        \\
			\\
			Here are some examples:                                                                                                                                                                                                                                                                                                                                                                                                                                                                                                \\
			\\
			```                                                                                                                                                                                                                                                                                                                                                                                                                                                                                                                    \\
			Object Category Label: malayan tapir                                                                                                                                                                                                                                                                                                                                                                                                                                                                                   \\
			\\
			Category Description: The Malayan tapir, a distinctive mammal found predominantly in Southeast Asia, is known for its unique coloration: It boasts a unique appearance, featuring a black body with a white front section. It resembles a large pig in shape, with a short, prehensile trunk, small eyes, and pointed ears. Primarily nocturnal, tapirs are herbivorous and thrive in dense forests near water sources.                                                                                                \\
			```                                                                                                                                                                                                                                                                                                                                                                                                                                                                                                                    \\
			......
			
			\\

			\\   
			\underline{\textbf{\textit{User Prompt}}} \\

			\# Object Category Description Generation                                                                                               \\
			\\
			Please directly generate an informative, concise, accurate and clear description for the given object category label in about 50 words. \\
			\\
			```                                                                                                                                     \\
			Object Category Label: \{category\_label\}                                                                                              \\
			\\
			Category Description:  

			\quad \\
			
            \bottomrule
        \end{tabular}
    }
\end{table}

\begin{table}[p]
    \centering
    \tablestyle{4pt}{1.1}
    \caption{
        Prompts for the attribute label description generation for Visual Genome. The examples are omitted.
    }
    \label{tab:label_des_gen_VG_attribute}
	\vspace{1em}
    \scalebox{0.85}{
        \begin{tabular}{p{15cm}}
            \toprule
            \underline{\textbf{\textit{System Prompt}}} \\
			
			\# Instructions for Attribute Category Description Generation                                                                                                                                                                                                                                                                                                                           \\
			\\
			You are a helpful, respectful, and honest assistant.                                                                                                                                                                                                                                                                                                                                    \\
			Now you are an expert in generating descriptions for attribute category labels of objects in an image.                                                                                                                                                                                                                                                                                  \\
			Your goal is to generate a description of the given attribute category label to help people better understand it from both vision and language modalities.                                                                                                                                                                                                                              \\
			Informative, concise, accurate and clear descriptions are expected.                                                                                                                                                                                                                                                                                                                     \\
			\\
			You are given some examples, each example is in the following format:                                                                                                                                                                                                                                                                                                                   \\
			\\
			```                                                                                                                                                                                                                                                                                                                                                                                     \\
			Attribute Category Label: the attribute category label required to generate a description.                                                                                                                                                                                                                                                                                              \\
			\\
			Related Object Category Labels: the category labels of the objects which may be associated with the attribute category label, separated by commas.                                                                                                                                                                                                                                      \\
			\\
			Attribute Category Description: the generated description of the attribute category label.                                                                                                                                                                                                                                                                                              \\
			```                                                                                                                                                                                                                                                                                                                                                                                     \\
			\\
			Attention, the given attribute category label and related object category labels may be noisy. Specifically:                                                                                                                                                                                                                                                                            \\
			\\
			1. The given related object category labels may be invalid or not be associated with the given attribute category label. You should disregard such noisy related object category labels.                                                                                                                                                                                                \\
			2. The given attribute category label may be invalid. You should generate "Invalid." as its description and then explain why it is invalid after "Invalid.".                                                                                                                                                                                                                            \\
			\\
			Here are some useful tips for generating attribute category descriptions when the given attribute category label is valid:                                                                                                                                                                                                                                                              \\
			\\
			1. Universality: Focus on features common to most instances of the attribute category.                                                                                                                                                                                                                                                                                                  \\
			2. Multiple Semantics: Select the meaning of the most relevant and likely attribute category label in an image. For instance, as an attribute category label, "awake" is an adjective meaning "not asleep" rather than a verb meaning "to stop sleeping or wake up from sleep".                                                                                                         \\
			3. Distinctive Features: Emphasize unique aspects differentiating the attribute from other similar attributes.                                                                                                                                                                                                                                                                          \\
			4. Appearance, Sensory and Effect: Describe how the attribute typically appears or is perceived on the objects. This could include color, texture, size, sense, function, or any other aspect that is significant for the attribute.                                                                                                                                                    \\
			5. You can optionally focus on the following aspects when generating attribute category descriptions:                                                                                                                                                                                                                                                                                   \\
			\quad 1. How is the attribute category label usually observed or sensed?                                                                                                                                                                                                                                                                                                                      \\
			\quad 2. What impact does this attribute have on the object, its perception or function?                                                                                                                                                                                                                                                                                                      \\
			\quad 3. What are the identifying characteristics of the given category label of an object in an image?                                                                                                                                                                                                                                                                                       \\
			\quad 4. What are the key visual indicators that help identify the attribute category label in an image?                                                                                                                                                                                                                                                                                      \\
			\\
			Here are some examples:                                                                                                                                                                                                                                                                                                                                                                 \\
			\\ \{examples\}
			                                                                                                                                                                                                                                                                                                                                                                                                                                                                                                                                                                                                                              \\
	
			......

			\\   
			\underline{\textbf{\textit{User Prompt}}} \\

			\# Attribute Category Description Generation                                                                                               \\ \\
			Please directly generate an informative, concise, accurate and clear description for the given attribute category label in about 50 words. \\ \\
			```                                                                                                                                        \\
			Attribute Category Label: \{attribute category\_label\}                                                                                    \\ \\
			Related Object Category Labels: \{related object category labels\}, ...                                                                    \\ \\
			Attribute Category Description:

			\quad \\

            \bottomrule
        \end{tabular}
    }
\end{table}

\begin{table}[p]
    \centering
    \tablestyle{4pt}{1.1}
    \caption{
        Prompts for the relationship label description generation for Visual Genome. The examples are all omitted.
    }
    \label{tab:label_des_gen_VG_relationship}
	\vspace{1em}
    \scalebox{0.85}{
        \begin{tabular}{p{15cm}}
            \toprule
            \underline{\textbf{\textit{System Prompt}}} \\
			
			\# Instructions for Relationship Category Description Generation                                                                                                                                                                                                                                               \\
			\\
			You are a helpful, respectful, and honest assistant.                                                                                                                                                                                                                                                           \\
			Now you are an expert in generating descriptions for relationship category labels of objects in an image.                                                                                                                                                                                                      \\
			Your goal is to generate a description of a given relationship label to help people better understand it from both vision and language modalities.                                                                                                                                                             \\
			Informative, concise, accurate and clear descriptions are expected.                                                                                                                                                                                                                                            \\
			\\
			You are given some examples, each example is in the following format:                                                                                                                                                                                                                                          \\
			\\
			```                                                                                                                                                                                                                                                                                                            \\
			Relationship Category Label: the relationship category label required to generate a description.                                                                                                                                                                                                               \\
			\\
			Related Subject-Object Pairs: the subject-object category label pairs associated with the relationship category label. Each pair in the form of {[}Subject Category Label, Object Category Label{]}, different pairs are separated by commas.                                                                  \\
			\\
			Relationship Category Description: the generated description of the relationship category label.                                                                                                                                                                                                               \\
			```                                                                                                                                                                                                                                                                                                            \\
			\\
			Attention, the given relationship category label and related subject-object pairs may be noisy. Specifically:                                                                                                                                                                                                  \\
			1. The given related subject-object pairs may be invalid or not be associated with the given relationship category label. You should disregard such noisy related subject-object pairs.                                                                                                                        \\
			2. The given relationship category label may be invalid. You should generate "Invalid." as its description and then explain why it is invalid after "Invalid.".                                                                                                                                                \\
			\\
			Here are some useful tips for generating attribute category descriptions when the given relationship category label is valid:                                                                                                                                                                                  \\
			\\
			1. Universality: Focus on features common to most instances of the relationship category.                                                                                                                                                                                                                      \\
			2. Multiple Semantics: Select the meaning of the most relevant and likely relationship category label in an image. For instance, as a relationship category label, "truck" is a verb meaning "convey by truck" rather than a noun meaning "a large, heavy road vehicle used for carrying goods and materials". \\
			3. Distinctive Features: Emphasize unique aspects differentiating the relationship from other similar relationships.                                                                                                                                                                                           \\
			4. Nature of Relationship: Describe the nature of the relationship (e.g., spatial, action, functional, hierarchical, temporal, social) and how the subject and object interact or relate to each other.                                                                                                        \\
			5. You can optionally focus on the following aspects when generating attribute category descriptions:                                                                                                                                                                                                          \\
			\quad 1. How is the relationship expressed or manifested in the image?                                                                                                                                                                                                                                               \\
			\quad 2. What are the key characteristics or significance of the relationship?                                                                                                                                                                                                                                       \\
			\quad 3. What are common or typical scenarios in which this relationship is observed?                                                                                                                                                                                                                                \\
			\quad 4. What are the key visual indicators that help identify the relationship category in an image?                                                                                                                                                                                                                \\
			\\
			Here are some examples: \\ \{examples\}
			                                                                                                                                                                                                                                                                                                                                                                                                                                                                                             \\
	
			......

			\\   
			\underline{\textbf{\textit{User Prompt}}} \\
			
			 \# Relationship Category Description Generation                                                                                               \\
			 \\
			 Please directly generate an informative, concise, accurate and clear description for the given relationship category label in about 50 words. \\
			 \\
			 ```                                                                                                                                           \\
			 Relationship Category Label: \{relationship category\_label\}                                                                                 \\
			 \\
			 Related Subject-Object Pairs: \{related subject-object pairs\}, ...                                                                           \\
			 \\
			 Relationship Category Description: \quad

			 \\

            \bottomrule
        \end{tabular}
    }
\end{table}

\begin{table}[p]
    \centering
    \tablestyle{4pt}{1.1}
    \caption{
       Image-first template. The instructions for the related region part will be repeated if the image has more than one region.
    }
    \label{tab:pretrain_template_image_first}
	\vspace{1em}
    \scalebox{0.85}{
        \begin{tabular}{p{15cm}}
            \toprule

			\textbf{\textit{image-first template with descriptions}} \\ \\
			\# Detailed Analysis of Objects in the Image \\ \\
			
			Image: \texttt{[IMG]} \\
			Caption: [image caption] \\
			Localized narrative caption: [localized narrative caption] \\
			
			Objects and their descriptions: \\
			- [object label]: [object label des] \\
			- [object label]: [object label des] \\
            Attributes of objects and their descriptions: \\
            - [attribute label] {[object label], [object label]} : [attribute label des] \\
            - [attribute label] {[object label], [object label]} : [attribute label des] \\

            Relationships between objects and their descriptions: \\
            - [relationship label] {[subject label]-[object label], [subject label]-[object label]} : [relationship label des] \\
            - [relationship label] {[subject label]-[object label], [subject label]-[object label]} : [relationship label des] \\

            \#\# Overview of Selected Object Regions \\ \\
            \#\#\# Overview of a Selected Object Region \\ \\
            Region: \texttt{[IMG]} \\ 
            Location of the selected region in the image: [location] \\
            Objects: \\
            - [object label] \\
            - [object label] \\ \\

            \textbf{\textit{image-first template without descriptions}} \\ \\
			\# Detailed Analysis of Objects in the Image \\ \\
			
			Image: \texttt{[IMG]} \\
			Caption: [image caption] \\
			Localized narrative caption: [localized narrative caption] \\
			
			Objects: \\
			- [object label] \\
			- [object label] \\
			Attributes of objects: \\
			- [attribute label] {[object label], [object label]} \\
			- [attribute label] {[object label], [object label]} \\

			Relationships between objects: \\
			- [relationship label] {[subject label]-[object label], [subject label]-[object label]} \\
			- [relationship label] {[subject label]-[object label], [subject label]-[object label]} \\

			\#\# Overview of Selected Object Regions \\ \\
			\#\#\# Overview of a Selected Object Region \\ \\
			Region: \texttt{[IMG]} \\ 
			Location of the selected region in the image: [location] \\
			Objects: \\
			- [object label] \\
			- [object label] \\ \\

            \bottomrule
        \end{tabular}
    }
\end{table}

\begin{table}[p]
    \centering
    \tablestyle{4pt}{1.1}
    \caption{
       Text-first template. The instructions for the related region part will be repeated if the image has more than one region.
    }
    \label{tab:pretrain_template_text_first}
    \vspace{1em}
    \scalebox{0.85}{
        \begin{tabular}{p{15cm}}
            \toprule
            
			\textbf{\textit{text-first template with descriptions}} \\ \\
			\# Detailed Analysis of Objects in the Image \\ \\
			Objects and their descriptions: \\
			- [object label]: [object label des] \\
			- [object label]: [object label des] \\
            Attributes of objects and their descriptions: \\
            - [attribute label] {[object label], [object label]} : [attribute label des] \\
            - [attribute label] {[object label], [object label]} : [attribute label des] \\

            Relationships between objects and their descriptions: \\
            - [relationship label] {[subject label]-[object label], [subject label]-[object label]} : [relationship label des] \\
            - [relationship label] {[subject label]-[object label], [subject label]-[object label]} : [relationship label des] \\
            
            Caption: [image caption] \\
            
            Localized narrative caption: [localized narrative caption] \\
            Image: \texttt{[IMG]} \\ \\
            
            \#\# Overview of Selected Object Regions \\ \\
            \#\#\# Overview of a Selected Object Region \\ \\
            Location of the selected region in the image: [location] \\
            Objects: \\
            - [object label] \\
            - [object label] \\
            Region: \texttt{[IMG]} \\ \\

            \textbf{\textit{text-first template without descriptions}} \\ \\
			\# Detailed Analysis of Objects in the Image \\ \\
			Objects and their descriptions: \\
			- [object label] \\
			- [object label] \\
			Attributes of objects: \\
			- [attribute label] {[object label], [object label]} \\
			- [attribute label] {[object label], [object label]} \\
			Relationships between objects: \\
			- [relationship label] {[subject label]-[object label], [subject label]-[object label]} \\
			- [relationship label] {[subject label]-[object label], [subject label]-[object label]} \\
			
			Caption: [image caption] \\ 
			
			Localized narrative caption: [localized narrative caption] \\
			Image: \texttt{[IMG]} \\ \\
			
			\#\# Overview of Selected Object Regions \\ \\
			\#\#\# Overview of a Selected Object Region \\ \\
			Location of the selected region in the image: [location] \\
			Objects: \\
			- [object label] \\
			- [object label] \\
			Region: \texttt{[IMG]} \\

            \bottomrule
        \end{tabular}
    }
\end{table}

\end{document}